\pdfoutput=1
\documentclass[11pt]{article}

\usepackage{acl}
\usepackage{times}
\usepackage{latexsym}
\usepackage{caption}
\usepackage{subcaption}
\usepackage{booktabs}

\newcommand*{\CA}{\texttt{CALIGN}}
\newcommand*{\CO}{\texttt{COLAP}}

\usepackage[T1]{fontenc} 
\usepackage{microtype}

\usepackage{amsmath,amsfonts,bm}

\def\eqref#1{equation~\ref{#1}}
\def\1{\bm{1}}

\DeclareMathAlphabet{\mathsfit}{\encodingdefault}{\sfdefault}{m}{sl}
\SetMathAlphabet{\mathsfit}{bold}{\encodingdefault}{\sfdefault}{bx}{n}

\usepackage[utf8]{inputenc}
\usepackage{arabtex}
\usepackage{utf8}
\setcode{utf8}

\usepackage{dsfont}
\usepackage{comment}

\usepackage{todonotes}

\title{Exploring Alignment in Shared Cross-lingual Spaces}
\author{
 Basel Mousi ~ Nadir Durrani ~ Fahim Dalvi \\  \textbf{Majd Hawasly} ~ \textbf{Ahmed Abdelali}\thanks{\hspace{0.5mm} Ahmed contributed to the project while he was at QCRI.} \\ 
{\tt \{bmousi,ndurrani,faimaduddin\}@hbku.edu.qa} \\ 
Qatar Computing Research Institute, HBKU Research Complex, Doha, Qatar \\
}

\begin{document}
\maketitle
\begin{abstract}

Despite their remarkable ability to capture linguistic nuances across diverse languages, questions persist regarding the degree of alignment between languages in multilingual embeddings. Drawing inspiration from research on high-dimensional representations in neural language models, we employ clustering to uncover latent concepts within multilingual models. Our analysis focuses on quantifying the \textit{alignment} and \textit{overlap} of these concepts across various languages within the latent space. To this end, we introduce two metrics \CA{} and \CO{} aimed at quantifying these aspects, enabling a deeper exploration of multilingual embeddings. Our study encompasses three multilingual  models (\texttt{mT5}, \texttt{mBERT}, and \texttt{XLM-R}) and three downstream tasks (Machine Translation, Named Entity Recognition, and Sentiment Analysis). Key findings from our analysis include: i) deeper layers in the network demonstrate increased cross-lingual \textit{alignment} due to the presence of language-agnostic concepts, ii) fine-tuning of the models enhances \textit{alignment} within the latent space, and iii) such task-specific calibration helps in explaining the emergence of zero-shot capabilities in the models.\footnote{The code is available at \url{https://github.com/qcri/multilingual-latent-concepts}}

\end{list}
\end{abstract}

\section{Introduction}

The emergence of multilingual contextualized embeddings has been a ground-breaking advancement, in the ever-evolving landscape of natural language processing. Adept at capturing the linguistic nuances across different languages, these embeddings have spurred a multitude of studies \cite{pires-etal-2019-multilingual,dufter-schutze-2020-identifying,papadimitriou-etal-2021-deep} seeking to understand the underlying mechanisms. How these models achieve multilinguality without explicit cross-lingual supervision during training is a particularly interesting question to answer.

Cross-lingual embeddings are designed to encode linguistic concepts that bridge equivalent semantic meaning across diverse languages. The question is: how well is this achieved in practice? When considering two arbitrary languages, \textit{how well aligned are the embeddings of those languages?} and \textit{how language agnostic are these multilingual embeddings in reality?} Addressing these questions necessitates a comprehensive approach.
% In this research endeavor, we aim to address these questions by %aligning latent concepts learned within the 
%studying the \textit{alignment} and \textit{overlap} in latent concepts learned within representations of multilingual models. 

\begin{figure*}
    \centering
    \includegraphics[width=\linewidth]{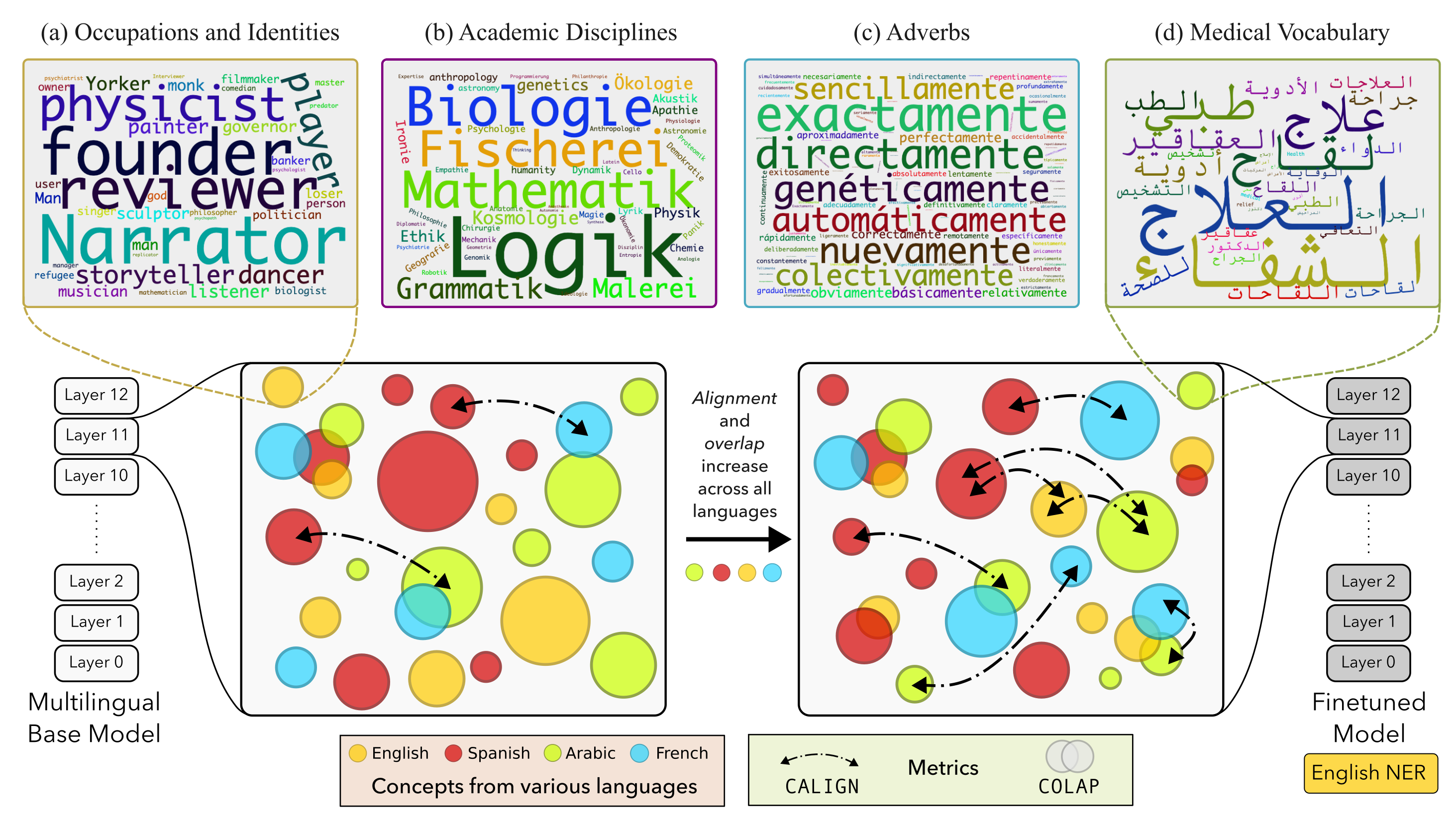}
    \caption{Overview of \CA{} and \CO{} metrics in latent spaces of multilingual models, and how the space re-calibrates after fine-tuning. The top row shows concepts learned in \texttt{mT5} across different languages:  (a) English (b) German, (c) Spanish,  (d) Arabic. }
    \label{fig:overall}
\end{figure*}

In high-dimensional spaces, neural language models exhibit a capability to group words with shared linguistic associations, as highlighted by  
\citet{Mikolov:2013:ICLR}.  
Expanding upon this foundational insight, recent research endeavors \cite{michael-etal-2020-asking,dalvi2022discovering, Yao-Latent} delved into conducting representation analysis within pre-trained models. Our objective, in this work, is to uncover encoded concepts within multilingual models and analyze their \textit{alignment} and \textit{overlap} across various languages within the latent space. We discover latent concepts by applying clustering to the underlying contextualized representations. The premise is that these clusters potentially signify latent concepts, encapsulating the language knowledge assimilated by the model. We build our work on top of this foundation to quantify concept \textit{alignment} and \textit{overlap} within multilingual latent space. We propose two metrics \CA{} and \CO{} to quantify these two aspects and carry out analysis to study the following questions:

%We design our analysis around this key insight and study the following questions:

\begin{itemize}
    \item To what extent do latent spaces across languages exhibit \textit{alignment} and \textit{overlap} in multilingual models?
    \item How does this change as the models are tuned towards any downstream NLP task?
    \item How do the multilingual latent spaces transform for zero-shot scenarios?
\end{itemize}

We conducted a study employing three multilingual transformer models: \texttt{mT5} \cite{xue-etal-2021-mt5}, \texttt{mBERT} \cite{devlin-etal-2019-bert}, and \texttt{XLM-RoBERTa} \cite{xlm-roberta}. These models were fine-tuned for three downstream NLP tasks: machine translation, named-entity recognition and sentiment analysis, spanning sequence generation, labeling and classification respectively. Our analysis revealed intriguing insights, including:

\begin{itemize}

    \item Deeper layers in multilingual models preserve semantic concepts, contrasting with language-dependent lexical learning in lower layers, resulting in a higher alignment.

    \item Fine-tuning calibrates the latent space towards higher alignment and the task-specific calibration of the latent space facilitates zero-shot capabilities.

    \item Divergent patterns emerge in the  encoder and decoder latent spaces in \texttt{seq2seq} models. The final layers  in the decoder tend to primarily retain language specific concepts.

    \item Closely related languages demonstrate higher
overlap in latent space.

    \item The complexity of optimization function affects the extent of overlap in latent spaces

    \item While many model concepts exhibit multilingual traits, later layers post fine-tuning tend to retain primarily language-specific characteristics.

    % \item Multilinguality of latent concepts is attributed to shared lingusitic properties between languages. Multilinguality in earlier layers is due to shared ngrams between languages. Monolingual clusters are due to properties specific to the language itself. As we progress to higher layers, the concepts become more semantic and mutlilinguality improves. 
    % \item Languages written in the same script tend to have higher concept multilinguality.  \textcolor{red}{same as the first point. not sure if we should mention it here}
    
    % \item Zero shot learning is attributed to improved cluster alignment after finetuning.  

    % \item Models tend to encode languages similarly. Identical clusters of names, numbers, and places are found in the same model across different language splits.

    % \item \textcolor{red}{What did we notice about XNLI models?} 

    % \item  \textcolor{red}{What did we notice about NER models?}
    
\end{itemize}

\section{Methodology}
\label{sec:methodology}

% Our work is based on the Latent Concept Analysis  \cite{dalvi2022discovering} used to interpret the representational spaces found in neural network models. We employ clustering on contextualized embeddings to identify what we term ``Encoded Concepts'' within the model. Furthermore, we investigate how these concepts align within multilingual models across various languages. An outline of our approach is depicted in Figure \ref{fig:pipeline}. In the subsequent sections, we elaborate on each stage of our methodology.

The high-dimensional latent spaces learned within neural language models have been shown to encapsulate concepts formed by common linguistic attributes \citep{Mikolov:2013:ICLR,bert_geometry:nips2019}.  Our approach is rooted in this foundational insight where we discover latent concepts for interpreting representational spaces in multilingual neural language models. More precisely, our study endeavors to gauge the degree of \textit{alignment} and \textit{overlap} of concepts across the latent spaces acquired through training models on a diverse array of languages. To this end, we introduce two metrics to quantify these phenomena.
%Our work builds upon the foundation of Latent Concept Analysis \cite{dalvi2022discovering}, a method employed for interpreting representational spaces in neural language models. We extend this framework by introducing two metrics to \emph{quantify} and \emph{analyze} the alignment of encoded concepts within multilingual models across various languages.  
The first metric \texttt{CALIGN} \textit{(Concept Alignment)} involves measuring alignment by identifying concepts that are semantically equivalent. This provides a nuanced understanding of how concepts in one language align with their counterparts in another, capturing the semantic coherence within the multilingual framework. Our second metric \texttt{COLAP} \textit{(Concept Overlap)} delves into investigating the existence of overlapping cross-lingual latent spaces within the model's representation. This metric aims to highlight multilingual concepts that maintain multiple languages in a close latent space. By probing the shared latent spaces, we gain insights into the intricate relationships between concepts across languages, contributing to a more comprehensive understanding of multilingual model representations, and how they evolve when the model is trained for specific tasks. 
Figure \ref{fig:overall} gives an overview of our approach. %, illustrating how we employ clustering on contextualized embeddings to identify \emph{Encoded Concepts} within multilingual models.
%Figure \ref{fig:overall} provides an overview of our approach, showcasing the stages where we employ clustering on contextualized embeddings to identify \emph{Encoded Concepts} and subsequently apply our proposed metrics to analyze the alignment and overlapping latent spaces within multilingual models. 
In the following sections, we detail each stage of our methodology.

 \subsection{Concept Discovery}
 \label{subsec:concept_discovery}

Our investigation builds upon the work on discovering Latent Concepts in contextualized representations \cite{dalvi2022discovering}. 
At a high level, feature vectors (contextualized representations) are initially generated by performing a forward pass on a neural language model. The representations are then clustered to uncover the encoded concepts of the model. A concept, in this context, can be understood as a collection of words from one or more languages grouped together based on some linguistic relationship, such as lexical, semantic, syntactic, and morphological connections. Figure \ref{fig:overall} illustrates concepts discovered within the latent space of the \texttt{mT5} model, where word representations are organized according to distinct linguistic concepts.

Formally, consider a pre-trained model $\mathbf{M}$ with $L$ layers: $l_1, l_2, \ldots, l_L$. Using a dataset of $S$ sentences totaling $N$ tokens, $\mathcal{D}=[w_1, w_2, \ldots, w_N]$, we generate feature vectors: $\mathcal{D} \xrightarrow{\mathbf{M}_l} \mathbf{z}^l = [\mathbf{z}^l_1, \ldots, \mathbf{z}^l_N]$, where $\mathbf{z}_i^l$ is the contextualized representation for the word $w_i$ from its sentence at layer $l$. A clustering algorithm is then employed in the per-layer feature vector space to discover layer-$l$ encoded concepts.

\subsection{Concept Alignment (\CA{})}
\label{subsec:concept_alignment_method}

Multilingual neural language models are crafted to encode linguistic concepts that bridge equivalent semantic meaning across diverse languages. A key question guiding our exploration is how well this alignment is actually achieved in practice. Specifically, when considering two arbitrary languages, we seek to quantify how well the embeddings of those languages from the same neural model are aligned. We propose an alignment metric, denoted as \CA{} to quantify the correspondence of concepts across different languages within the latent space of multilingual models.
%Once we have obtained a set of encoded concepts in neural language models, we want to align them across languages  \citet{dalvi2022discovering} calibrated representational space in transformer models with different linguistic ontologies to generate their explanations. We formalize their alignment function to discover overlapping latent spaces within multilingual models.
Given a concept $C_s$ (in language \textbf{s}) and a concept $C_t$ (in language \textbf{t}), the number of aligned tokens ${\mathcal{A}_C}_s$ in $C_s$ is:
\vspace{-2mm}
\begin{equation}
    {\mathcal{A}_C}_s = \sum_{w_s \in C_s} \mathbb{I} \left ( \left ( \sum_{w_t \in C_t} \mathcal{T}(w_s, w_t) \right ) > 0 \right )
    \nonumber
\end{equation}

\noindent where function $\mathcal{T}(w_s,w_t) = 1$ if $w_s$ and $w_t$ represent equivalent semantic meaning across the two languages. We simulate $\mathcal{T}(w_s,w_t)$ using a translation dictionary of N-best translations. We consider $C_s$ and $C_t$ to be $\theta_A$-aligned ($\Lambda_{\theta_A}$), if the following constraint is satisfied: 

\begin{equation}
    \Lambda_{\theta_A}(C_s, C_t) = \begin{cases}
    \begin{array}{@{}ll@{}}
    1, & \text{if}\ \frac{{\mathcal{A}_C}_s}{|C_s|} \geq \theta_A  \\
    0, & \text{otherwise}
    \end{array}
    \end{cases}
    \nonumber
    \label{eq:e1}
\end{equation}

\noindent We use a threshold $\theta_A$ to control the extent of alignment i.e. the percentage of words within a cluster required to  satisfy the constraint. The alignment function proves valuable for identifying concepts that exhibit shared semantic meaning in multilingual latent spaces. %or across various latent spaces, such as aligning concepts within the encoder and decoder components in models like \texttt{mT5}.
Finally, \CA{} is the percentage of concepts from language \textbf{s} which are $\theta_A$-aligned to some concept in another language.

\subsection{Concept Overlap (\CO{})} 
\label{subsec:concept_multilinguality_method}

While the alignment metric \CA{} helps to understand whether the model preserves encoded concepts $(C_s, C_t)$ that can be aligned to each other, indicating their shared semantic meaning, it does not explicitly look at overlapping latent spaces across multiple languages in the same model. To investigate these overlapping latent spaces, we introduce another metric denoted as \CO~(\texttt{Concept Overlap}). This metric highlights concepts that encode words from multiple languages in a close latent space. Given $k$ languages, and a set of tokens from language $i$ as $L_i$, We identify a concept as overlapping if it satisfies the following constraint:
\vspace{-0.5mm}
\begin{equation}
    \mathcal{O}_C = \begin{cases}
    \begin{array}{@{}ll@{}}
    1, & \sum_{i=1}^k \mathbb{I}\left(\frac{|{C \cap L_i}| }{|C|} \geq \theta_O \right) \ge 2 \\
    0, & \text{otherwise}
    \end{array}
    \end{cases}
    \nonumber
    \label{eq:e2}
\end{equation}

\noindent where $\theta_O$ defines the minimum threshold of words that must be present in the concept from at least two languages. \CO{} is then computed as the percentage of total concepts that satisfy the above constraint.

% \begin{equation}
% \Lambda_{\theta}(C_x)=
% \frac{\sum_{j=1}^{N} \delta(w_{j} \in \text{L}_i)}{N}  \geq \theta_{\text{L}_i} \quad \forall i = 1, 2 \dots k
%  \nonumber
%   \label{eq:e2}
% \end{equation} 

% \noindent The term $\frac{\sum_{j=1}^{N} \delta(w_{j} \in \text{L}_i)}{N}$ calculates the ratio of words $w_j$ within a concept $C_x$ that belongs to specific language $L_i$. The threshold $\theta_{\text{L}_i}$ signifies that a certain ratio of words belong to language $L_i$. We can probe a concept using $k$ languages. 

Note that, the multilingual concepts may overlap while also being aligned. In such cases, both the \CA{} and \CO{} metrics would identify these concepts. However, there are instances where an overlapping concept may contain related words that are not semantically equivalent, or where the concepts do not overlap but have semantic correspondence. In these scenarios, the two metrics capture distinct aspects.

\section{Experimental Setup}
\label {sec:experiments}

\subsection{Models and Tasks} 
\label{subsec:models_and_tasks}

We experimented with three multilingual transformer architectures namely: mT5,  %\cite{xue-etal-2021-mt5}, 
mBERT, 
%\cite{devlin-etal-2019-bert} 
and XLM-RoBERTa %\cite{xlm-roberta} 
using the base versions (13 layers and 768 dimensions). The former is a state-of-the-art multilingual variant of the T5 (encoder-decoder Transformer) model and the latter two are the cross-lingual variants of the BERT and RoBERTa.  To conduct the analysis, we tuned the mT5 model for the tasks of machine translation (\textit{sequence generation}) using the TED corpus \cite{ansari-etal-2020-findings}. %while 
The mBERT and XLM-R models were tuned for NER-tagging (\textit{sequence labeling}) with the Xtreme dataset \cite{hu2020xtreme} and Sentiment Analysis (\textit{sequence classification}) with the SST-2 dataset \cite{socher-etal-2013-recursive}. We experimented with English, German, French, Spanish, and Arabic.

%XNLI \cite{xue-etal-2021-mt5}. The former is a sequence generation task and the latter is the sequence classification. 

\begin{figure*}
    \centering
      \begin{subfigure}[b]{0.24\textwidth}
         \centering
         \includegraphics[width=\textwidth]{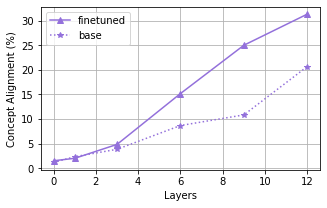}
         \caption{mT5 -- encoder (MT)}
         \label{fig:mt5-encoder-alignment}
     \end{subfigure}
          \begin{subfigure}[b]{0.23\textwidth}
         \centering
         \includegraphics[width=\textwidth]{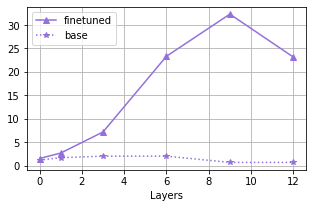}
         \caption{mT5 -- decoder (MT)}
         \label{fig:mt5-decoder-alignment}
     \end{subfigure}
     \begin{subfigure}[b]{0.23\textwidth}
         \centering
         \includegraphics[width=\textwidth]{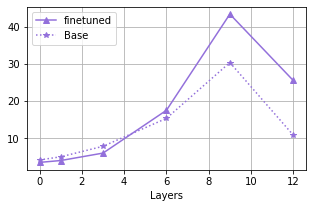}
         \caption{mBERT (NER)}
         \label{fig:mbert-ner-de-en-alignment}
     \end{subfigure}
     % \begin{subfigure}[b]{0.24\textwidth}
     %     \centering
     %     \includegraphics[width=\textwidth]{figures/ner-alignments/mt-data-on-de-en-ner-xlm.png}
     %     \caption{XLM-R (NER)}
     %     \label{fig:xlm-r-ner-de-en-alignment}
     % \end{subfigure}
      \begin{subfigure}[b]{0.23\textwidth}
         \centering
         \includegraphics[width=\textwidth]{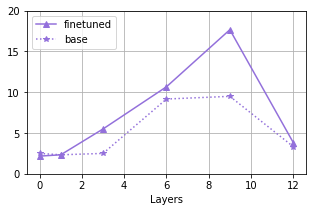}
         \caption{XLM-R (SST-2)}
         \label{fig:de-en-sts-cluster-alignment}
     \end{subfigure}

    \caption{Quantifying Concept Alignment \CA{} (\%) in German–English Concepts: Dotted lines depict base models, while solid lines represent fine-tuned models across different multilingual models.}
    \label{fig:concept-alignment}
\end{figure*}

\begin{figure*}
     \centering
     
     \begin{subfigure}[b]{0.24\textwidth}
         \centering
         \includegraphics[width=\textwidth]{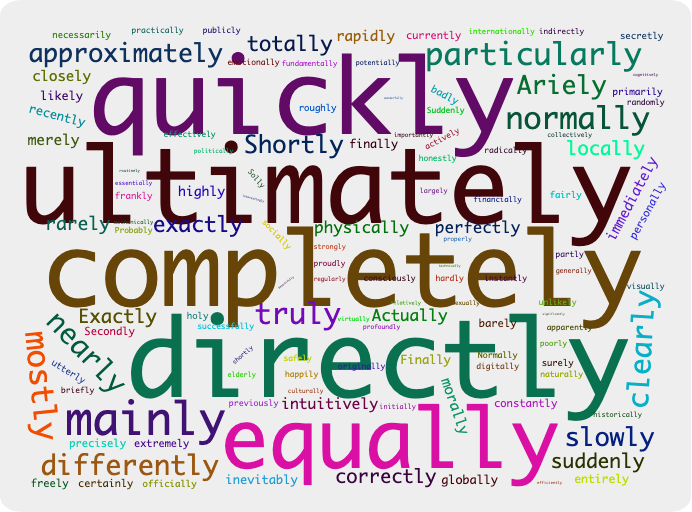}
         \caption{Words ending with "ly"}
         \label{fig:base-en-gr-encoder-0-c13}
     \end{subfigure}
     \begin{subfigure}[b]{0.24\textwidth}
         \centering
         \includegraphics[width=\textwidth]{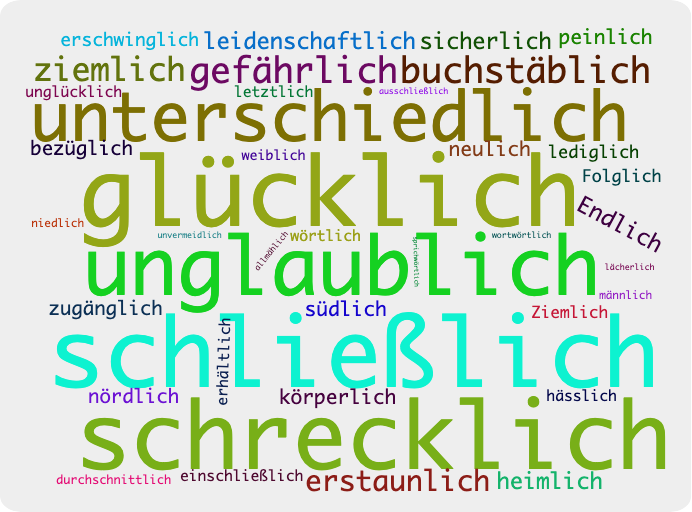}
         \caption{Words ending with "lich"}
         \label{fig:base-gr-en-encoder-0-c3}
     \end{subfigure}
     \begin{subfigure}[b]{0.24\textwidth}
         \centering
         \includegraphics[width=\textwidth]{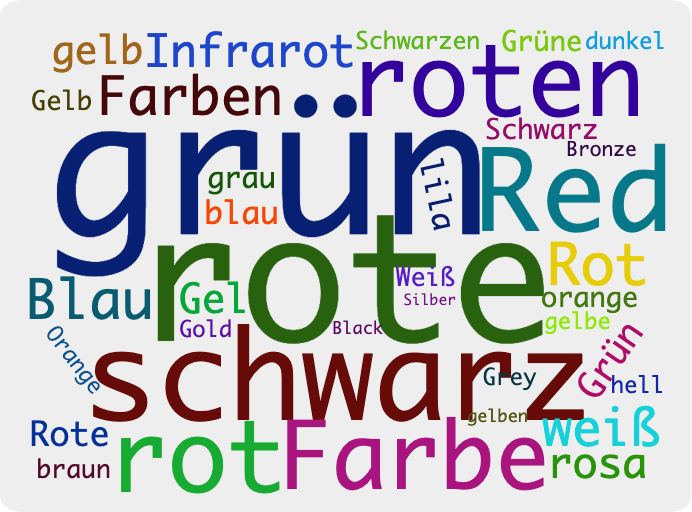}
         \caption{Colors in German}
         \label{fig:de-en-german-encoder-12-c339-0}
     \end{subfigure}
     \begin{subfigure}[b]{0.24\textwidth}
         \centering
         \includegraphics[width=\textwidth]{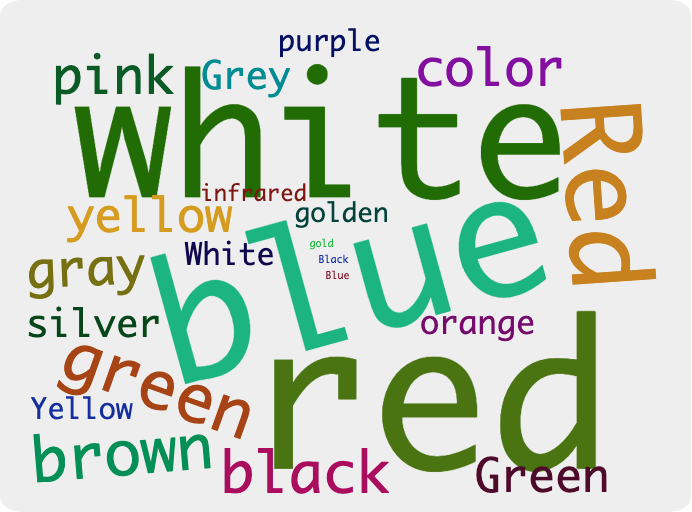}
         \caption{Colors in English}
         \label{fig:de-en-english-encoder-12-c531-0}
     \end{subfigure}     
    \caption{Lower layers capture lexical concepts (a,b), while higher layers focus on semantic concepts (c,d).}
    \label{fig:sample-clusters-mt5}
\end{figure*}

\subsection{Concept Discovery}
\label{subsec:clustering_settings}

We perform a forward pass through the models to generate contextualized feature vectors.\footnote{We use NeuroX toolkit \cite{neurox-aaai19:demo}.} Subsequently, we apply K-means clustering\footnote{\citet{hawasly-etal-2024-scaling} showed K-means to be a viable alternative to the originally proposed agglomerative hierarchical clustering in studying latent spaces.} to the feature vectors, yielding $K$ clusters (also referred to as encoded concepts) for both base and fine-tuned models. We set $K = 600$ and filter out representations that appear at least 10 times, following the settings prescribed by \citet{dalvi2022discovering}.\footnote{
The range of clusters ($K$) between 600 and 1400 yields consistent patterns, as also noted by \citet{sajjad:naacl:2022}. We validated this observation in our initial experiments.} We utilized the parallel data across languages to obtain the encoded concepts. This enables us to accurately compare the representational spaces generated by the same data across multiple languages. It also allows us to estimate the translation dictionary $\mathcal{T}(w_s,w_t)$. We computed word alignments using fast-align \cite{dyer-etal-2013-simple} and then estimated lexical dictionaries using Moses toolkit~\cite{koehn-etal-2007-moses}. The dictionary contains the N-best target translations of a source word. We used GPT-3.5 to annotate the latent concepts for our qualitative analysis \cite{mousi-etal-2023-llms}.

%\fd{Should we clarify that SST-2 uses auto translation data and not TED?}\nd{we used the concatenation to get the dictionary, so I think we can leave it like this} %to filter 10-best\footnote{A word may have many semantic meaning based on different contexts.} $w_t$ translations for each $w_s$. 

% We used the parallel data to obtain clusters. This enables to accurately compare the representational spaces generated by the same data across multiple languages. We do a forward-pass over both base and fine-tuned models to generate contextualized feature vectors of words in the data. We filter out representations based on translation pairs\footnote{We used fast-align \cite{dyer-etal-2013-simple} for alignment.} that have appeared at least 10 times. We then run K-means clustering over the remaining vectors. We do this for every layer independently, obtaining $K$ clusters (a.k.a encoded concepts) for both base and fine-tuned models. We used $K=600$ for our experiments, following the settings prescribed by \citet{dalvi2022discovering}.\footnote{\citet{dalvi2022discovering} found using $K$ ranging between 600 and 1400 to give similar patterns. We also found this to be true in the preliminary experiments we carried out varying the number of clusters.}  For the mT5 model, representations were clustered on both the encoder and decoder sides of the model. 

\subsection{Thresholds}
\label{subsec:threshold}

%\paragraph{Concept Alignment} 
For \CA, we consider $C_s$ (a concept in language \textbf{s}) to be aligned to $C_t$ a concept in language \textbf{t}) if $80\%$ of its types have a semantically equivalent word in $C_t$, i.e. $\theta_A=0.8$. We use 10-best translations\footnote{A word may have many semantic meaning and translations based on different contexts.} %The encoded concepts are based on contextualized embedding where a word has different embeddings depending on the context.} 
of a word $w_s \in C_s$ to define this equivalence. We only consider concepts that have more than 5 word-types. Finally, we also only align concepts $C_s$/$C_t$ if their sizes do not differ by more than 40\%, to avoid aligning very small concepts in one language to a single large concept in another language. We also perform concept discovery independently across languages before aligning the concepts.

For computing \CO{}, we perform concept discovery on multilingual data (mixed sentences from all languages). We deem a concept $C$ to be multilingual or overlapping if all languages being considered form at least 30\% ($\theta_O = 0.3$) of the concept.  %for computing \CO{}.

%Note that 
While the choice of these parameters may seem arbitrary, we experimented with various configurations, such as using a $\theta_A=0.7$--$0.9$ or using $5$--$20$ best translations. The overall patterns of the results remained consistent across different configurations.\footnote{Please refer to Figure \ref{fig:parameter-variations} in Appendix \ref{sec:appendix:thresholds}.} The selected thresholds were based on a qualitative examination of the concepts, allowing for some noise in the concept representations.

%\paragraph{Concept Overlap} assesses the degree of overlap in latent spaces across languages. A concept $C_x$ is classified as multilingual if fewer than 70\% of its words originate from a single language.

%\fd{Change title to Analysis and Results? Feels weird to not have a "Results" section}

\section{Results and Analysis}
\label{sec:analysis}

Cross-lingual representations are deemed to capture unified linguistic concepts across languages which enables them to generalize and to carry out the tasks for low resource languages and zero-shot scenarios. We use latent concept analysis of multilingual models to address the following questions: i) how latent space aligns and overlaps across languages in multilingual model? ii) how is the representation space calibrated as the model is tuned towards different downstream tasks? and iii) what impact does this re-calibration have on the alignment and overlap of concepts representing zero-shot languages? (which were not used for fine-tuning).

% Cross-lingual representations are deemed to capture unified linguistic concepts across languages which enables them to carry out tasks in few and zero-shot scenarios. We
% analyze: i) how latent space aligns across languages in multilingual model? ii) whether the representational calibrates towards better alignment as the model is tuned towards different downstream tasks, iii) does it improve alignment for a zero-shot language? and iv) are there overlapping latent spaces across languages in these models? 

\subsection{Concept Alignment}
\label{subsec:concept_alignment_analysis}

In Figure \ref{fig:concept-alignment}, we illustrate \CA{} across latent spaces in three models: mT5, mBERT, and XLM-R. Dotted lines represent base models, while solid lines denote fine-tuned models. Here the mT5 model is fine-tuned for the task of Machine Translation, mBERT for the NER-tagging and XLM-R for SST-2. The models are jointly trained using German and English samples.  We discover latent concepts in both the base and fine-tuned models for English and German across different layers ($0, 1, 3, 6, 9,$ and $12$),\footnote{We aimed to investigate the embedding layer, as well as the lower, middle, higher middle, and final layers.} plotting the number of aligned concepts (please refer to Section \ref{subsec:concept_alignment_method} for the definition of alignment). %Several insights can be drawn from the obtained results: 
Here are some insights from the results:

\paragraph{Deeper layers in multilingual models reveal increased alignment and preserve semantic concepts, contrasting with language-dependent lexical learning in lower layers.} We observed a significant number of concepts that exhibited alignment within the latent spaces of these models. Notably, up to 42\% of concepts demonstrated alignment within the German-English latent space of the mBERT-NER model. We noted an interesting trend where the number of aligned concepts increased with the depth of the network, reaching its peak in the higher layers of the model. In our qualitative analysis, we found that lower layers of the models are predominantly engaged in learning word morphology, including lexical concepts such as suffixation.\footnote{We also verified this quantitatively. See Figure \ref{fig:mbert-layerwise-trends} in Appendix \ref{sec:appendix:CA} where we count the number of lexical and semantic concepts across different layers of the model.} These aspects are often language-dependent, resulting in a comparatively lower alignment of latent spaces. However, as we go deeper into the network, we uncover more semantic concepts that are preserved across latent spaces in a language-agnostic manner. For example, Figures \ref{fig:base-en-gr-encoder-0-c13} and \ref{fig:base-gr-en-encoder-0-c3} present concepts in lower layers, depicting the learning of lexical concepts like derivational morphology. In contrast, Figures \ref{fig:de-en-german-encoder-12-c339-0} and \ref{fig:de-en-english-encoder-12-c531-0} showcase concepts learned in layer 12, highlighting the higher layers' focus on capturing similar semantic concepts (colors in this case). We found these results to hold consistently across other languages. Please refer to Appendix \ref{sec:appendix:CA} for additional results.

\paragraph{Fine-tuning calibrates the latent space towards higher alignment.} Comparing base models (dotted lines) to fine-tuned models (solid lines) revealed a notable increase in aligned concepts, particularly in higher layers. We posit that base models, trained with a multilingual MLM (mBERT and XLM-R) and ``span-corruption'' (mT5) objectives yield generic linguistic concepts that may not align fully across languages. However, fine-tuning models for specific tasks such as NER or translation leads to calibration of the latent space toward task-specific concepts. This aligns with prior research~\cite{kovaleva-etal-2019-revealing,merchant-etal-2020-happens,durrani-etal-2021-transfer,durrani-etal-2022-latent}, which indicates that higher layers of generic models become optimized for the downstream task.
%the model is trained on.

\begin{table}[!ht]
    \centering
    \setlength{\tabcolsep}{10pt}
    \scalebox{0.90}{
    \begin{tabular}{rrrrr}
    \toprule
         & \multicolumn{1}{r}{test11} & \multicolumn{1}{r}{test12} & \multicolumn{1}{r}{test13} & \multicolumn{1}{l}{test14}  \\
         \midrule
        fr-en (ft) & 49.0& 43.8& 40.8& 42.7\\
        \midrule
        de-en (ft) & 39.9& 36.4& 36.9& 35.5\\
        de-en (zs) & 28.2& 18.9& 23.1& 21.7\\
        \midrule
        es-en (ft) & 43.3& 35.9& 44.7& 44.5\\
        es-en (zs) & 32.0& 26.7& 24.0& 28.2\\
        \midrule
        *-en (bs) & 0.01& 0.02& 0.10& 0.20\\
        %gr-en BS & 0& 0.001& 0& 0.001\\
        %es-en BS& 0& 0 & 0& 0.001 \\ 
        \bottomrule
\end{tabular}}
\caption{BLEU Scores for IWSLT tests: \textbf{ft} = the model fine-tuned for \textbf{fr--en} translation, \textbf{zs} = zero-shot performance of the pair using the \textbf{fr--en} tuned model and \textbf{bs} = the scores when using the base mT5 model.}
\label{tab:french-alignment-all-variants-bleu}
\end{table}

\begin{figure*}
         % \centering
         %     \begin{subfigure}[b]{0.23\textwidth}
         %     \centering
         %     \includegraphics[width=\textwidth]{figures/french-alignments/fr-en-de-en-encoder-decoder-french-model-cluster-alignment.png}
         %     \caption{enc-dec (zero-shot de)}
         %     \label{fig:fr-en-de-en-enc-dec-align-fr-model}
         % \end{subfigure}
         % \begin{subfigure}[b]{0.23\textwidth}
         %     \centering
         %     \includegraphics[width=\textwidth]{figures/french-alignments/fr-en-es-en-encoder-decoder-french-model-cluster-alignment.png}
         %     \caption{enc-dec (zero-shot es)}
         %     \label{fig:mt5-fr-en-es-en-enc-dec-align-fr-model}
         % \end{subfigure}
         % \medskip
         \begin{subfigure}[b]{0.24\textwidth}
             \centering
             \includegraphics[width=\textwidth]{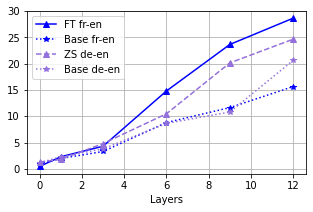}
             \caption{zero-shot de (encoder)}
             \label{fig:mt5-fr-en-de-en-encoder-align-fr-model}
         \end{subfigure}
         \begin{subfigure}[b]{0.24\textwidth}
             \centering
             \includegraphics[width=\textwidth]{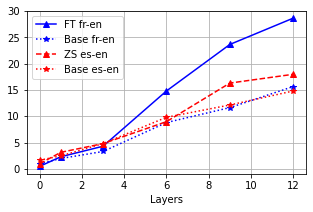}
             \caption{zero-shot es (encoder)}
             \label{fig:mt5-fr-en-es-en-encoder-align-fr-model}
         \end{subfigure}
         %\medskip
         \begin{subfigure}[b]{0.24\textwidth}
             \centering
             \includegraphics[width=\textwidth]{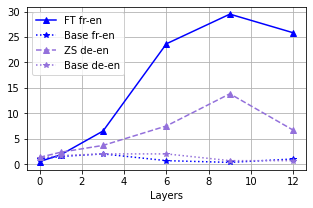}
             \caption{zero-shot de (decoder)}
             \label{fig:mt5-fr-en-de-en-decoder-align-fr-model}
         \end{subfigure}
         \begin{subfigure}[b]{0.24\textwidth}
             \centering
             \includegraphics[width=\textwidth]{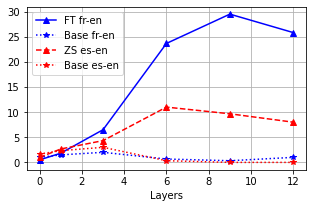}
             \caption{zero-shot es (decoder)}
             \label{fig:mt5-fr-en-es-en-decoder-align-fr-model}
         \end{subfigure}
         
        \caption{Concept Alignment (\%) in \texttt{mT5}. Dotted lines represent base models, solid lines denote fine-tuned French–English MT models, and dashed lines depict zero-shot alignment for German–English and Spanish–English.}
        \label{fig:french-alignment-all-variants}
    \end{figure*}

\begin{figure}
    \centering

      \begin{subfigure}[b]{0.23\textwidth}
         \centering
         \includegraphics[width=\textwidth]{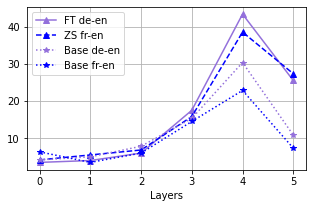}
         \caption{zero-shot fr}
         \label{fig:zero-shot-fr-mbert}
     \end{subfigure}
     \begin{subfigure}[b]{0.23\textwidth}
         \centering
         \includegraphics[width=\textwidth]{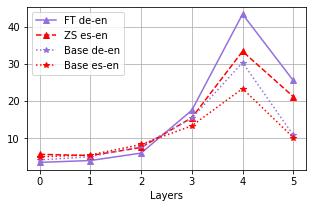}
         \caption{zero-shot es}
         \label{fig:zero-shot-es-mbert}
     \end{subfigure}
    \caption{Concept Alignment (\%) in \texttt{mBERT}. %Dotted lines represent base models, 
    Solid lines: fine-tuned German–English NER model. Dashed lines: zero-shot alignment for French and Spanish.}
    \label{fig:zero-shot-mbert}
\end{figure}

\begin{table}[!ht]

\centering
\begin{tabular}{lrr|rr}
\toprule
language      & en & de & fr  & es  \\
      \midrule
      & \multicolumn{2}{c}{fine-tuned} & \multicolumn{2}{c}{zero-shot} \\
      \midrule
mbert (NER) & 84.6 & 89.7 & 77.9 &  68.0\\ 
\midrule
mbert (base) & 3.0 & 4.9 & 5.8 &  4.1\\ 
\bottomrule
\end{tabular}
\caption{F1 scores for \texttt{mBERT--NER} (German,English). French and Spanish represent the zero-shot scenario.}
%\textbf{ft} refers to the model fine-tuned for translation  using German-English fine-tuned model, \textbf{zs} refers to the zero-shot performance and \text{bs} refers to scores using base model.}
\label{tab:german-english-alignment-F1}
\end{table}

We also observed that \textbf{task-specific calibration of the latent space facilitates zero-shot capabilities}. To substantiate this claim quantitatively, we extract latent concepts for zero-shot languages (not used during fine-tuning) and evaluate their alignment. Figure \ref{fig:french-alignment-all-variants} illustrates concept alignment in the \texttt{mT5} model tuned towards the task of French–English translation. We extract concepts for French, English, German, and Spanish from these models on both the encoder and decoder sides, with the latter two representing zero-shot scenarios. The dashed lines indicate concept alignment for German and Spanish within these models. Notably, we observe a substantial increase in the percentage of aligned concepts, despite the model not being fine-tuned for German– or Spanish–English translation. This suggests that the presence of language-agnostic concepts within the latent space of these models facilitates performance in zero- and few-shot scenarios. Our findings correlate with the BLEU scores~\cite{post-2018-call}, as shown in Table \ref{tab:french-alignment-all-variants-bleu}. Note that while the zero-shot German and Spanish translations show significantly lower performance compared to their respective models after fine-tuning, the model still performs reasonably well considering it was never explicitly trained for German- and Spanish-English translation tasks.  We consistently observed these trends across various language settings in the \texttt{mT5} model\footnote{Please see Figures \ref{fig:german-model-alignment}--\ref{fig:arabic-model-alignment} in the Appendix \ref{sec:appendix:CA} for results.} and in the \texttt{mBERT} model fine-tuned for the NER  task for German and English. Notably the alignment improved in zero-shot French and Spanish languages (compare dashed lines (zs) to dotted lines (base) in Figure \ref{fig:zero-shot-mbert}). Again, these findings correlate with the F1 Scores (see Table \ref{tab:german-english-alignment-F1}). We see similar results for \texttt{XLM-R} model fine-tuned for the SST2 task as well.

\begin{figure*}
     \centering
     \begin{subfigure}[b]{0.24\textwidth}
         \centering
         \includegraphics[width=\textwidth]{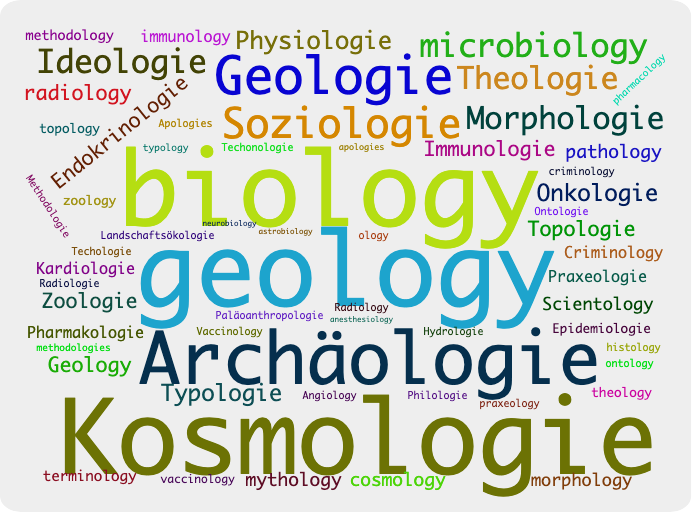}
         \caption{Shared infix ``olog'' (de, en) }
         \label{fig:de-en-combined-cluster-encoder-0-de-en-model-c443}
     \end{subfigure}
     \begin{subfigure}[b]{0.24\textwidth}
         \centering
         \includegraphics[width=\textwidth]{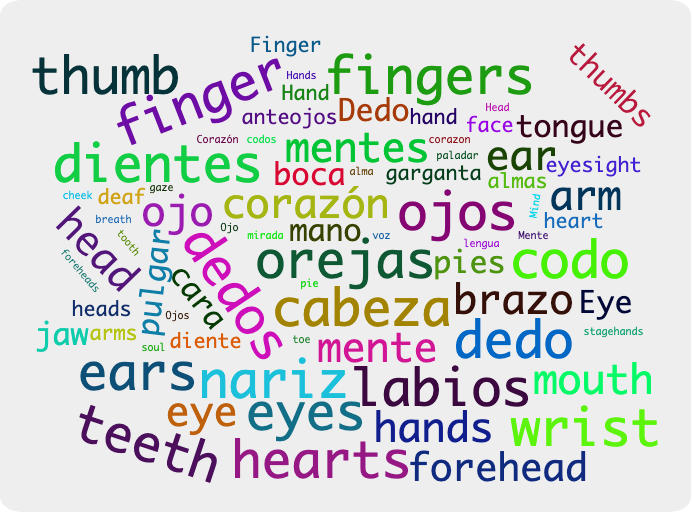}
         \caption{Anatomy \& Senses (es, en)}
         \label{fig:es-en-combined-cluster-c308-encoder-12}
     \end{subfigure}
     \begin{subfigure}[b]{0.24\textwidth}
         \centering
         \includegraphics[width=\textwidth]{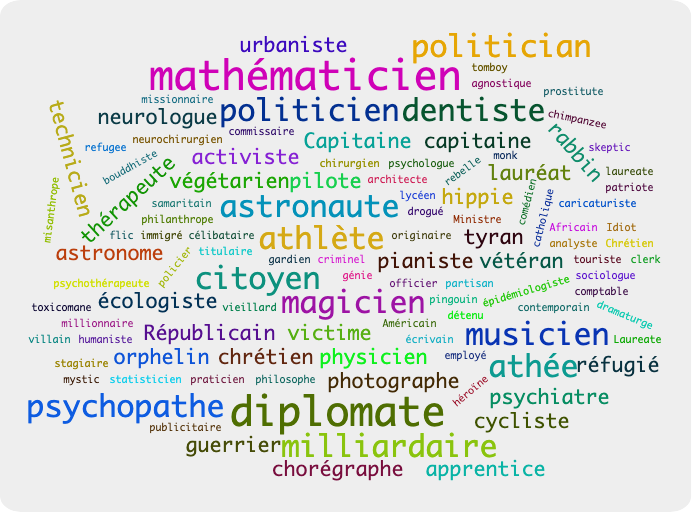}
         \caption{Occupations (fr, en)}
         \label{fig:fr-en-combined-french-english-12-c10}
     \end{subfigure}
     \begin{subfigure}[b]{0.24\textwidth}
         \centering
         \includegraphics[width=\textwidth]{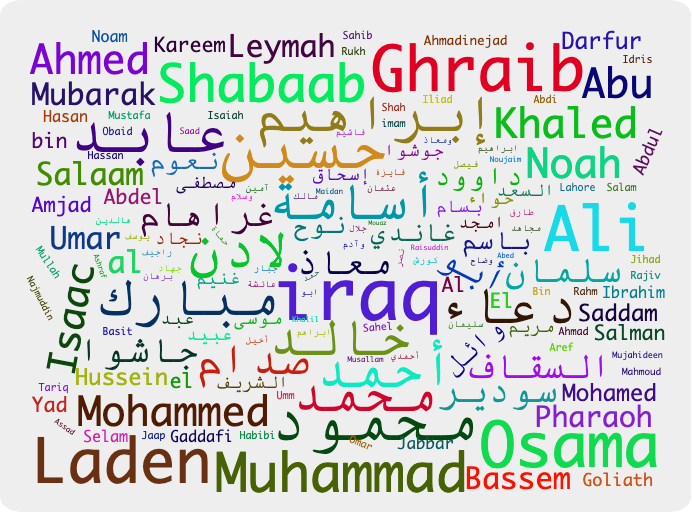}
         \caption{Names (ar, en)}
         \label{fig:ar-en-combined-cluster-encoder-ar-en-model-c12}
     \end{subfigure}
    \caption{Sample Overlapping Concepts in the \texttt{mT5} model.}
    \label{fig:sample-multilingual-clusters}
\end{figure*}

\begin{figure*}
    \centering

      \begin{subfigure}[b]{0.24\textwidth}
         \centering
         \includegraphics[width=\textwidth]{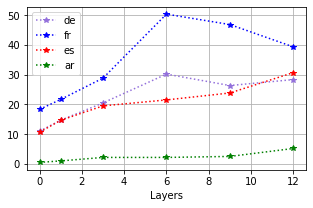}
         \caption{mT5 encoder -- Base}
         \label{fig:mt5-base-multilinguality}
     \end{subfigure}
          \begin{subfigure}[b]{0.24\textwidth}
         \centering
         \includegraphics[width=\textwidth]{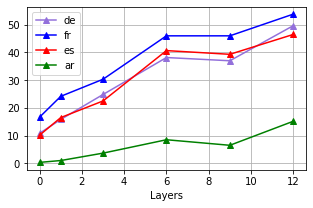}
         \caption{mT5 encoder -- MT-tuned}
         \label{fig:mt5-ft-multilinguality-encoder}
     \end{subfigure}
     \begin{subfigure}[b]{0.24\textwidth}
         \centering
         \includegraphics[width=\textwidth]{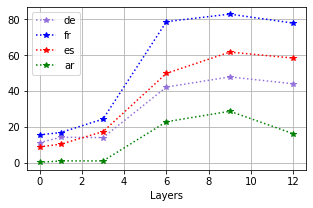}
         \caption{mT5 decoder -- Base}
         \label{fig:mt5-decoder-base-multilinguality}
     \end{subfigure}
     \begin{subfigure}[b]{0.24\textwidth}
         \centering
         \includegraphics[width=\textwidth]{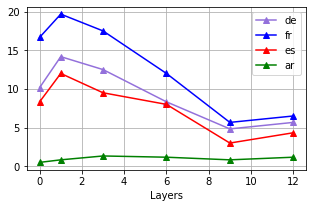}
         \caption{mT5 decoder -- MT-tuned}
         \label{fig:mt5-decoder-ft-multilinguality}
     \end{subfigure}
    \caption{Quantifying Overlapping Concepts in different languages in \texttt{mT5} encoder and decoder}
    \label{fig:concept-multilinguality}
\end{figure*}

\paragraph{Divergent patterns emerge in the encoder and decoder latent spaces.} %Revisiting our 
Comparing our findings in \texttt{mT5}, as depicted in Figure \ref{fig:concept-alignment},
%\ref{fig:french-alignment-all-variants}, 
we noted disparities in alignment between the encoder and decoder spaces: i) while the base model demonstrated reasonable alignment on the encoder side (up to $20\%$), indicated by the dotted line in Figure \ref{fig:mt5-encoder-alignment},
%\ref{fig:mt5-fr-en-de-en-encoder-align-fr-model} and \ref{fig:mt5-fr-en-es-en-encoder-align-fr-model}), 
alignment on the decoder side was minimal ($<3\%$), as shown in Figure \ref{fig:mt5-decoder-alignment}.
%\ref{fig:mt5-fr-en-de-en-decoder-align-fr-model} and \ref{fig:mt5-fr-en-es-en-decoder-align-fr-model}). 
Decoders in transformer models are responsible for generating target language sequences based on the encoded input. We speculate that since its primary focus is on generating fluent and accurate translations, it may prioritize language-specific nuances and idiosyncrasies, leading to lesser aligned concepts across languages. This also explains a decrease in alignment observed in the final layers of the fine-tuned decoder.

We see a similar dip in the last layer of encoder-only \texttt{mBERT} and \texttt{XLM-R} models for the NER and SST-2 tasks (refer to Figures \ref{fig:mbert-ner-de-en-alignment} and \ref{fig:de-en-sts-cluster-alignment}), which again can be attributed to the layers adapting to the task at hand instead of maintaining semantic alignment across languages.

\subsection{Concept Overlap}
\label{subsec:concept_multilinguality_analysis}

\CA{} serves to assess whether the model captures concepts that exhibit alignment across languages, signifying shared semantic space. Our \CO{} metric delves into this aspect further by exploring the presence of overlapping latent spaces within the model's representation. This sheds light on how the model effectively maintains multiple languages within a shared latent space. We demonstrate a selection of concepts demonstrating multilinguality. Figure \ref{fig:de-en-combined-cluster-encoder-0-de-en-model-c443} illustrates a concept at the lower layer where German and English intersect, sharing the common infix ``olog''.  Various multilingual semantic concepts, including Anatomy \& senses, Occupations and Names are depicted across different languages. Note that while \CA{} can identify the concept in Figure \ref{fig:es-en-combined-cluster-c308-encoder-12} because its constituent words are semantically equivalent, the cross-lingual words in Figure \ref{fig:de-en-combined-cluster-encoder-0-de-en-model-c443} are grouped based on lexical, rather than semantic similarity. \CO{} helps us detect such concepts. 

\begin{figure}
    \centering
      \begin{subfigure}[b]{0.23\textwidth}
         \centering
         \includegraphics[width=\textwidth]{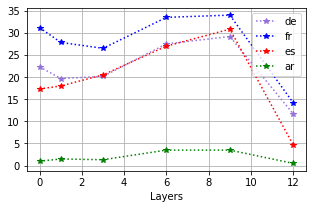}
         \caption{mBERT Base}
         \label{fig:multilinguality-mbert-base}
     \end{subfigure}
     \begin{subfigure}[b]{0.23\textwidth}
         \centering
         \includegraphics[width=\textwidth]{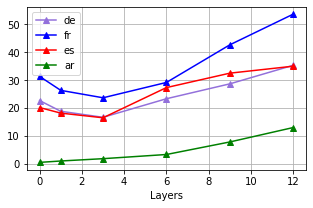}
         \caption{mBERT (NER)}
         \label{fig:mutlilinguality-mbert-finetuned}
     \end{subfigure}

    \caption{Quantifying Concept Overlap in \texttt{mBERT}}
    \label{fig:mbert-multilinguality}
    \vspace{-1mm}
\end{figure}

In Figures \ref{fig:concept-multilinguality}--\ref{fig:mbert-multilinguality} we quantify overlap across latent spaces in various layers of \texttt{mT5} and \texttt{mBERT} models. %Dotted lines represent the base models, whereas solid lines depict the models post fine-tuning for machine translation and NER tasks respectively.
We note a significant number of concepts across layers with a high \CO{} score in both \texttt{mT5} and \texttt{mBERT}. The overlap typically peaks around 50\% across most settings (refer to Figures \ref{fig:concept-multilinguality} and \ref{fig:mbert-multilinguality}). %Notably, in the case of French-English overlap, up to 80\% of the concepts satisfied the multilinguality constraint %established in Section \ref{subsec:concept_multilinguality_method} 
%(Figure \ref{fig:mt5-decoder-base-multilinguality}).
We draw the following insights from these results:

%\paragraph{A significant portion of the concepts demonstrate multilinguality.} %We observe that a high number of concepts across layers exhibit a high \CO{} score in both \texttt{mT5} and \texttt{mBERT}. The overlap peaks around 50\% for most settings (See Figures \ref{fig:concept-multilinguality} and \ref{fig:mbert-multilinguality}). Remarkably, up to 80\% (Figure \ref{fig:mt5-decoder-base-multilinguality}) of the concepts satisfied the constraint for multilinguality that we established in Section \ref{subsec:concept_multilinguality_method} in the case of French-English overlap.

\paragraph{Closely related languages demonstrate higher overlap in latent space.} We observe a spectrum of overlap across languages, with the highest degree found in French (peaking around 80\%) and the lowest in Arabic (peaking around 25\%) -- please see  Figure \ref{fig:mt5-decoder-base-multilinguality}. English and French showcase substantial overlap in their latent spaces, attributed to their shared linguistic roots within the Indo-European language family. Specifically, French stems from the Romance branch, while English belongs to the Germanic branch. This common linguistic heritage manifests in similarities in vocabulary and syntactic structures between the two languages. In contrast, Arabic exhibits notable differences in orthography and morphology when compared to English. As a Semitic language, Arabic presents unique linguistic characteristics absent in Indo-European languages like English and French. Its script diverges significantly from the Latin script, while its intricate root-and-pattern morphology stands in stark contrast to English morphology. These linguistic disparities contribute to a reduced degree of overlap in the latent space between English and Arabic compared to English and French.

\paragraph{The complexity of optimization function affects the extent of overlap in latent spaces} While German and English share a closer linguistic relationship, and belong to the Germanic language branch within the Indo-European family, it exhibits a lesser overlap compared to French. The extent of their overlap in the latent space may be influenced by the differences in syntax, such as word order and grammatical structure, despite their linguistic closeness. Note that the base \texttt{mT5} model employs span correction as its optimization function, which may primarily requires a focus on short-range dependencies. In contrast, the translation task requires the handling of long-range syntactic dependencies. Consequently, as the models are fine-tuned for machine translation tasks, we also observe a higher overlap for German in latent spaces of the fine-tuned models (See Figures \ref{fig:mt5-ft-multilinguality-encoder} and \ref{fig:mt5-decoder-ft-multilinguality}). We even notice an increase in overlapping concepts for Arabic-English in the higher layers post fine-tuning. A comprehensive investigation, however, is required to examine this further, and we defer this exploration to future studies.

% \paragraph {While the majority of a model's concepts maintain multilingual characteristics, the final layers after fine-tuning tend to primarily retain language-specific concepts.} While we see significant overlap across languages in general, the number of overlapping concepts diminished to less than 20\% (See Figure \ref{fig:mt5-decoder-ft-multilinguality}) as the model was fine-tuned towards the task of machine translation, dropping even below 5\% in the final layers.  This demonstrates that while the majority of the model's concepts maintain multilingual characteristics, the final layers in the decoder primarily retain language-specific concepts. However, it is noteworthy that these concepts may be semantically equivalent and satisfy \CA{}, as illustrated in Section \ref{subsec:concept_alignment_analysis} (recall Figure \ref{fig:french-alignment-all-variants}).

\paragraph {While most of the concepts in a model exhibit multilingual traits, the later layers, post fine-tuning, tend to preserve predominantly language-specific characteristics.} Although substantial overlap is evident across languages in general, the proportion of concepts that overlap diminishes to less than 20\% (See Figure \ref{fig:mt5-decoder-ft-multilinguality}) as the model undergoes fine-tuning for machine translation, dropping further below 5\% in the final layers. This underscores that while the bulk of a model's concepts maintain multilingual attributes, the final layers within the decoder predominantly preserve language-specific traits. It's worth noting, however, that these concepts may still be semantically equivalent and satisfy \CA{}, as demonstrated in Section \ref{subsec:concept_alignment_analysis} (refer to Figure \ref{fig:french-alignment-all-variants}).

% We do not notice any such drop in the \texttt{mBERT} NER model (Figure \ref{fig:mutlilinguality-mbert-finetuned}), where the consistently high overlap can be attributed to output class specific concepts (such as location concepts), where semantic alignment may not be as important as just bringing locations from various languages close together for prediction.

We do not observe a similar drop in the \texttt{mBERT} NER model (Figure \ref{fig:mutlilinguality-mbert-finetuned}), where the consistently high overlap can be ascribed to concepts specific to output classes (e.g. \texttt{location} concepts), where semantic alignment may be less crucial than merely grouping locations from different languages closely together for prediction.

%\paragraph{Compare mt5 with mbert and xlm}

\section{Related Work}
\label{sec:related_work}

Numerous studies have explored the domain of multilingual embedding, investigating how deep neural language models encode knowledge across various languages without explicit supervision.  \citet{pires-etal-2019-multilingual} demonstrated mBERT's ability to learn multilingual representations, enabling cross-lingual transfer even for languages with different scripts, provided they share topological similarities. \citet{cao2020multilingual} employ a contextual word retrieval task where the model is tasked with finding corresponding words and sentences across parallel corpora.  \citet{dufter-schutze-2020-identifying} identified critical architectural and linguistic properties for multilinguality, emphasizing the necessity of common positional embeddings, shared special tokens, and a restricted parameter space. \citet{papadimitriou-etal-2021-deep} investigated higher-order grammatical feature representation across languages using probing classifiers trained on mBERT embeddings. Their successful zero-shot cross-lingual transfer demonstrated parallel representation of grammatical features. \citet{wen-yi-mimno-2023-hyperpolyglot} conducted analysis on the embedding layer of mT5 and XLM-R, uncovering the diverse language encoding patterns within these models and highlighting the semantic encoding across languages. \citet{xu-etal-2023-structural} investigated the conceptual correspondence between structural concepts in linguistically diverse languages, emphasizing the correlation between conceptual alignment and cross-lingual transfer. They proposed a meta-learning approach to align these linguistic spaces, enabling zero-shot and few-shot generalization. 
% Our approach diverges from prior research methodologies. We use an unsupervised approach to unveil multilingual concepts learned within the latent space of these models. We illustrate the alignment and overlap within these spaces and track their recalibration as the models undergo fine-tuning for downstream tasks. Our findings suggest that this calibration of latent space enhances the model's performance in zero-shot scenarios.

Our approach diverges from prior research methodologies by using an unsupervised approach to unveil multilingual concepts learned within the latent space of these models. We identify latent concepts across different languages and assess alignment across these concepts using our proposed metrics CALIGN and COLAP. Unlike previous approaches that focus on individual words and local alignment, our multilingual concept analysis provides insight into how different linguistic concepts align and overlap across multilingual spaces. We illustrate the alignment and overlap within these spaces and track their recalibration as the models undergo fine-tuning for downstream tasks. While prior research often examines if individual words have aligned counterparts in target languages, our work extends this by enforcing whether the latent spaces themselves are similarly constructed. This means that the neighbors of a word in one language correspond to neighbors of the target word in another language, introducing a stronger evidence of multilinguality at a fundamental level. Our findings suggest that this calibration of latent space enhances the model's performance in zero-shot scenarios, presenting a distinct analysis and revealing results that significantly differ from previous research.

\section{Conclusion} 
\label{sec:conclusion}

The emergence of multilingual contextualized embeddings has sparked interest in understanding their mechanisms. We introduce two metrics, Concept Alignment (\texttt{CALIGN}) and Concept Overlap (\texttt{COLAP}), to quantify \textit{alignment} and \textit{overlap} within multilingual models. Our analysis reveals: i) deeper layers exhibit increased alignment due to presence of semantic concepts, ii) fine-tuning enhances alignment across cross-lingual concepts, facilitating zero-shot capabilities, iii) divergent patterns in encoder and decoder spaces and higher overlaps between closely related languages are observed. Our insights shed light on the dynamics of multilingual embeddings and lay the groundwork for a more comprehensive understanding of
multilingual NLP models.

\section{Limitations} 

We list below limitations of our work:

\begin{itemize}

    \item  While our approach effectively analyzes how multilingual models encode concepts across languages within their learned representations, it does not shed light on how these concepts are utilized by the model during prediction. Our results demonstrate a correlation between our metrics and the model's performance (as measured by BLEU and F1 scores) in the zero-shot scenarios. However, establishing causation from this correlation is not straightforward. In future research, we aim to integrate our method with ablation and knowledge attribution techniques to establish a direct connection between the encoded concepts and their impact on prediction.

    \item Due to the high dimensionality of contextual representations, only a restricted amount of data can be clustered to extract latent concepts. This limitation affects the goal of concept discovery, providing only a partial view of the spectrum of concepts that could be learned within the model. Our experiments were constrained by time and memory limitations. It is possible that with large-scale experimentation, we could uncover many other intriguing concepts. Additionally, time and memory constraints prevent us from exploring other clustering algorithms that may yield a superior hierarchy of concepts but are computationally infeasible.

\end{itemize}

% \textcolor{red}{in progress - Basel}

% we list below the limitations of our work 

% \begin{itemize}
%     \item Running clustering algorithms on hidden representations requires a large amount of compute whose availability is contingent upon circumstances. In future work, we aim to make the process of activation-extraction and clustering more compute efficient.
%     \item  We only described what could be the reason behind the zero-shot ability of models without connecting the knowledge analyzed to how the model is making predictions. In future work we aim to extend our work to analyze how the model utilizes the cross-lingual knowledge in making predictions.
% \end{itemize}
\label{sec:limitations}

\bibliography{anthology,custom}
\bibliographystyle{acl_natbib}

%\newpage
\section*{Appendix}
\label{sec:appendix}
\appendix
\section{Latent Concepts}
\label{sec:LC}

\begin{figure*}
     \centering
     % \begin{subfigure}{0.3\textwidth}
     %     \centering
     %     \includegraphics[width=\textwidth]{figures/cluster_plots/de_c3_mt5_encoder_0.png}
     %     \caption{Words ending with ``lich''}
     %     \label{fig:words_ending_with_lich}
     % \end{subfigure}
      \begin{subfigure}{0.3\textwidth}
         \centering
         \includegraphics[width=\textwidth]{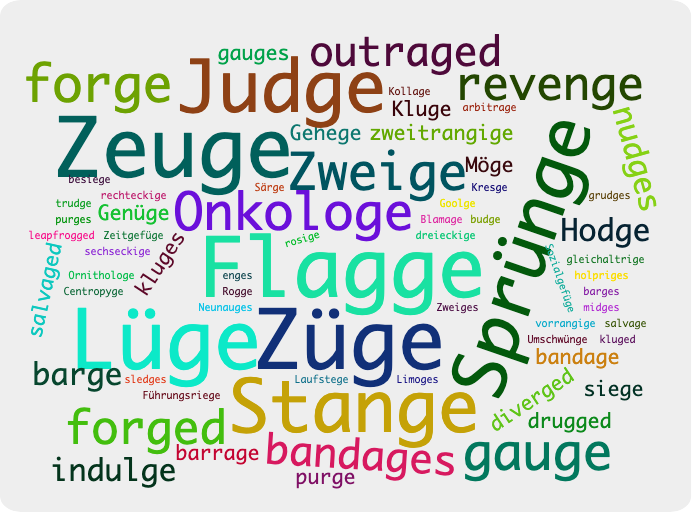}
         \caption{``ge'' infix}
         \label{fig:words_ending_with_lich}
     \end{subfigure}
     \hfill
     % \begin{subfigure}{0.3\textwidth}
     %     \centering
     %     \includegraphics[width=\textwidth]{figures/cluster_plots/de_c5_mt5_encoder_12.png}
     %     \caption{Sciences}
     %     \label{fig:sciences}
     % \end{subfigure}
     \begin{subfigure}{0.3\textwidth}
         \centering
         \includegraphics[width=\textwidth]{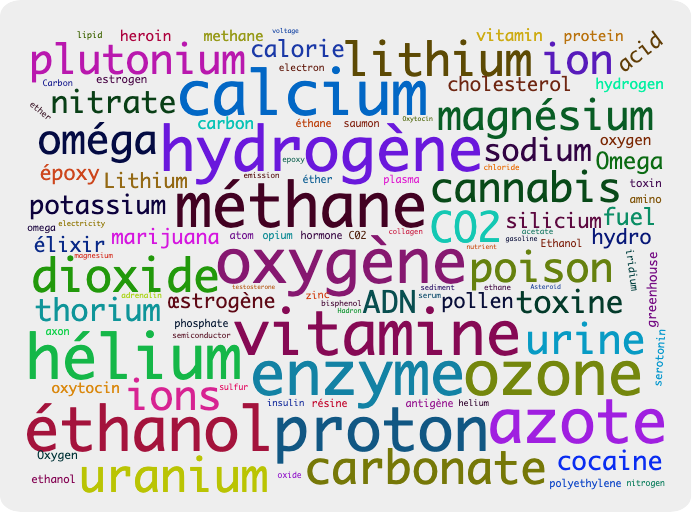}
         \caption{Chemical Elements}
         \label{fig:sciences}
     \end{subfigure}
     \hfill
     % \begin{subfigure}{0.3\textwidth}
     %     \centering
     %     \includegraphics[width=\textwidth]{figures/cluster_plots/en_c4_mt5_base_encoder.png}
     %     \caption{Military related terms}
     %     \label{fig:military_related_terms}
     % \end{subfigure}
     \begin{subfigure}{0.3\textwidth}
         \centering
         \includegraphics[width=\textwidth]{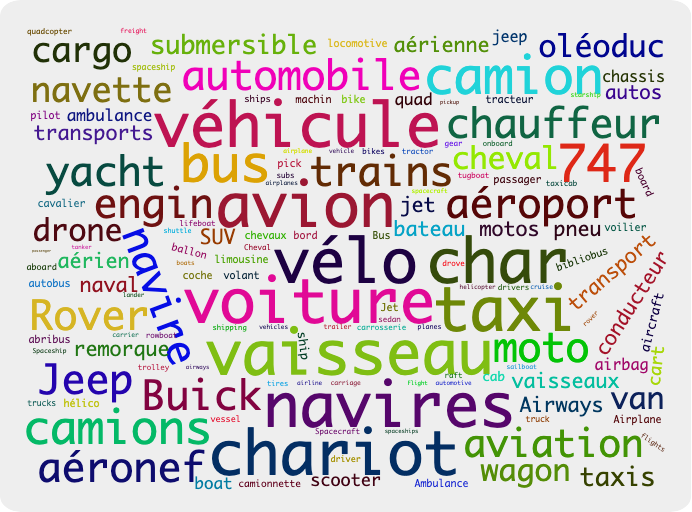}
         \caption{Modes of Transportation}
         \label{fig:mtm-fr-en-12-encoder-c451_}
     \end{subfigure}
     %  \begin{subfigure}{0.3\textwidth}
     %     \centering
     %     \includegraphics[width=\textwidth]{figures/cluster_plots/en_c5_mt5_base_encoder_0.png}
     %     \caption{Words ending with ``ly''}
     %     \label{fig:en_words_ending_with_ly}
     % \end{subfigure}
      \begin{subfigure}{0.3\textwidth}
         \centering
         \includegraphics[width=\textwidth]{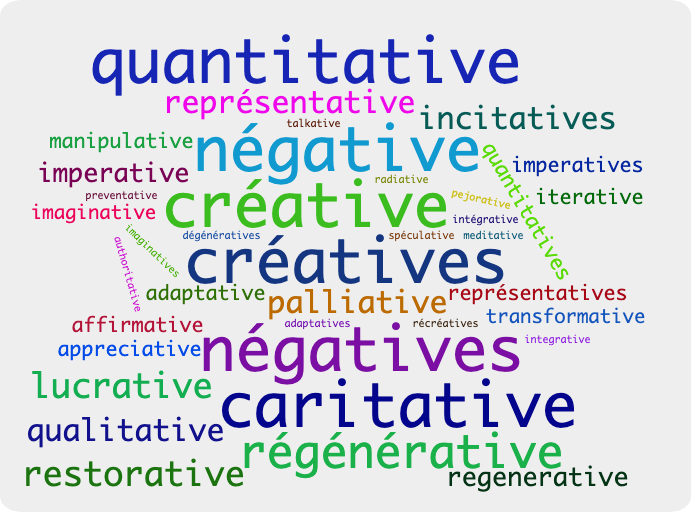}
         \caption{Words ending with ``tive''}
         \label{fig:en_words_ending_with_ly}
     \end{subfigure}
     \hfill
     %  \begin{subfigure}{0.3\textwidth}
     %     \centering
     %     \includegraphics[width=\textwidth]{figures/cluster_plots/en_c22_mt5_base_encoder_12.png}
     %     \caption{Software related terms}
     %     \label{fig:software_related_terms}
     % \end{subfigure}
     \begin{subfigure}{0.3\textwidth}
         \centering
         \includegraphics[width=\textwidth]{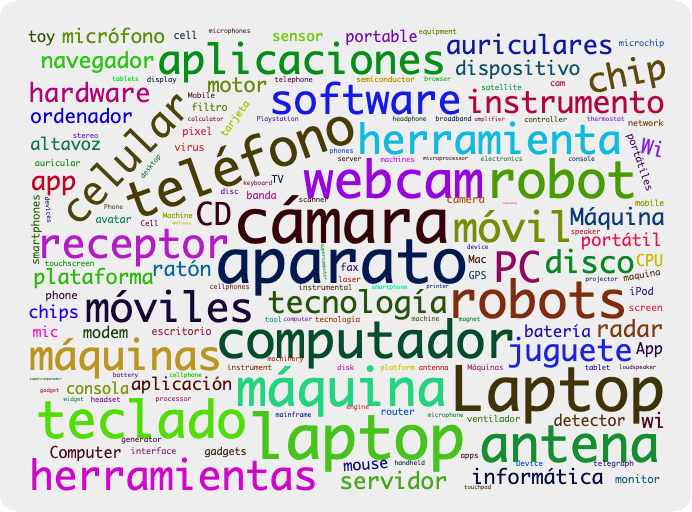}
         \caption{Technological Devices and tools}
         \label{fig:mtm-es-en-12-encoder-c211_}
     \end{subfigure}
     \hfill
     %  \begin{subfigure}{0.3\textwidth}
     %     \centering
     %     \includegraphics[width=\textwidth]{figures/cluster_plots/es_c22_mt5_base_encoder_12.png}
     %     \caption{Spanish Nouns}
     %     \label{fig:spanish_nouns}
     % \end{subfigure}
     \begin{subfigure}{0.3\textwidth}
         \centering
         \includegraphics[width=\textwidth]{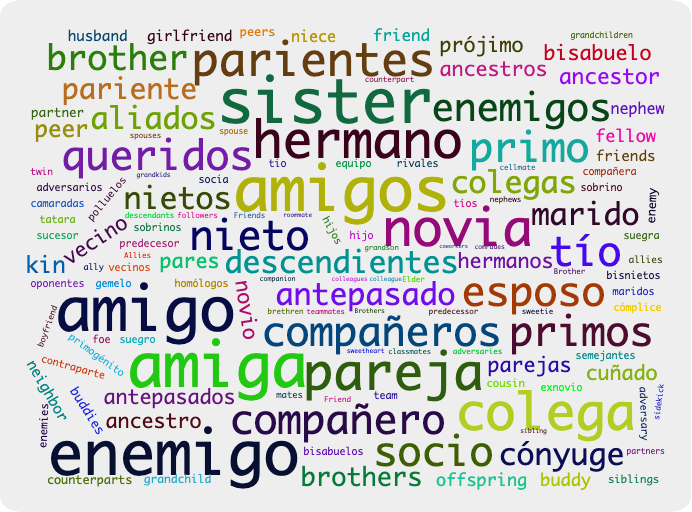}
         \caption{Family and Relationships}
         \label{fig:mtm-es-en-12-encoder-c28_}
     \end{subfigure}
     % \begin{subfigure}{0.3\textwidth}
     %     \centering
     %     \includegraphics[width=\textwidth]{figures/cluster_plots/es_c28_mt5_base_encoder_12.png}
     %     \caption{Spanish Adverbs}
     %     \label{fig:spanish_adverbs}
     % \end{subfigure}
      \begin{subfigure}{0.3\textwidth}
         \centering
         \includegraphics[width=\textwidth]{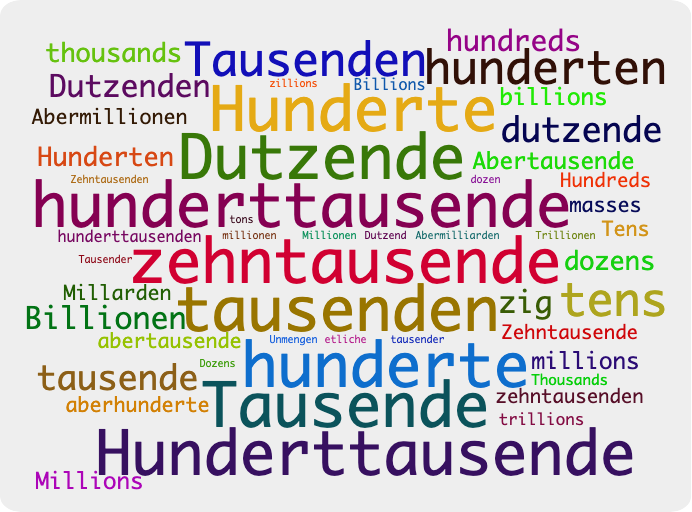}
         \caption{Qualities and Numbers}
         \label{fig:um-1}
     \end{subfigure}
     \hfill 
    % \begin{subfigure}{0.3\textwidth}
    %      \centering
    %      \includegraphics[width=\textwidth]{figures/cluster_plots/fr_c28_mt5_encoder_12.png}
    %      \caption{Middle East Related terms}
    %      \label{fig:fr_middle_east}
    %  \end{subfigure}
    \begin{subfigure}{0.3\textwidth}
         \centering
         \includegraphics[width=\textwidth]{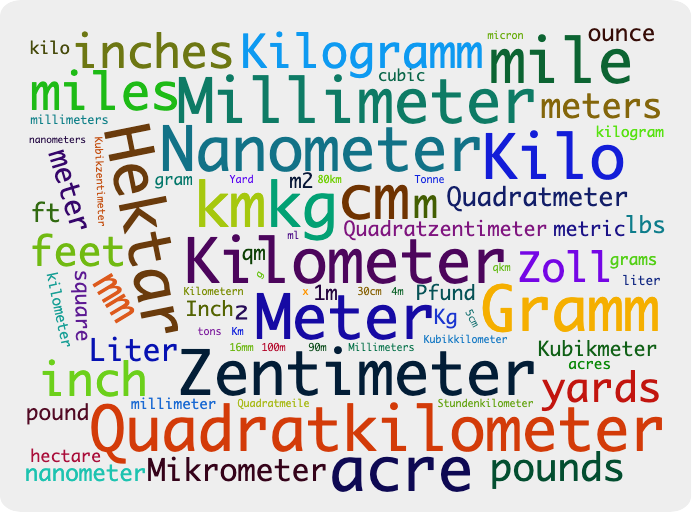}
         \caption{Units of Measurement}
         \label{fig:um-2}
     \end{subfigure}
     \hfill 
    % \begin{subfigure}{0.3\textwidth}
    %      \centering
    %      \includegraphics[width=\textwidth]{figures/cluster_plots/fr_c66_mt5_encoder_12.png}
    %      \caption{Identities}
    %      \label{fig:fr_people_identities}
    %  \end{subfigure}
 \begin{subfigure}{0.3\textwidth}
         \centering
         \includegraphics[width=\textwidth]{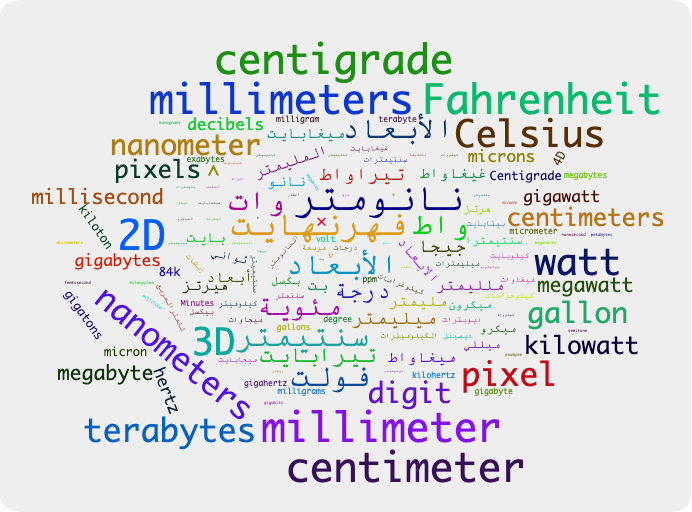}
         \caption{Units of Measurment}
         \label{fig:um-3}
     \end{subfigure}     
        \caption{Sample Concepts learned in the mT5 model}
        \label{fig:mt5_concepts}
\end{figure*}

In Figure \ref{fig:mt5_concepts}, we present a selection of concepts learned within the latent space of the multilingual mT5 model. These figures showcase a diverse array of encoded concepts, encompassing lexical concepts (e.g., Figures \ref{fig:words_ending_with_lich} and \ref{fig:en_words_ending_with_ly}, which depict German and English words with affixes ``ge'' and ``able'' respectively), semantic concepts (e.g., Figures \ref{fig:um-1} -- \ref{fig:um-3}, highlighting quantities, numbers and units of measurement in different languages), and more intricate semantic concepts illustrating fine-grained taxonomies (e.g., Figures \ref{fig:sciences}, capturing various scientific disciplines).

\section{Concept Alignment}
\label{sec:appendix:CA}

In Section \ref{subsec:concept_alignment_analysis}
we discussed several results. Here we demonstrate that our findings generalize to other languages.

\begin{figure}
    \centering
    \includegraphics[width=0.9\linewidth]{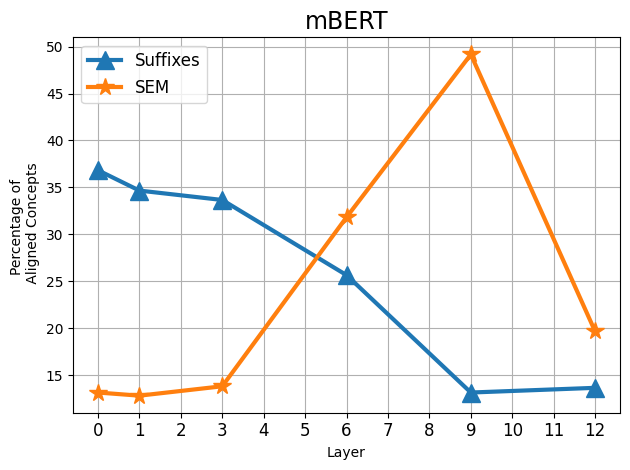}
    \caption{Layer-wise alignment of clusters to lexical and semantic properties in mBERT}
    \label{fig:mbert-layerwise-trends}
\end{figure}

\begin{figure*}
     \centering
     \begin{subfigure}[b]{0.24\textwidth}
         \centering
         \includegraphics[width=\textwidth]{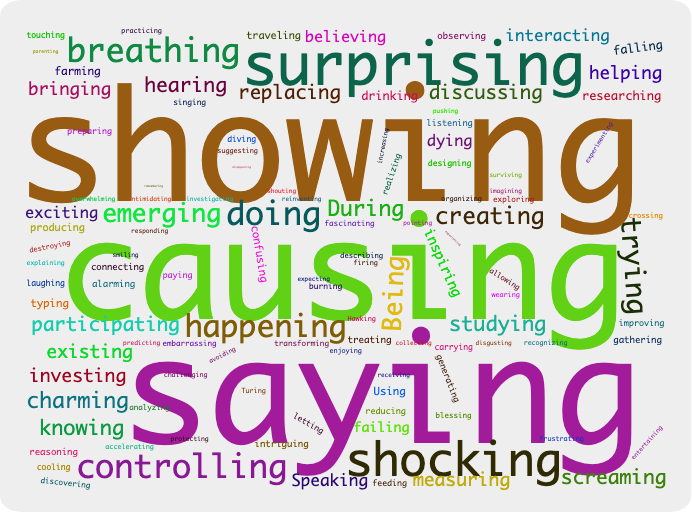}
         \caption{Words ending with ``ing''}
         \label{fig:base-es-en-en-encoder-c36-0}
     \end{subfigure}
     \begin{subfigure}[b]{0.24\textwidth}
         \centering
         \includegraphics[width=\textwidth]{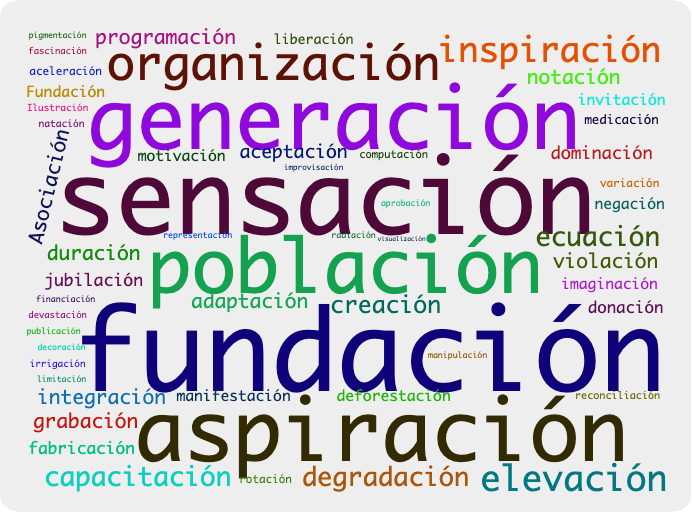}
         \caption{words ending with ``ión''}
         \label{fig:base-es-en-en-encoder-c2-0}
     \end{subfigure}
     \begin{subfigure}[b]{0.24\textwidth}
         \centering
         \includegraphics[width=\textwidth]{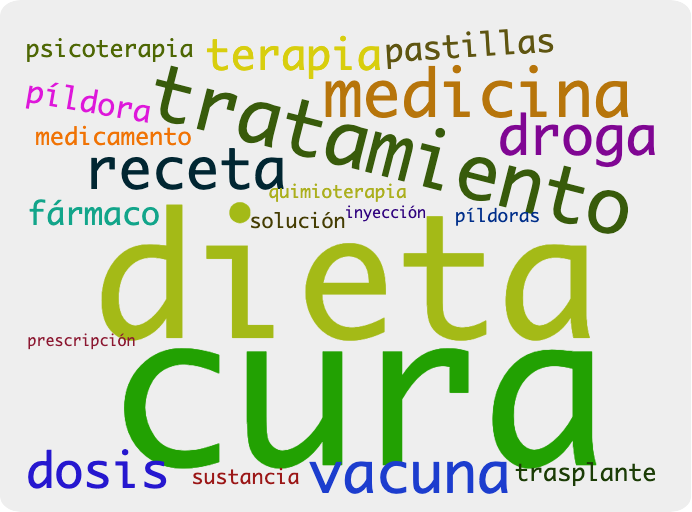}
         \caption{Medical terms in Spanish}
         \label{fig:es-en-es-encoder-6-c29}
     \end{subfigure}
     \begin{subfigure}[b]{0.24\textwidth}
         \centering
         \includegraphics[width=\textwidth]{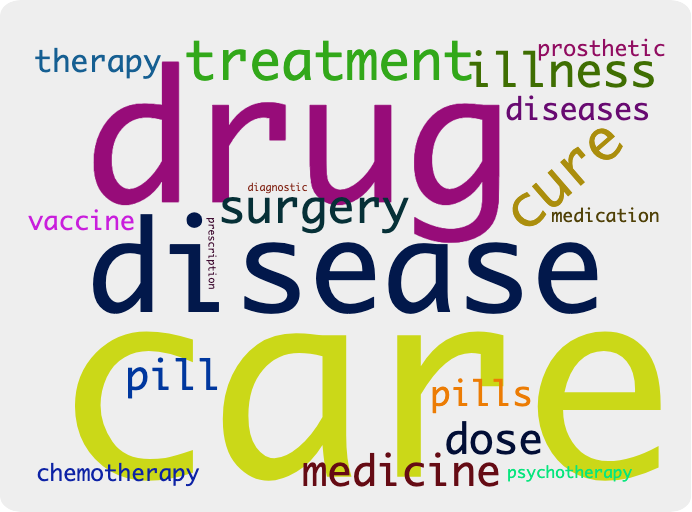}
         \caption{Medical Terms in English}
         \label{fig:es-en-encoder-6-c189}
     \end{subfigure}     
    \caption{Spanish-English Concepts learned in the mT5 model: Lower layers (a and b) capture lexical concepts, while higher layers focus on semantic concepts (c and d).}
    \label{fig:sample-clusters-spanish-english}
\end{figure*}

\begin{figure*}
     \centering
     
     \begin{subfigure}[b]{0.24\textwidth}
         \centering
         \includegraphics[width=\textwidth]{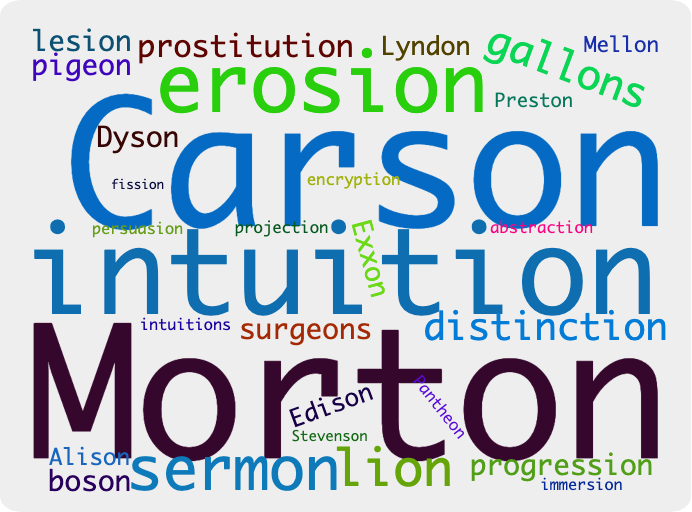}
         \caption{Words ending with ``on''}
         \label{fig:base-fr-en-en-encoder-c30-0}
     \end{subfigure}
     \begin{subfigure}[b]{0.24\textwidth}
         \centering
         \includegraphics[width=\textwidth]{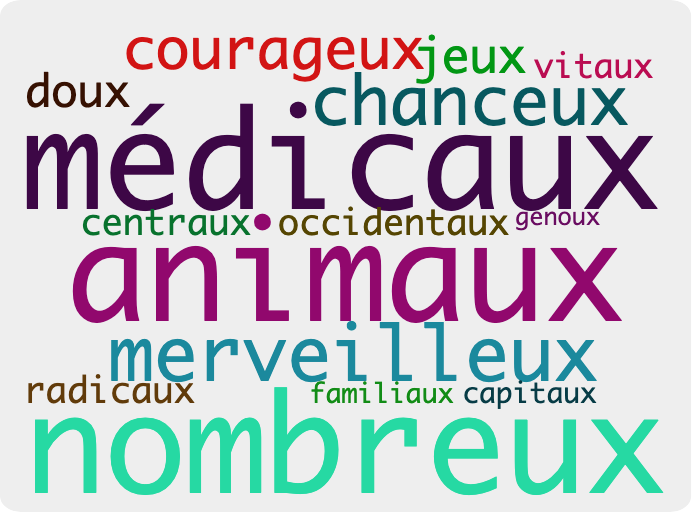}
         \caption{Words ending with ``ux''}
         \label{fig:base-fr-en-fr-encoder-c11-0}
     \end{subfigure}
     \begin{subfigure}[b]{0.24\textwidth}
         \centering
         \includegraphics[width=\textwidth]{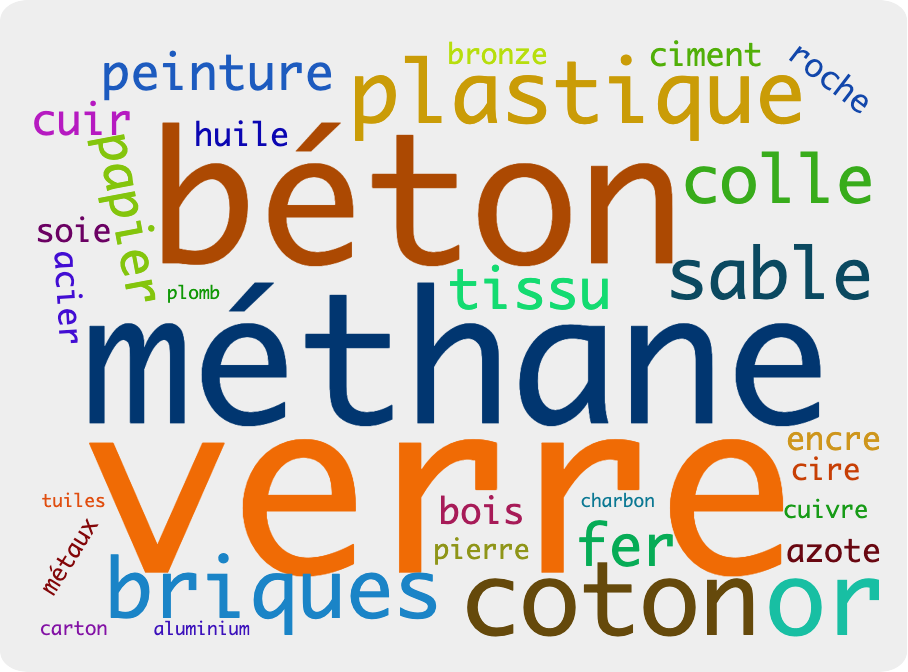}
         \caption{Materials and Substances}
         \label{fig:mt5-ft-c302-encoder-6}
     \end{subfigure}
     \begin{subfigure}[b]{0.24\textwidth}
         \centering
         \includegraphics[width=\textwidth]{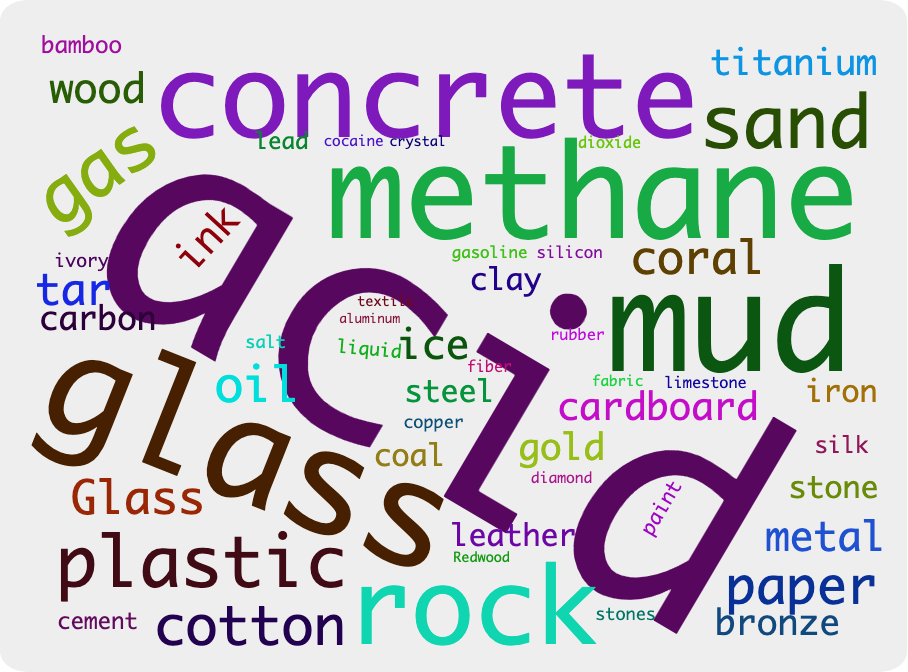}
         \caption{Materials and Substances}
         \label{fig:mt5-encoder-c21_en_encoder}
     \end{subfigure}     
    \caption{French-English Concepts learned in the mT5 model: Lower layers (a and b) capture lexical concepts, while higher layers focus on semantic concepts (c and d).}
    \label{fig:sample-clusters-french-english}
\end{figure*}

\begin{figure*}
     \centering
     
     \begin{subfigure}[b]{0.24\textwidth}
         \centering
         \includegraphics[width=\textwidth]{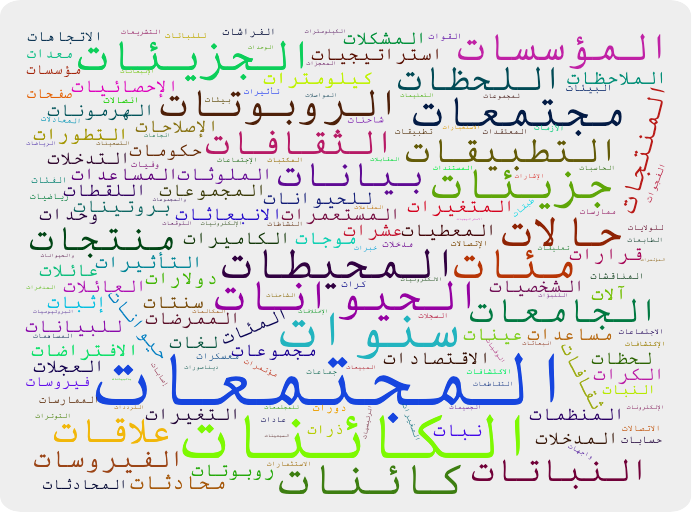}
         %\caption {\begin{arabtext}ات\end{arabtext}}
         \caption{Words ending with ``At''}
         \label{fig:mt5-ar-en-finetuned-layer-0-ar-encoder-11}
     \end{subfigure}
     \begin{subfigure}[b]{0.24\textwidth}
         \centering
         \includegraphics[width=\textwidth]{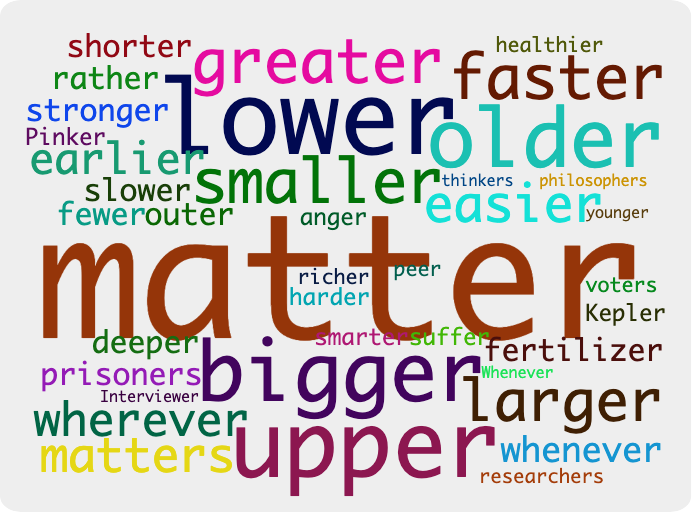}
         \caption{Shared infix ``er''}
         \label{fig:mt5-ar-en-finetuned-0-en-encoder-c187}
     \end{subfigure}
     \begin{subfigure}[b]{0.24\textwidth}
         \centering
         \includegraphics[width=\textwidth]{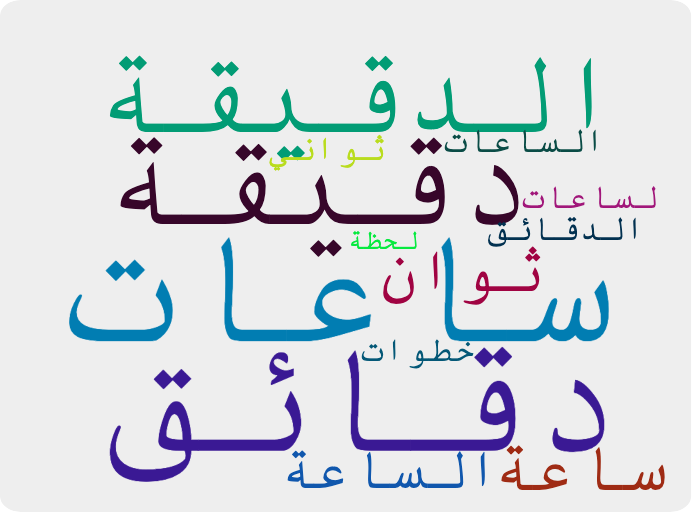}
         \caption{Time phrases in Arabic}
         \label{fig:mt5_finetuned_c17_ar_encoder}
     \end{subfigure}
     \begin{subfigure}[b]{0.24\textwidth}
         \centering
         \includegraphics[width=\textwidth]{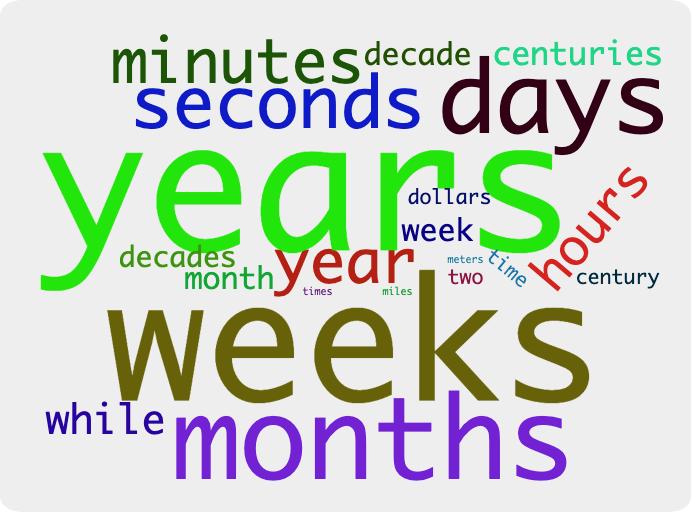}
         \caption{Time phrases in English}
         \label{fig:mt5_finetuned_c217_en_decoder}
     \end{subfigure}     
    \caption{Arabic-English Concepts learned in the mT5 model: Lower layers (a and b) capture lexical concepts, while higher layers focus on semantic concepts (c and d).}
    \label{fig:sample-clusters-arabic-english}
\end{figure*}

\paragraph{Deeper layers in multilingual models reveal increased alignment and preserve semantic concepts, contrasting with language-dependent lexical learning in lower layers.} We made this observation through qualitative analysis of concepts across different languages we studied in this paper. In Figures \ref{fig:sample-clusters-spanish-english}--\ref{fig:sample-clusters-arabic-english}, we present lexical concepts learned within the lower layers of the multilingual models, contrasting with the aligned semantic concepts found in the higher layers. To verify our hypothesis, we quantify the number of lexical (suffix-based concepts) and semantic concepts in English within the mBERT model. Please see Figure \ref{fig:mbert-layerwise-trends} for a layer-wise pattern of concepts.
%We consistently observed a significant number of concepts that exhibited alignment within the latent spaces of these models. Notably, up to 49\% of concepts demonstrated alignment within the Arabic-English latent space of the mBERT-NER model. The results also show that Please refer to Figures \ref{fig:spanish-concept-alignment}--\ref{fig:arabic-concept-alignment} for results across different languages. Please refer to Figures \ref{fig:german-english-alignment-encoder}--\ref{fig:arabic-alignment-encoder} for additional examples of concepts aligned across various languages.

\begin{figure*}
    \centering

      \begin{subfigure}[b]{0.24\textwidth}
         \centering
         \includegraphics[width=\textwidth]{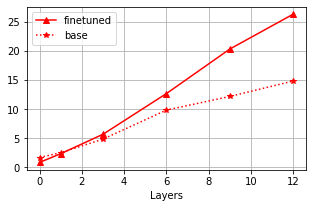}
         \caption{mT5--encoder}
         \label{fig:mt5-encoder-alignment-es}
     \end{subfigure}
     \begin{subfigure}[b]{0.24\textwidth}
         \centering
         \includegraphics[width=\textwidth]{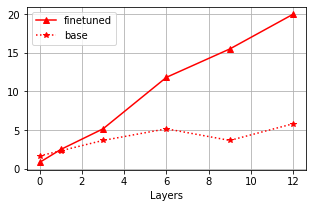}
         \caption{mT5--encoder-decoder}
         \label{fig:mt5-encoder-decoder-alignment-es}
     \end{subfigure}
    \begin{subfigure}[b]{0.24\textwidth}
         \centering
         \includegraphics[width=\textwidth]{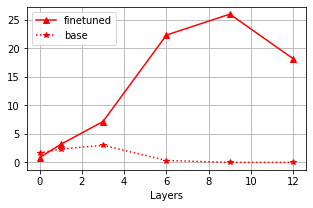}
         \caption{mT5--decoder}
         \label{fig:mt5-decoder-alignment-es}
     \end{subfigure}
     \begin{subfigure}[b]{0.24\textwidth}
         \centering
         \includegraphics[width=\textwidth]{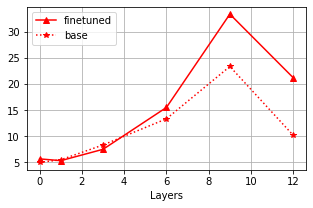}
         \caption{mBERT}
         \label{fig:mbert-ner-es-en-alignment}
     \end{subfigure}
     
    \caption{Quantifying Alignment Percentage in Spanish–English Concepts: Dotted lines depict base models, while solid lines represent fine-tuned models across different multilingual models.}
    \label{fig:spanish-concept-alignment}
\end{figure*}

\begin{figure*}
    \centering
      \begin{subfigure}[b]{0.24\textwidth}
         \centering
         \includegraphics[width=\textwidth]{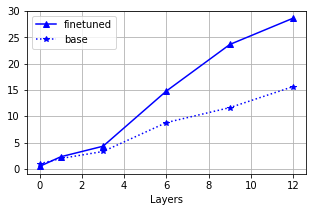}
         \caption{mT5--encoder}
         \label{fig:mt5-encoder-alignment-fr}
     \end{subfigure}
    \begin{subfigure}[b]{0.24\textwidth}
         \centering
         \includegraphics[width=\textwidth]{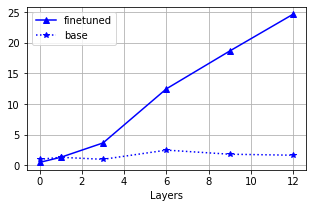}
         \caption{mT5--encoder-decoder}
         \label{fig:mt5-encoder-decoder-alignment-fr}
     \end{subfigure}
         \begin{subfigure}[b]{0.24\textwidth}
         \centering
         \includegraphics[width=\textwidth]{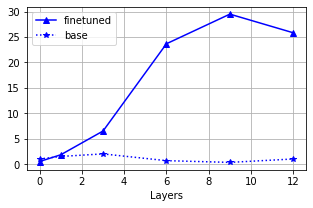}
         \caption{mT5--decoder}
         \label{fig:mt5-decoder-alignment-fr}
     \end{subfigure}
     \begin{subfigure}[b]{0.24\textwidth}
         \centering
         \includegraphics[width=\textwidth]{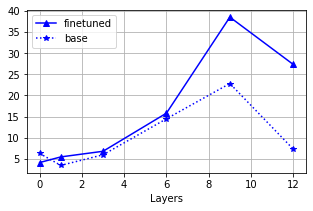}
         \caption{mBERT}
         \label{fig:mbert-ner-fr-en-alignment}
     \end{subfigure}
    \caption{Quantifying Alignment Percentage in French–English Concepts: Dotted lines depict base models, while solid lines represent fine-tuned models across different multilingual models.}
    \label{fig:french-concept-alignment}
\end{figure*}

\begin{figure*}
    \centering

      \begin{subfigure}[b]{0.24\textwidth}
         \centering
         \includegraphics[width=\textwidth]{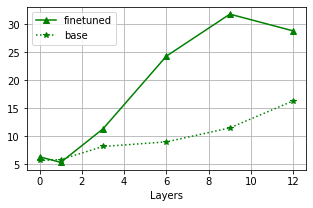}
         \caption{mT5--encoder}
         \label{fig:mt5-encoder-alignment-ar}
     \end{subfigure}
      \begin{subfigure}[b]{0.24\textwidth}
         \centering
         \includegraphics[width=\textwidth]{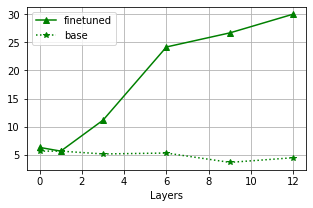}
         \caption{mT5 encoder-decoder}
         \label{fig:mt5-encoder-decoder-ar}
     \end{subfigure}
    \begin{subfigure}[b]{0.24\textwidth}
         \centering
         \includegraphics[width=\textwidth]{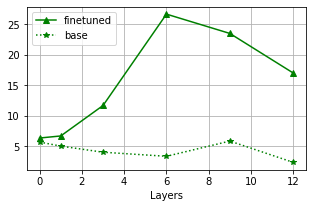}
         \caption{mT5 decoder}
         \label{fig:mt5-decoder-ar}
     \end{subfigure}
     \begin{subfigure}[b]{0.24\textwidth}
         \centering
         \includegraphics[width=\textwidth]{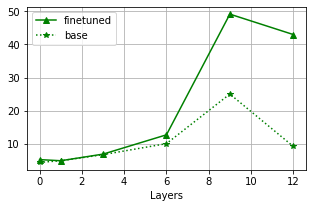}
         \caption{mBERT}
         \label{fig:mbert-arabic-alignment}
     \end{subfigure}
    \caption{Quantifying Alignment Percentage in Arabic–English Concepts: Dotted lines depict base models, while solid lines represent fine-tuned models across different multilingual models.}
    \label{fig:arabic-concept-alignment}
\end{figure*}

\begin{figure*}
     \centering
     \begin{subfigure}[b]{0.24\textwidth}
         \centering
         \includegraphics[width=\textwidth]{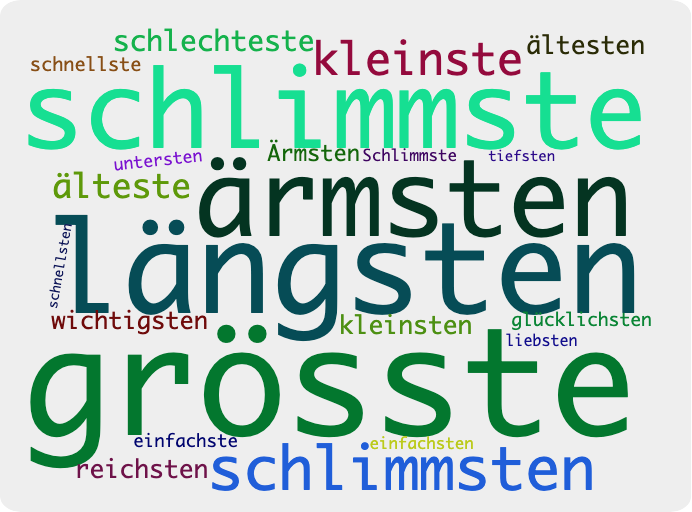}
         \caption{Superlatives in German}
         \label{fig:de-german-encoder-6-c507}
     \end{subfigure}
     \begin{subfigure}[b]{0.24\textwidth}
         \centering
         \includegraphics[width=\textwidth]{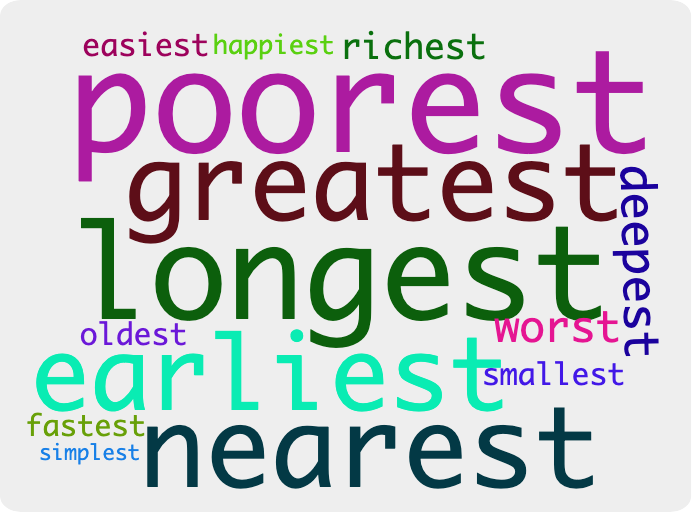}
         \caption{Superlatives in English}
         \label{fig:de-en-english-encoder-6-c93}
     \end{subfigure}
    \begin{subfigure}[b]{0.24\textwidth}
         \centering
         \includegraphics[width=\textwidth]{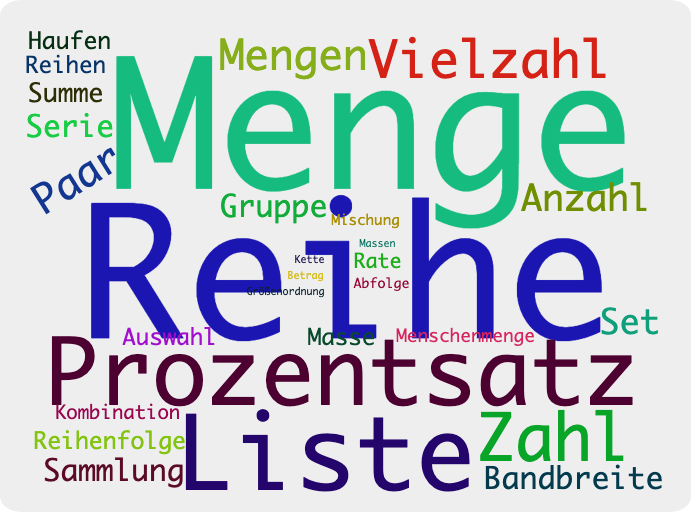}
         \caption{Math related terms (de)}
         \label{fig:de-en-german-encoder-9-c349}
     \end{subfigure}
     \begin{subfigure}[b]{0.24\textwidth}
         \centering
         \includegraphics[width=\textwidth]{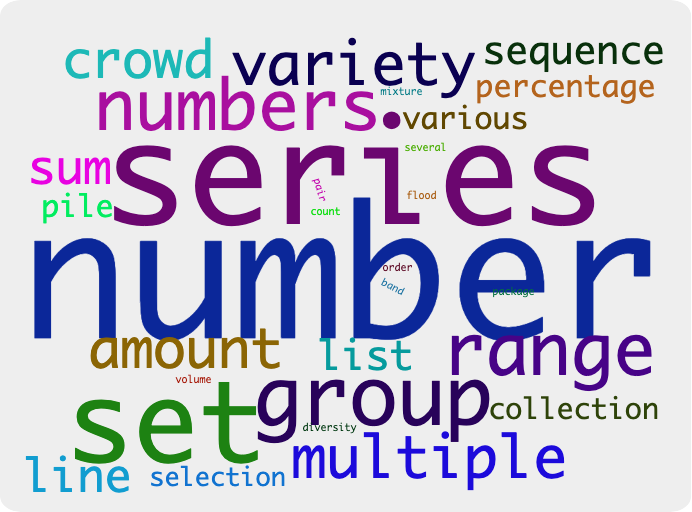}
         \caption{Math related terms (en)}
         \label{fig:de-en-english-encoder-8-c533}
     \end{subfigure}
     \begin{subfigure}[b]{0.24\textwidth}
         \centering
         \includegraphics[width=\textwidth]{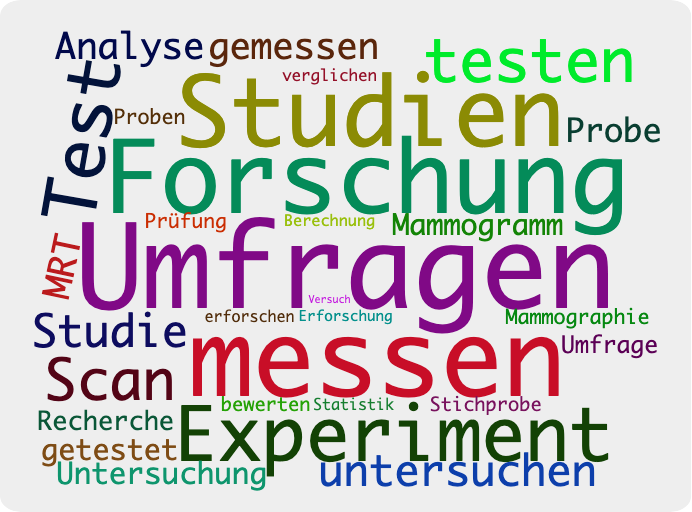}
         \caption{Study related (de)}
         \label{fig:de-en-german-encoder-12-c287}
     \end{subfigure}
     \begin{subfigure}[b]{0.24\textwidth}
         \centering
         \includegraphics[width=\textwidth]{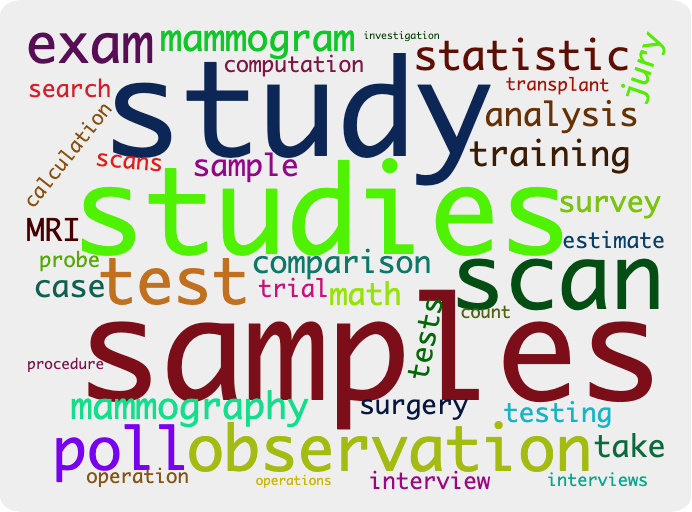}
         \caption{Study related (en)}
         \label{fig:de-en-english-encoder-12-c444}
     \end{subfigure}
     \begin{subfigure}[b]{0.24\textwidth}
         \centering
         \includegraphics[width=\textwidth]{figures/cluster-alignment-plots/de-en/de-en-model-german-encoder-12-c339.png}
         \caption{Colors in German}
         \label{fig:de-en-german-encoder-12-c399}
     \end{subfigure}
     \begin{subfigure}[b]{0.24\textwidth}
         \centering
         \includegraphics[width=\textwidth]{figures/cluster-alignment-plots/de-en/de-en-model-english-encoder-12-c531.png}
         \caption{Colors in English}
         \label{fig:de-en-english-encoder-12-c531}
     \end{subfigure}
     
    \caption{Pairs of Concepts in German-English mT5 model}
    \label{fig:german-english-alignment-encoder}
\end{figure*}

\begin{figure*}
         \centering
         \begin{subfigure}[b]{0.24\textwidth}
             \centering
             \includegraphics[width=\textwidth]{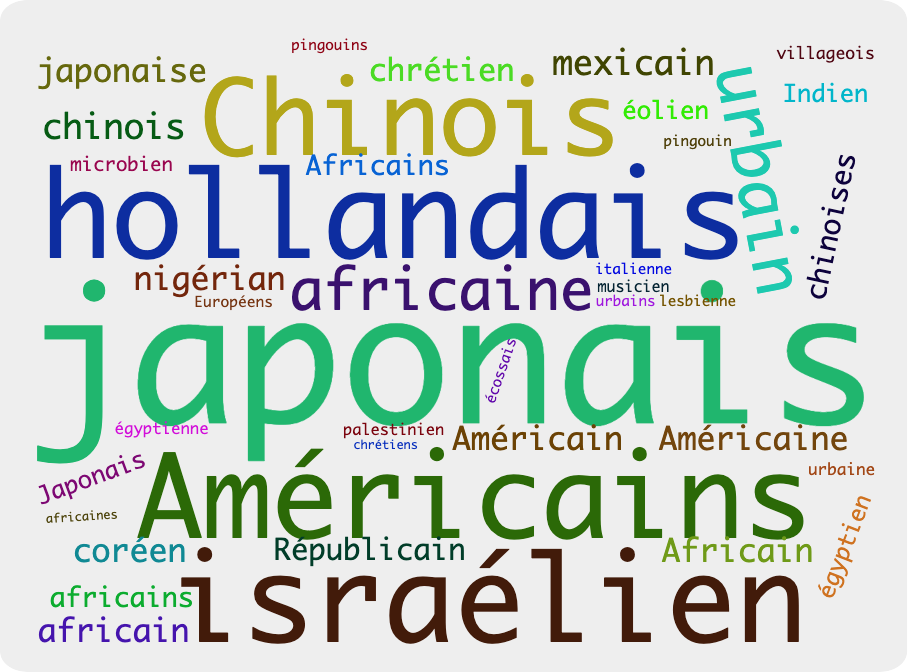}
             \caption{Nationality \& Identity (fr)}
             \label{fig:mt5-finetuned-6-c219}
         \end{subfigure}
         \begin{subfigure}[b]{0.24\textwidth}
             \centering
             \includegraphics[width=\textwidth]{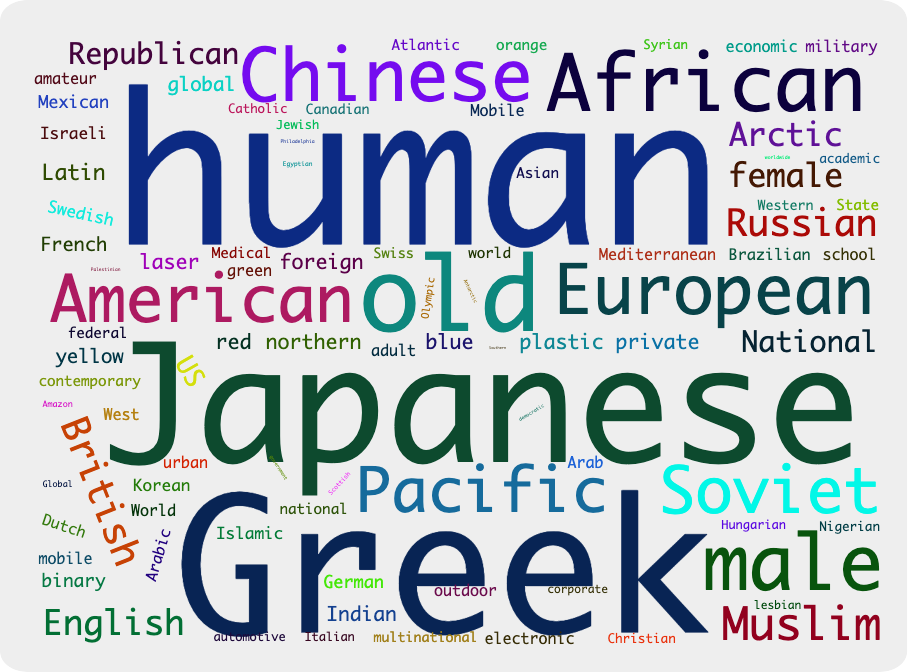}
             \caption{Nationality \& Identity (en)}
             \label{fig:mt5-finetuned-6-decoder}
         \end{subfigure}
        \begin{subfigure}[b]{0.24\textwidth}
             \centering
             \includegraphics[width=\textwidth]{figures/cluster-alignment-plots/fr-en/mt5_finetuned_c302_fr_encoder_6.png}
             \caption{Chemical Materials (fr)}
             \label{fig:mt5-finetuned-6-c302}
         \end{subfigure}
         \begin{subfigure}[b]{0.24\textwidth}
             \centering
             \includegraphics[width=\textwidth]{figures/cluster-alignment-plots/fr-en/mt5_finetuned_c21_en_encoder_6.png}
             \caption{Chemical Material (en)}
             \label{fig:mt5-finetuned-c21-6-enc}
         \end{subfigure}
         \begin{subfigure}[b]{0.24\textwidth}
             \centering
             \includegraphics[width=\textwidth]{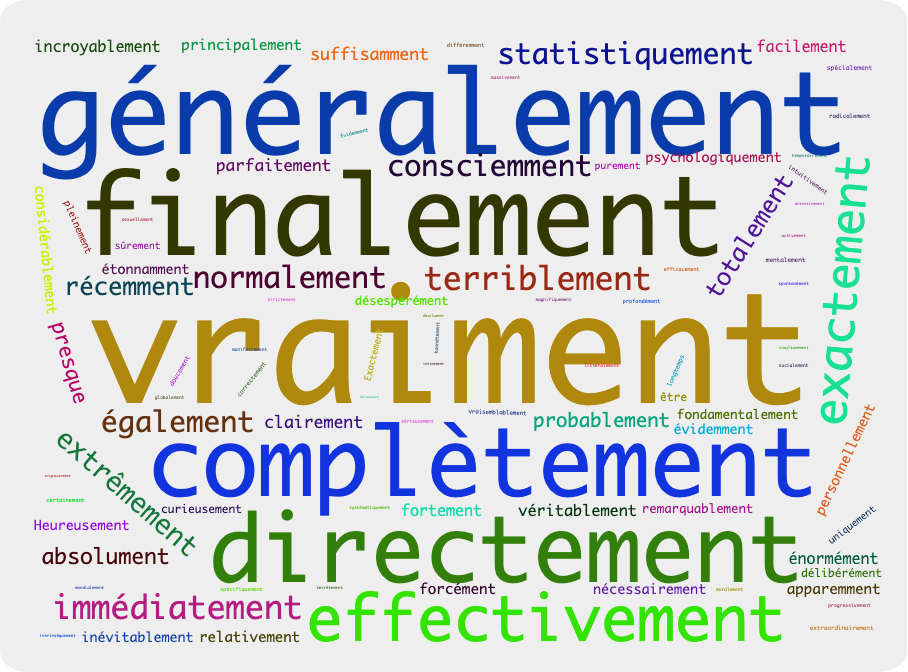}
             \caption{Adverbs (fr)}
             \label{fig:mt5-finetuned-3-c46-fr-decoder}
         \end{subfigure}
         \begin{subfigure}[b]{0.24\textwidth}
             \centering
             \includegraphics[width=\textwidth]{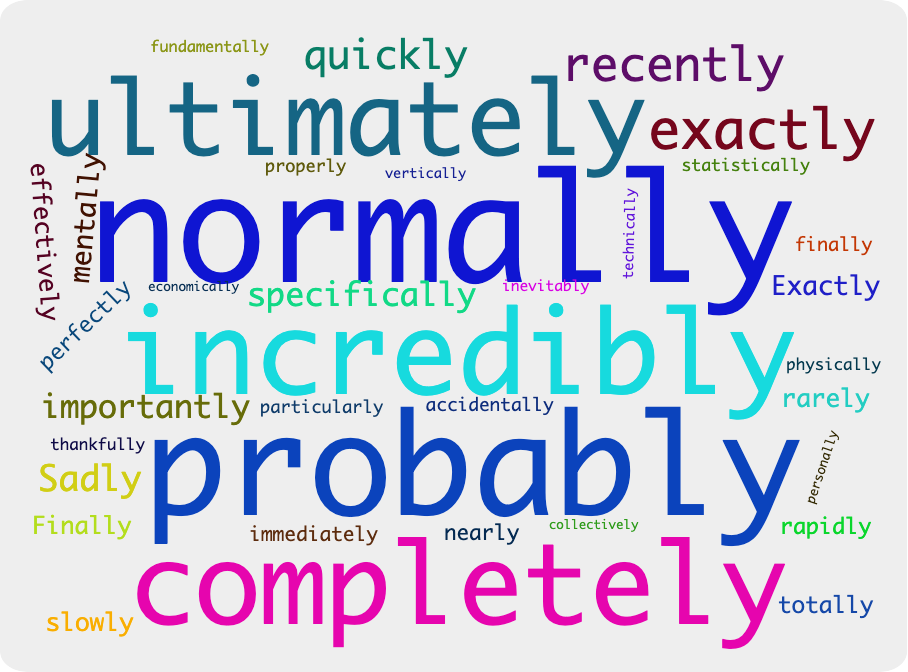}
             \caption{Adverbs (en)}
             \label{fig:mt5-finetuned-3-c67-en-decoder}
         \end{subfigure}
        \caption{Pairs of Concepts in French-English mT5 model}
        \label{fig:french-alignment}
    \end{figure*}

\begin{figure*}
     \centering
    \begin{subfigure}[b]{0.24\textwidth}
         \centering
         \includegraphics[width=\textwidth]{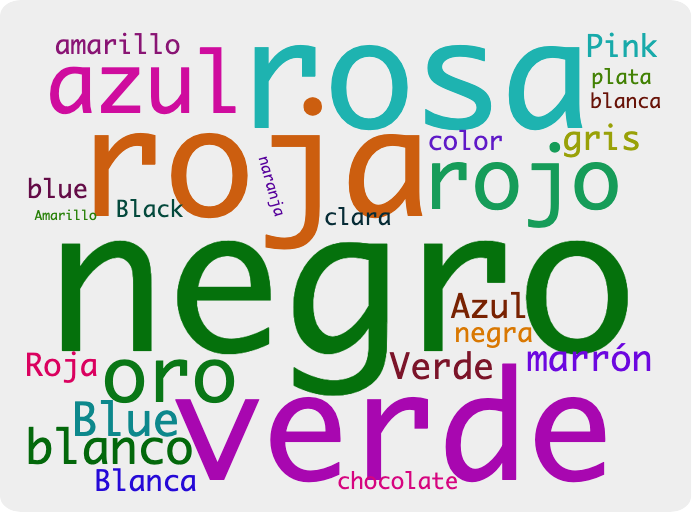}
         \caption{Colors in English}
         \label{fig:es-encoder-12-c575}
     \end{subfigure}
     \begin{subfigure}[b]{0.24\textwidth}
         \centering
         \includegraphics[width=\textwidth]{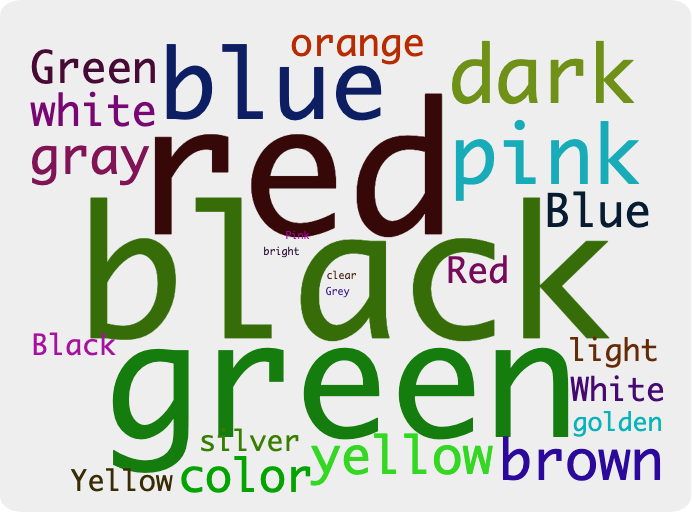}
         \caption{Colors in English}
         \label{fig:en-encoder-12-c232}
     \end{subfigure}
     \begin{subfigure}[b]{0.24\textwidth}
         \centering
         \includegraphics[width=\textwidth]{figures/cluster-alignment-plots/es-en/encoder-encoder/spanish-encoder-6-c29.png}
         \caption{Medical terms in Spanish}
         \label{fig:es-encoder-6-c29}
     \end{subfigure}
     \begin{subfigure}[b]{0.24\textwidth}
         \centering
         \includegraphics[width=\textwidth]{figures/cluster-alignment-plots/es-en/encoder-encoder/english-encoder-6-c189.png}
         \caption{Medical terms in English}
         \label{fig:en-encoder-6-c189}
     \end{subfigure}
     \begin{subfigure}[b]{0.24\textwidth}
         \centering
          \includegraphics[width=\textwidth]{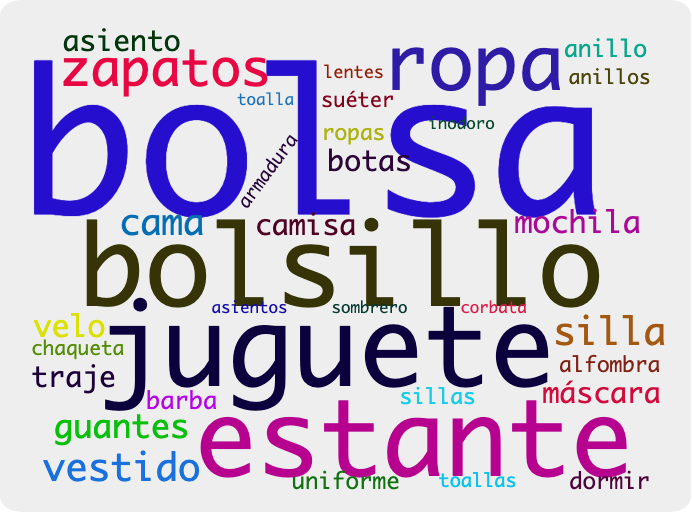}
         \caption{Assorted Items (es)}
         \label{fig:es-encoder-9-c345}
     \end{subfigure}
     \begin{subfigure}[b]{0.24\textwidth}
         \centering
         \includegraphics[width=\textwidth]{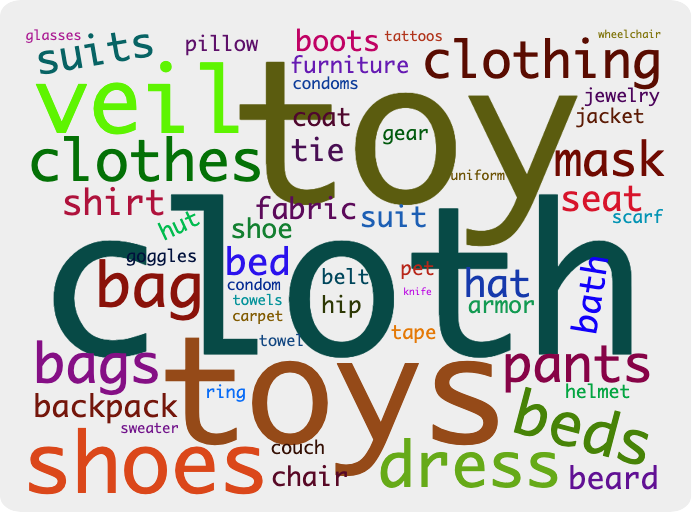}
         \caption{Assorted Items (en)}
         \label{fig:en-encoder-9-c44}
     \end{subfigure}
     % \begin{subfigure}[b]{0.24\textwidth}
     %     \centering
     %     \includegraphics[width=\textwidth]{figures/cluster-alignment-plots/es-en/encoder-encoder/spanish-encoder-1-c40.png}
     %     \caption{Variations in states (es)}
     %     \label{fig:es-encoder-1-c40}
     % \end{subfigure}
     % \begin{subfigure}[b]{0.24\textwidth}
     %     \centering
     %     \includegraphics[width=\textwidth]{figures/cluster-alignment-plots/es-en/encoder-encoder/english-encoder-1-c120.png}
     %     \caption{Variations in states (en)}
     %     \label{fig:en-encoder-1-c120}
     % \end{subfigure}
       \begin{subfigure}[b]{0.24\textwidth}
         \centering
         \includegraphics[width=\textwidth]{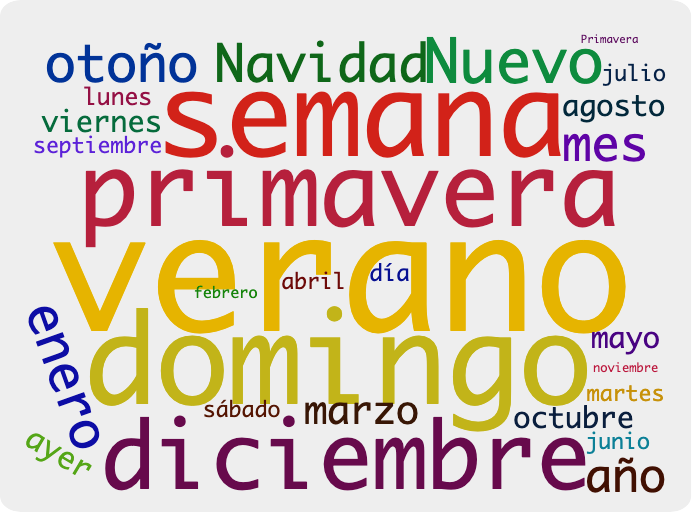}
         \caption{Temporal terms in Spanish}
         \label{fig:es-encoder-9-c212}
     \end{subfigure}
     \begin{subfigure}[b]{0.24\textwidth}
         \centering
         \includegraphics[width=\textwidth]{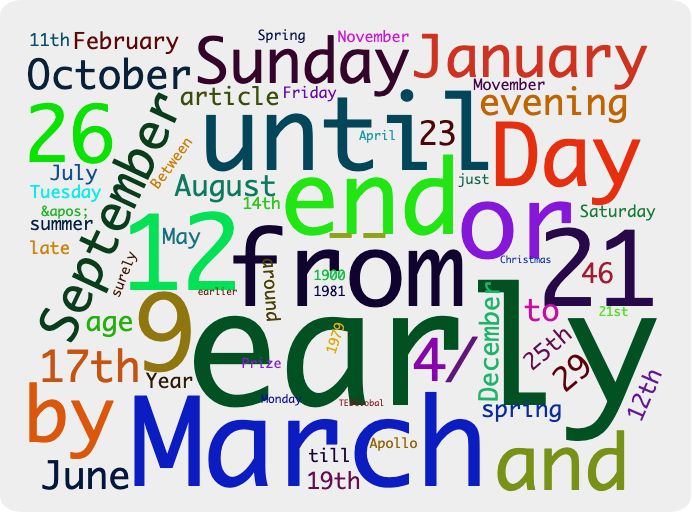}
         \caption{Temporal terms in English}
         \label{fig:en-decoder-9-c526}
     \end{subfigure}
    \caption{Pairs of Concepts in Spanish-English mT5 model}
    \label{fig:spanish-english-alignment-encoder}
\end{figure*}

\begin{figure*}
     \centering
      \begin{subfigure}[b]{0.24\textwidth}
         \centering
         \includegraphics[width=\textwidth]{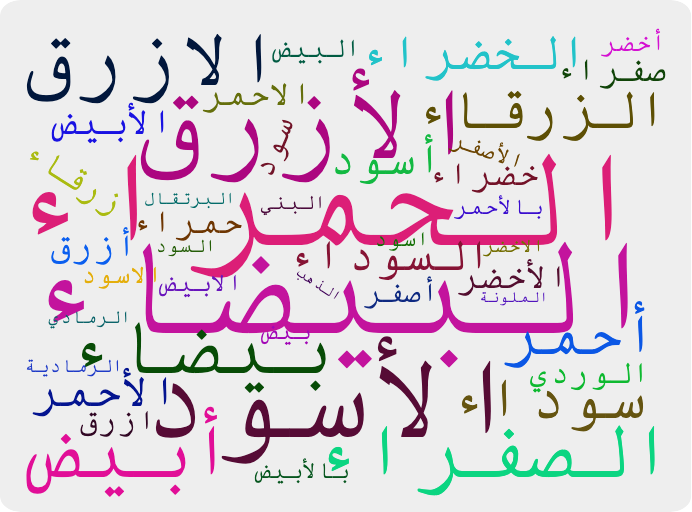}
         \caption{Colors in Arabic}
         \label{fig:mt5-ar-c204-12-encoder}
     \end{subfigure}
     \begin{subfigure}[b]{0.24\textwidth}
         \centering
         \includegraphics[width=\textwidth]{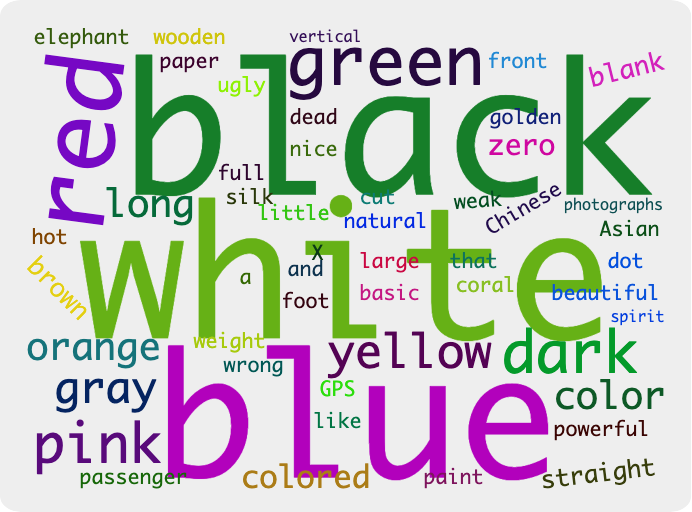}
         \caption{Colors in English}
         \label{fig:mt5-en-c374-12-decoder}
     \end{subfigure}
     \begin{subfigure}[b]{0.24\textwidth}
         \centering
         \includegraphics[width=\textwidth]{figures/cluster-alignment-plots/ar-en/mt5_finetuned_c17_ar_encoder.png}
         \caption{Time spans in Arabic}
         \label{fig:mt5-ar-c17-12-encoder}
     \end{subfigure}
     \begin{subfigure}[b]{0.24\textwidth}
         \centering
         \includegraphics[width=\textwidth]{figures/cluster-alignment-plots/ar-en/mt5_finetuned_c217_en_decoder.png}
         \caption{Time spans in English}
         \label{fig:mt5-en-c217-12-decoder}
     \end{subfigure}

     \begin{subfigure}[b]{0.24\textwidth}
         \centering
         \includegraphics[width=\textwidth]{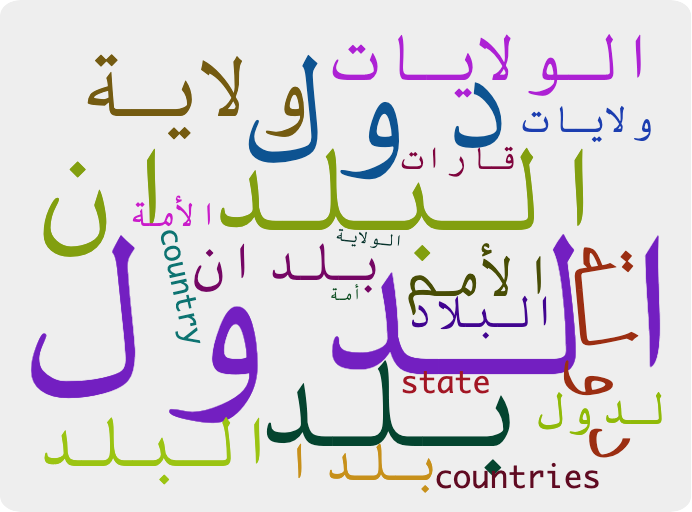}
         \caption{Geographical entities (ar)}
         \label{fig:mt5-ar-c12-12-encoder}
     \end{subfigure}
     \begin{subfigure}[b]{0.24\textwidth}
         \centering
         \includegraphics[width=\textwidth]{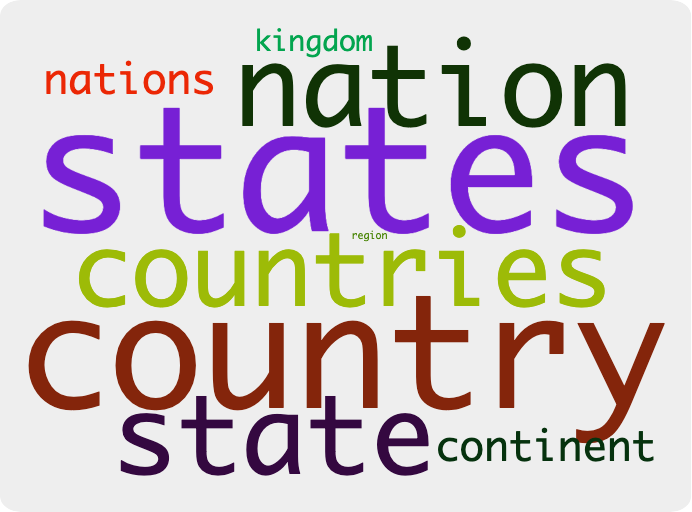}
         \caption{Geographical entities (en)}
         \label{fig:mt5-en-c374-12-encoder}
     \end{subfigure}
     \begin{subfigure}[b]{0.24\textwidth}
         \centering
         \includegraphics[width=\textwidth]{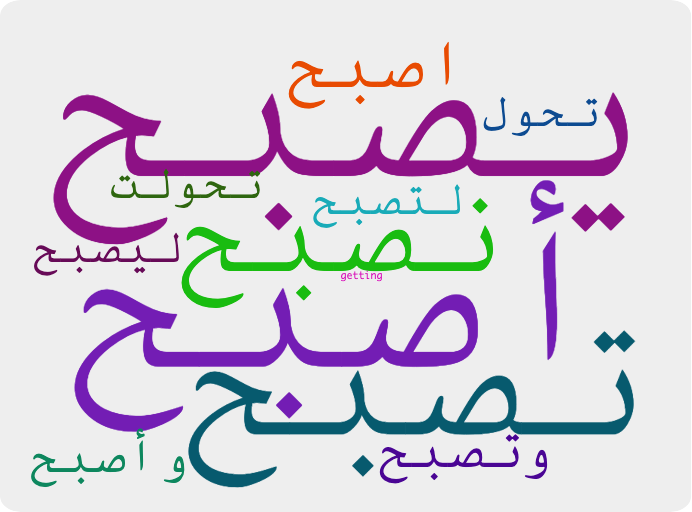}
         \caption{Morphological variations}
         \label{fig:mt5-ar-c82-12-encoder}
     \end{subfigure}
     \begin{subfigure}[b]{0.24\textwidth}
         \centering
         \includegraphics[width=\textwidth]{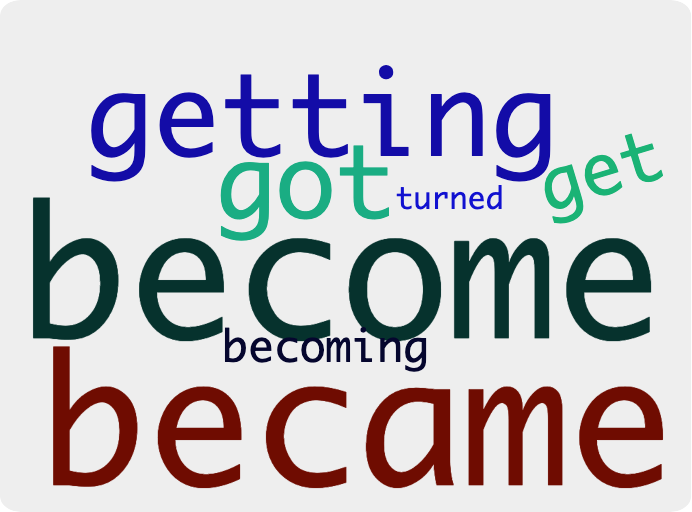}
         \caption{Verb transformations}
         \label{fig:mt5-en-c123-12-encoder}
     \end{subfigure}
    
    \caption{Pairs of Aligned Concepts in Arabic-English mT5 model}
    \label{fig:arabic-alignment-encoder}
\end{figure*}

\paragraph{Fine-tuning calibrates the latent space towards higher alignment} We consistently higher alignment of concepts as the models were fine-tuned towards a downstream NLP task. Please refer to Figures \ref{fig:spanish-concept-alignment}--\ref{fig:arabic-concept-alignment} for results across different architectures and languages. We display alignment outcomes in base models (dotted lines) and after they were fine-tuned (solid lines). Please refer to Figures \ref{fig:german-english-alignment-encoder}--\ref{fig:arabic-alignment-encoder} for additional examples of concepts aligned across various languages.

\begin{figure*}
     \centering
     \begin{subfigure}[b]{0.3\textwidth}
         \centering
         \includegraphics[width=\textwidth]{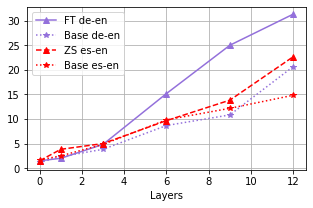}
         \caption{zero-shot \textbf{es} on \textbf{de} encoder}
         \label{fig:de-en-es-en-encoder-german-model-cluster-alignment}
     \end{subfigure}
     \hfill
     \begin{subfigure}[b]{0.3\textwidth}
         \centering
         \includegraphics[width=\textwidth]{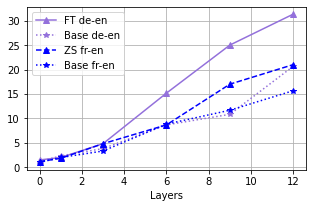}
         \caption{zero-shot \textbf{fr} on \textbf{de} encoder}
         \label{fig:de-en-fr-en-encoder-german-model-cluster-alignment}
     \end{subfigure}
     \hfill
     \begin{subfigure}[b]{0.3\textwidth}
         \centering
         \includegraphics[width=\textwidth]{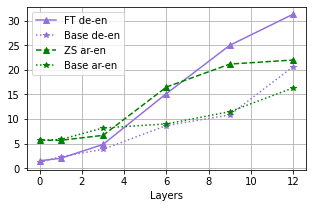}
         \caption{zero-shot \textbf{ar} on \textbf{de} encoder}
         \label{fig:de-en-ar-en-encoder-german-model-cluster-alignment}
     \end{subfigure}
       \begin{subfigure}[b]{0.3\textwidth}
         \centering
         \includegraphics[width=\textwidth]{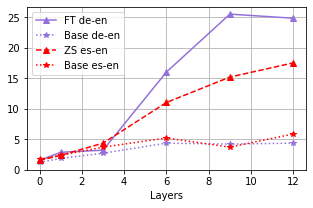}
         \caption{zero-shot \textbf{es} on \textbf{de$\leftrightarrow$en}}
         \label{fig:de-en-es-en-encoder-decoder-german-model-cluster-alignment}
     \end{subfigure}
     \hfill
     \begin{subfigure}[b]{0.3\textwidth}
         \centering
         \includegraphics[width=\textwidth]{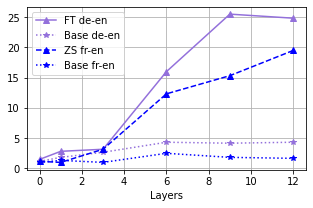}
         \caption{zero-shot \textbf{fr} on \textbf{de$\leftrightarrow$en}}
         \label{fig:de-en-fr-en-encoder-decoder-german-model-cluster-alignment}
     \end{subfigure}
     \hfill
     \begin{subfigure}[b]{0.3\textwidth}
         \centering
         \includegraphics[width=\textwidth]{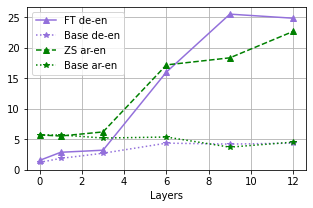}
         \caption{zero-shot \textbf{ar} on \textbf{de$\leftrightarrow$en}}
         \label{fig:de-en-ar-en-encoder-decoder-german-model-cluster-alignment}
     \end{subfigure}

     \begin{subfigure}[b]{0.3\textwidth}
         \centering
         \includegraphics[width=\textwidth]{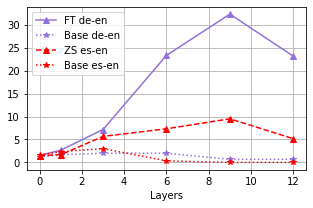}
         \caption{zero-shot \textbf{es} on \textbf{de} decoder}
         \label{fig:de-en-es-en-decoder-german-model-cluster-alignment}
     \end{subfigure}
     \hfill
     \begin{subfigure}[b]{0.3\textwidth}
         \centering
         \includegraphics[width=\textwidth]{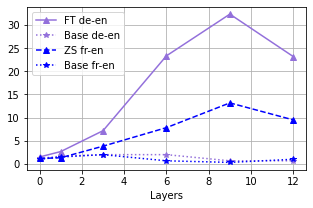}
         \caption{zero-shot \textbf{fr} on \textbf{de} decoder}
         \label{fig:de-en-fr-en-decoder-german-model-cluster-alignment}
     \end{subfigure}
     \hfill
     \begin{subfigure}[b]{0.3\textwidth}
         \centering
         \includegraphics[width=\textwidth]{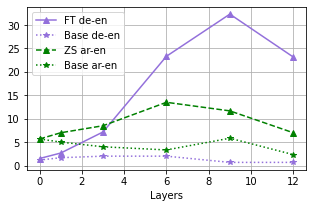}
         \caption{zero-shot \textbf{ar} on \textbf{de} decoder}
         \label{fig:de-en-ar-en-decoder-german-model-cluster-alignment}
     \end{subfigure}

     \caption{Percentage of Aligned Concepts: Dotted lines represent base models, solid lines denote fine-tuned German–English model, and dashed lines depict zero-shot alignment for spanish (left column), French–English (Middle column) and Arabic-English (right column); enc: Encoder, dec: Decoder}
    \label{fig:german-model-alignment}
\end{figure*}

\begin{figure*}
     \centering
     \begin{subfigure}[b]{0.3\textwidth}
         \centering
         \includegraphics[width=\textwidth]{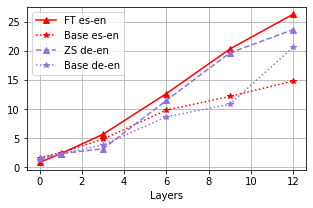}
         \caption{zero-shot \textbf{de} on \textbf{es} encoder}
         \label{fig:es-en-de-en-encoder-spanish-model-cluster-alignment}
     \end{subfigure}
     \hfill
     \begin{subfigure}[b]{0.3\textwidth}
         \centering
         \includegraphics[width=\textwidth]{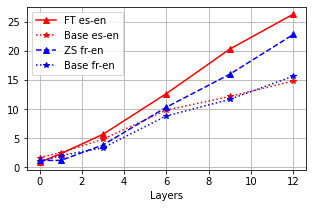}
         \caption{zero-shot \textbf{fr} on \textbf{es} encoder}
         \label{fig:es-en-fr-en-encoder-spanish-model-cluster-alignment}
     \end{subfigure}
     \hfill
     \begin{subfigure}[b]{0.3\textwidth}
         \centering
         \includegraphics[width=\textwidth]{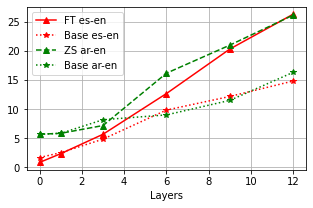}
         \caption{zero-shot \textbf{ar} on \textbf{es} encoder}
         \label{fig:es-en-ar-en-encoder-spanish-model-cluster-alignment}
     \end{subfigure}

       \begin{subfigure}[b]{0.3\textwidth}
         \centering
         \includegraphics[width=\textwidth]{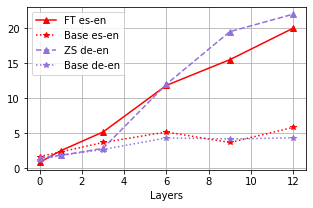}
         \caption{zero-shot \textbf{de} on \textbf{es$\leftrightarrow$en}}
         \label{fig:es-en-de-en-encoder-decoder-spanish-model-cluster-alignment}
     \end{subfigure}
     \hfill
     \begin{subfigure}[b]{0.3\textwidth}
         \centering
         \includegraphics[width=\textwidth]{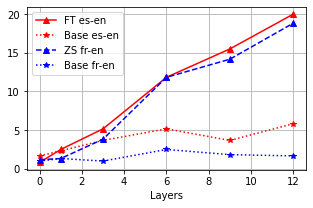}
         \caption{zero-shot \textbf{fr} on \textbf{es$\leftrightarrow$en}}
         \label{fig:es-en-fr-en-encoder-decoder-spanish-model-cluster-alignment}
     \end{subfigure}
     \hfill
     \begin{subfigure}[b]{0.3\textwidth}
         \centering
         \includegraphics[width=\textwidth]{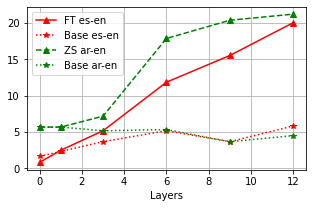}
         \caption{zero-shot \textbf{ar} on \textbf{es$\leftrightarrow$en}}
         \label{fig:es-en-ar-en-encoder-decoder-spanish-model-cluster-alignment}
     \end{subfigure}

     \begin{subfigure}[b]{0.3\textwidth}
         \centering
         \includegraphics[width=\textwidth]{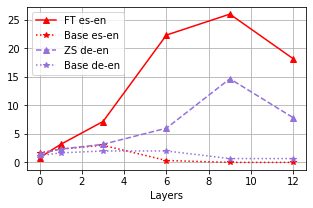}
         \caption{zero-shot \textbf{de} on \textbf{es} decoder}
         \label{fig:es-en-de-en-decoder-cluster-alignment-spanish-model}
     \end{subfigure}
     \hfill
     \begin{subfigure}[b]{0.3\textwidth}
         \centering
         \includegraphics[width=\textwidth]{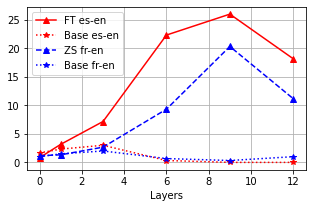}
         \caption{zero-shot \textbf{fr} on \textbf{es} decoder}
         \label{fig:es-en-fr-en-decoder-spanish-model-cluster-alignment}
     \end{subfigure}
     \hfill
     \begin{subfigure}[b]{0.3\textwidth}
         \centering
         \includegraphics[width=\textwidth]{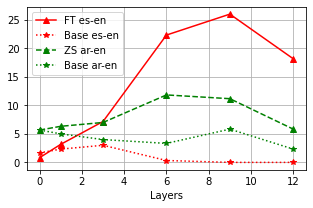}
         \caption{zero-shot \textbf{ar} on \textbf{es} decoder}
         \label{fig:tbr}
     \end{subfigure}

     \caption{Percentage of Aligned Concepts: Dotted lines represent base models, solid lines denote fine-tuned Spanish–English model, and dashed lines depict zero-shot alignment for German-English (left column), French–English (Middle column) and Arabic-English (right column); enc: Encoder, dec: Decoder}
    \label{fig:spanish-model-alignment}
\end{figure*}

\begin{figure*}
     \centering
     \begin{subfigure}[b]{0.3\textwidth}
         \centering
         \includegraphics[width=\textwidth]{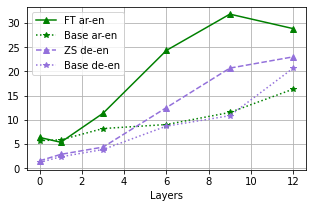}
         \caption{zero-shot \textbf{de} on \textbf{ar} encoder}
         \label{fig:ar-en-de-en-encoder-arabic-model-cluster-alignment}
     \end{subfigure}
     \hfill
     \begin{subfigure}[b]{0.3\textwidth}
         \centering
         \includegraphics[width=\textwidth]{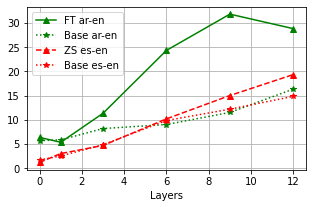}
         \caption{zero-shot \textbf{es} on \textbf{ar} encoder}
         \label{fig:ar-en-es-en-encoder-arabic-model-cluster-alignment}
     \end{subfigure}
     \hfill
     \begin{subfigure}[b]{0.3\textwidth}
         \centering
         \includegraphics[width=\textwidth]{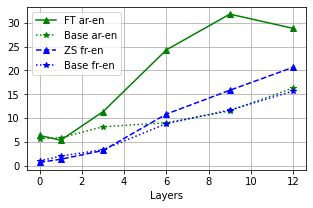}
         \caption{zero-shot \textbf{de} on \textbf{fr} encoder}
         \label{fig:ar-en-fr-en-encoder-arabic-model-cluster-alignment}
     \end{subfigure}

       \begin{subfigure}[b]{0.3\textwidth}
         \centering
         \includegraphics[width=\textwidth]{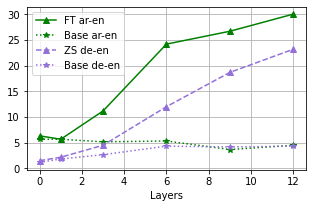}
         \caption{zero-shot \textbf{de} on \textbf{ar$\leftrightarrow$en}}
         \label{fig:ar-en-de-en-encoder-decoder-arabic-model-cluster-alignment}
     \end{subfigure}
     \hfill
     \begin{subfigure}[b]{0.3\textwidth}
         \centering
         \includegraphics[width=\textwidth]{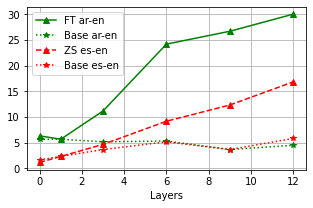}
         \caption{zero-shot \textbf{es} on \textbf{ar$\leftrightarrow$en}}
         \label{fig:ar-en-es-en-encoder-decoder-arabic-model-cluster-alignment}
     \end{subfigure}
     \hfill
     \begin{subfigure}[b]{0.3\textwidth}
         \centering
         \includegraphics[width=\textwidth]{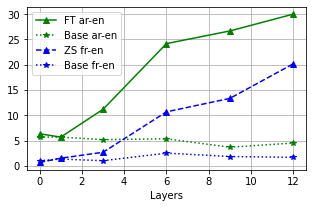}
         \caption{zero-shot \textbf{fr} on \textbf{ar$\leftrightarrow$en}}
         \label{fig:ar-en-fr-en-encoder-decoder-arabic-model-cluster-alignment}
     \end{subfigure}

     \begin{subfigure}[b]{0.3\textwidth}
         \centering
         \includegraphics[width=\textwidth]{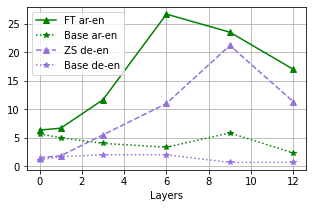}
         \caption{zero-shot \textbf{de} on \textbf{ar} decoder}
         \label{fig:ar-en-de-en-decoder-arabic-model-cluster-alignment}
     \end{subfigure}
     \hfill
     \begin{subfigure}[b]{0.3\textwidth}
         \centering
         \includegraphics[width=\textwidth]{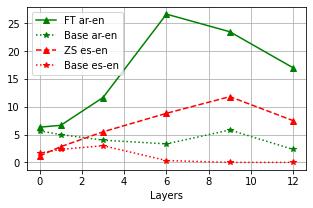}
         \caption{zero-shot \textbf{es} on \textbf{ar} decoder}
         \label{fig:ar-en-es-en-decoder-arabic-model-cluster-alignment}
     \end{subfigure}
     \hfill
     \begin{subfigure}[b]{0.3\textwidth}
         \centering
         \includegraphics[width=\textwidth]{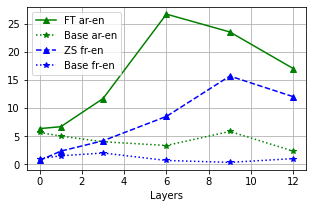}
         \caption{zero-shot \textbf{fr} on \textbf{ar} decoder}
         \label{fig:ar-en-fr-en-decoder-arabic-model-cluster-alignment}
     \end{subfigure}

     \caption{Percentage of Aligned Concepts: Dotted lines represent base models, solid lines denote fine-tuned Arabic–English model, and dashed lines depict zero-shot alignment for German-English (left column), French–English (Middle column) and Arabic-English (right column); enc: Encoder, dec: Decoder}
    \label{fig:arabic-model-alignment}
\end{figure*}

\paragraph{The task-specific calibration of the latent space facilitates zero-shot capabilities.}  In Figures \ref{fig:german-model-alignment}--\ref{fig:arabic-model-alignment}, we display alignment outcomes using mT5 base models and after tuning them for the machine translation task. We examine language alignment within the encoder, decoder, and between the encoder and decoder. We observe that fine-tuning the models enhances the alignment of latent spaces. Interestingly, this increase in alignment also extends to other languages, despite the fact that the model was not specifically tuned for these zero-shot languages.



% \begin{figure*}
%      \centering
%      \begin{subfigure}[b]{0.3\textwidth}
%          \centering
%          \includegraphics[width=\textwidth]{figures/decoder-decoder-plots/ar-en-parallel/es-en.png}
%          \caption{}
%          \label{fig:es-en-decoder-arabic-model}
%      \end{subfigure}
%      \hfill
%      \begin{subfigure}[b]{0.3\textwidth}
%          \centering
%          \includegraphics[width=\textwidth]{figures/decoder-decoder-plots/ar-en-parallel/fr-en.png}
%          \caption{}
%          \label{fig:fr-en-decoder-arabic-model}
%      \end{subfigure}
%      \hfill
%      \begin{subfigure}[b]{0.3\textwidth}
%          \centering
%          \includegraphics[width=\textwidth]{figures/decoder-decoder-plots/ar-en-parallel/de-en.png}
%          \caption{}
%          \label{fig:de-en-decoder-arabic-model}
%      \end{subfigure}
%     \caption{Cluster alignment on decoder side of the Arabic-English finetuned mT5 model. Each subfigure shows the comparison of the clusters alignment between the Arabic-English language split and a zero-shot language split (a) Shows the comparison with the Spanish-English data split (b) shows the comparison with the French-English data split. (c) Shows the comparison with the German-English data split}
%     \label{fig:encoder-encoder-arabic-model}
% \end{figure*}


\begin{figure*}
     \centering
     \begin{subfigure}[b]{0.24\textwidth}
         \centering
         \includegraphics[width=\textwidth]{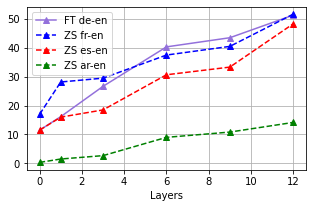}
         \caption{0-shot \textbf{fr, es, ar} on \textbf{de} enc}
         \label{fig:multilinguality-german-model-encoder}
     \end{subfigure}
     \begin{subfigure}[b]{0.24\textwidth}
         \centering
         \includegraphics[width=\textwidth]{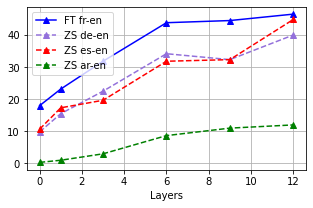}
         \caption{0-shot \textbf{de, es, ar} on \textbf{fr} enc}
         \label{fig:french-model-multilinguality-encoder}
     \end{subfigure}
   \begin{subfigure}[b]{0.24\textwidth}
         \centering
         \includegraphics[width=\textwidth]{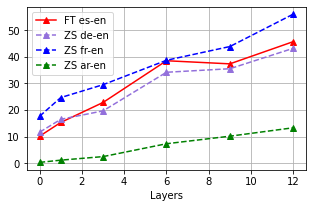}
         \caption{0-shot \textbf{de, fr, ar} on \textbf{es} enc}
         \label{fig:spanish-model-multilinguality-encoder}
     \end{subfigure}
     \begin{subfigure}[b]{0.24\textwidth}
         \centering
         \includegraphics[width=\textwidth]{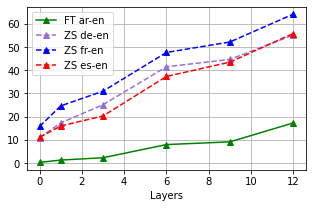}
         \caption{0-shot \textbf{de, fr, es} on \textbf{ar} enc}
         \label{fig:arabic-model-multilinguality-encoder}
     \end{subfigure}
          \begin{subfigure}[b]{0.24\textwidth}
         \centering
         \includegraphics[width=\textwidth]{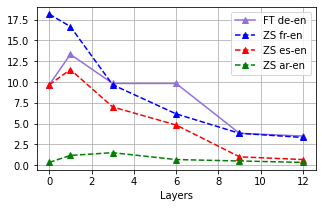}
         \caption{0-shot \textbf{fr, es, ar} on \textbf{de} dec}
         \label{fig:mutlilinguality-german-model-decoder}
     \end{subfigure}
     \begin{subfigure}[b]{0.24\textwidth}
         \centering
         \includegraphics[width=\textwidth]{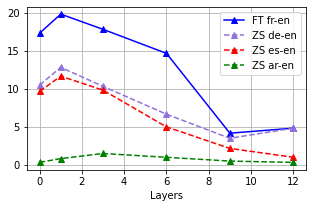}
         \caption{0-shot \textbf{de, es, ar} on \textbf{fr} dec}
         \label{fig:french-model-multilinguality-decoder}
     \end{subfigure} 
       \begin{subfigure}[b]{0.24\textwidth}
         \centering
         \includegraphics[width=\textwidth]{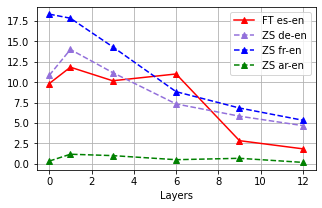}
         \caption{0-shot \textbf{de, fr, ar} on \textbf{es} dec}
         \label{fig:spanish-model-multilinguality-decoder}
     \end{subfigure} 
      \begin{subfigure}[b]{0.24\textwidth}
         \centering
         \includegraphics[width=\textwidth]{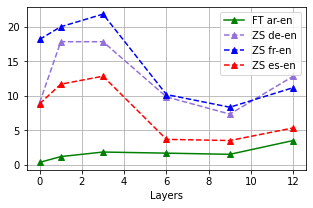}
         \caption{0-shot \textbf{de, fr, es} on \textbf{ar} dec}
         \label{fig:arabic-model-multilinguality-decoder}
     \end{subfigure} 
     \caption{Quantifying Concept Overlap in different languages in \texttt{mT5} encoder and decoders.}
    \label{fig:multilinguality-zs-plots}
\end{figure*}

\begin{figure*}
    \centering
      \begin{subfigure}[b]{0.3\textwidth}
         \centering
         \includegraphics[width=\textwidth]{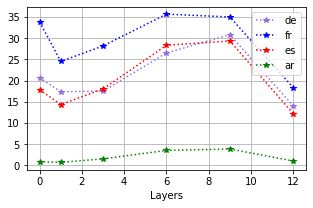}
         \caption{XLM-R Base}
         \label{fig:multilinguality-xlm-base}
     \end{subfigure}
     \begin{subfigure}[b]{0.3\textwidth}
         \centering
         \includegraphics[width=\textwidth]{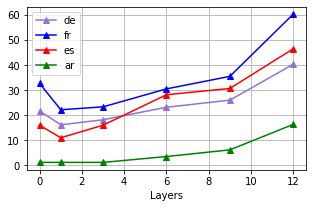}
         \caption{XLM-R (NER)}
         \label{fig:mutlilinguality-xlm-finetuned}
     \end{subfigure}
    \begin{subfigure}[b]{0.3\textwidth}
         \centering
         \includegraphics[width=\textwidth]{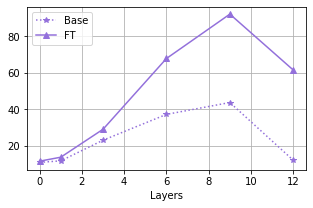}
         \caption{XLM-R (SST)}
         \label{fig:mutlilinguality-xlm-finetuned-sst}
     \end{subfigure}
    \caption{Quantifying Concept Overlap in \texttt{XLM-R}}
    \label{fig:xlm-multilinguality}
\end{figure*}

\section{Concept Multilinguality}
\label{sec:appendix:CM}

In Section \ref{subsec:concept_multilinguality_analysis}, we illustrated how both the base and fine-tuned models manifest concepts with overlapping latent spaces. Figure \ref{fig:multilinguality-zs-plots} showcases that these models display similar patterns even in the zero-shot scenario. Specifically, in this figure, we present the multilinguality of concepts in the mT5 encoder and decoder models under zero-shot conditions. Notably, we observe that the zero-shot overlap (depicted by dashed lines) follows a comparable pattern to the overlap of latent spaces after fine-tuning (indicated by solid lines).

\begin{figure*}
     \centering
     \begin{subfigure}{0.3\textwidth}
         \centering
         \includegraphics[width=\textwidth]{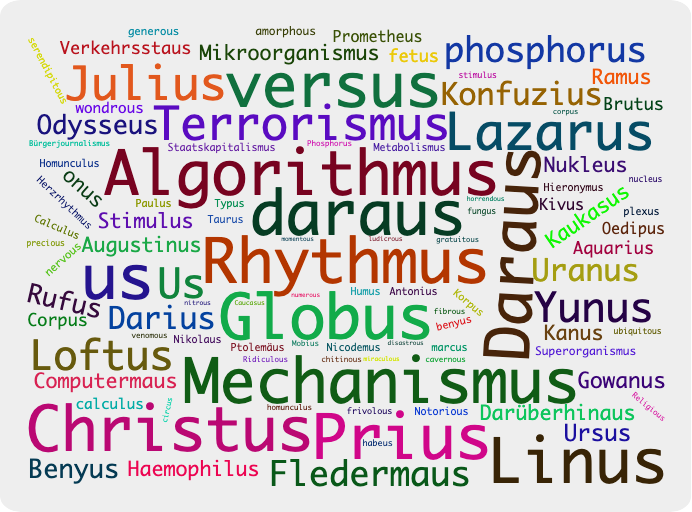}
         \caption{Words ending with ``us''}
         \label{fig:mtm-de-en-0-encoder-c35}
     \end{subfigure}
     \hfill
     \begin{subfigure}{0.3\textwidth}
         \centering
         \includegraphics[width=\textwidth]{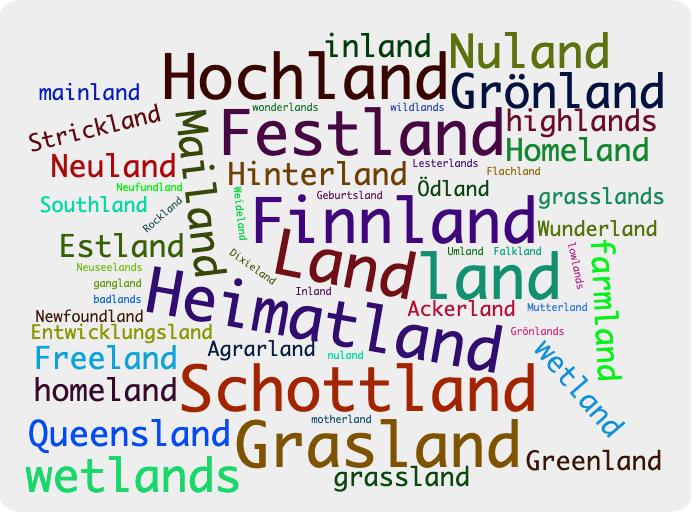}
         \caption{Words containing ``land''}
         \label{fig:mtm-de-en-0-encoder-c65}
     \end{subfigure}
     \hfill
     \begin{subfigure}{0.3\textwidth}
         \centering
         \includegraphics[width=\textwidth]{figures/many-to-many-sample-clusters/de-en/mtm-de-en-0-encoder-c101.png}
         \caption{``ge'' infix}
         \label{fig:mtm-de-en-0-encoder-c101}
     \end{subfigure}

      \begin{subfigure}{0.3\textwidth}
         \centering
         \includegraphics[width=\textwidth]{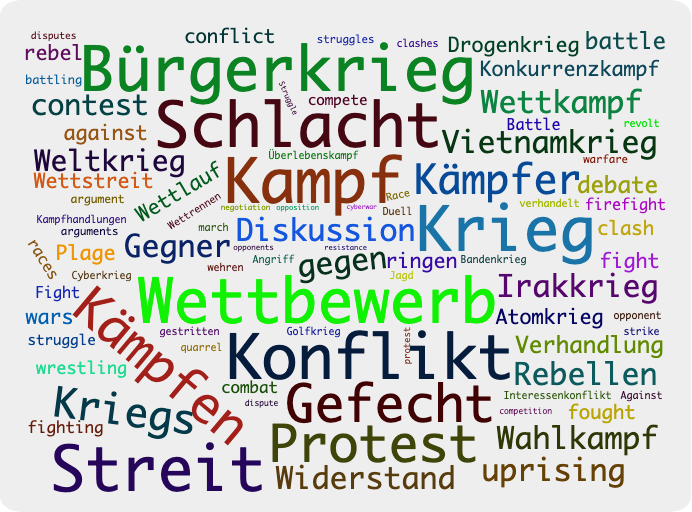}
         \caption{Conflict and competition}
         \label{fig:mtm-de-en-12-encoder-c3}
     \end{subfigure}
        \hfill
      \begin{subfigure}{0.3\textwidth}
         \centering
         \includegraphics[width=\textwidth]{figures/many-to-many-sample-clusters/de-en/mtm-de-en-12-encoder-c291.png}
         \caption{Qualities and Numbers}
         \label{fig:mtm-de-en-12-encoder-c291}
     \end{subfigure}
     \hfill
      \begin{subfigure}{0.3\textwidth}
         \centering
         \includegraphics[width=\textwidth]{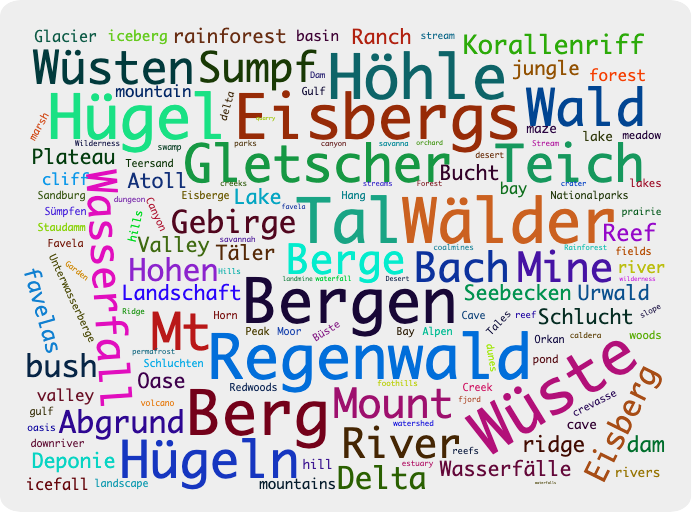}
         \caption{Landforms and Natural Features}
         \label{fig:mtm-de-en-12-encoder-c35}
     \end{subfigure}

      \begin{subfigure}{0.3\textwidth}
         \centering
         \includegraphics[width=\textwidth]{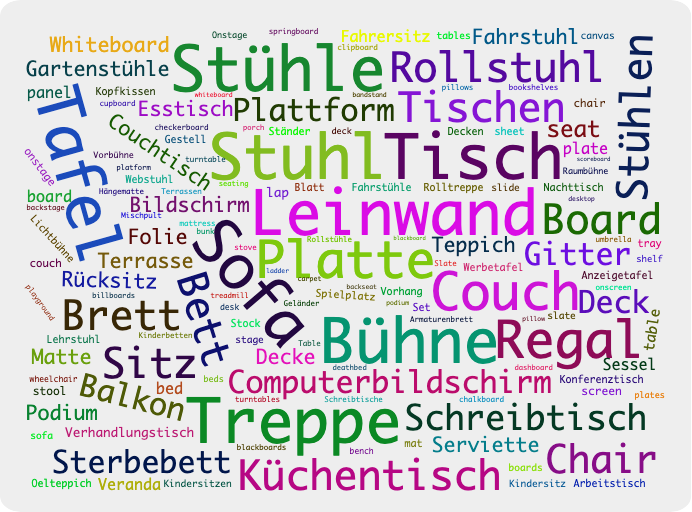}
         \caption{Furniture and Surfaces}
         \label{fig:mtm-de-en-12-encoder-c45}
     \end{subfigure}
     \hfill
     \begin{subfigure}{0.3\textwidth}
         \centering
         \includegraphics[width=\textwidth]{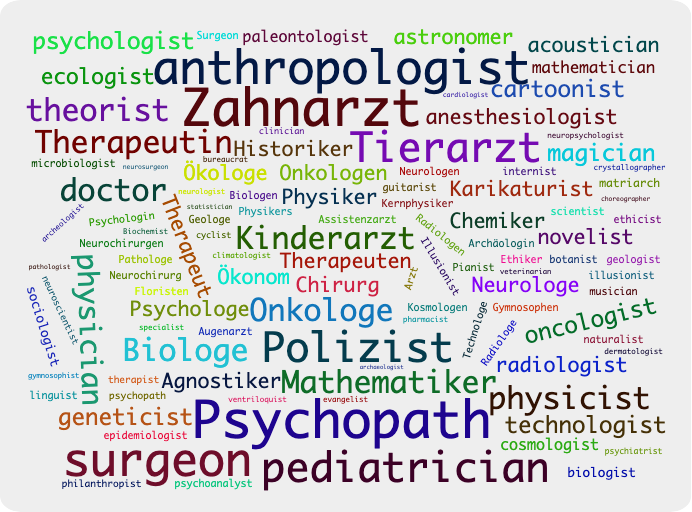}
         \caption{Medical and Scientific professions}
         \label{fig:mtm-de-en-12-encoder-c205}
     \end{subfigure}
     \hfill 
    \begin{subfigure}{0.3\textwidth}
         \centering
         \includegraphics[width=\textwidth]{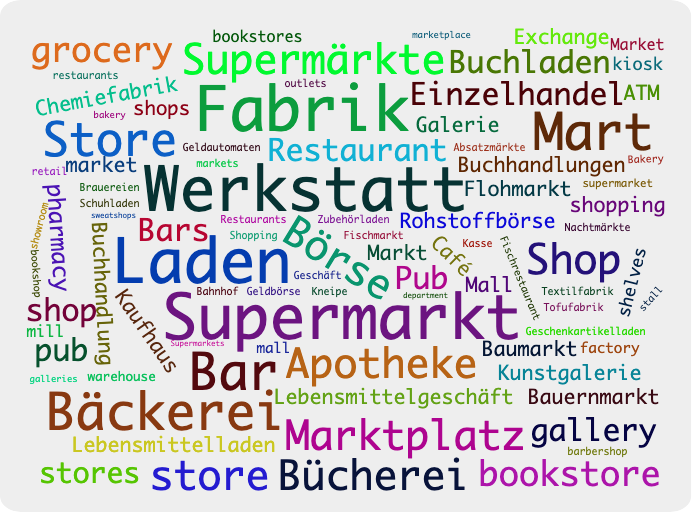}
         \caption{Commercial Establishments}
         \label{fig:mtm-de-en-12-encoder-c209}
     \end{subfigure}
     
    \begin{subfigure}{0.3\textwidth}
         \centering
         \includegraphics[width=\textwidth]{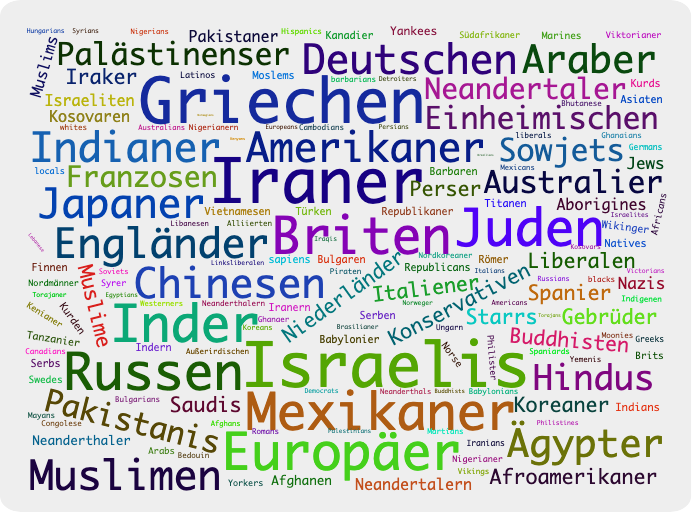}
         \caption{Nationalities and Ethnicities}
         \label{fig:mtm-de-en-12-encoder-c314}
     \end{subfigure}
    \hfill
     \begin{subfigure}{0.3\textwidth}
         \centering
         \includegraphics[width=\textwidth]{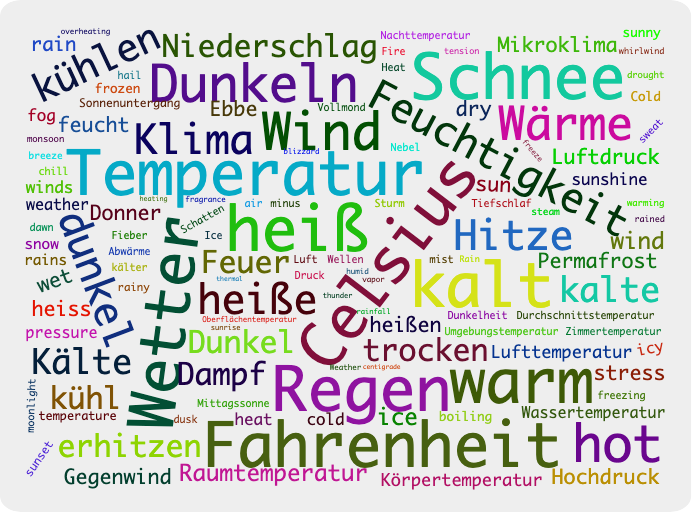}
         \caption{Weather and Tempratures}
         \label{fig:mtm-de-en-12-encoder-c333}
     \end{subfigure}
     \hfill
     \begin{subfigure}{0.3\textwidth}
         \centering
         \includegraphics[width=\textwidth]{figures/many-to-many-sample-clusters/de-en/mtm-de-en-12-encoder-c370.png}
         \caption{Units of Measurement}
         \label{fig:mtm-de-en-12-encoder-c370}
     \end{subfigure}
 
    \caption{Overlapping German-English Concepts in the MT-tuned mT5 model}
    \label{fig:gr_en_mixed_clusters}
\end{figure*}

\begin{figure*}
     \centering
     \begin{subfigure}{0.3\textwidth}
         \centering
         \includegraphics[width=\textwidth]{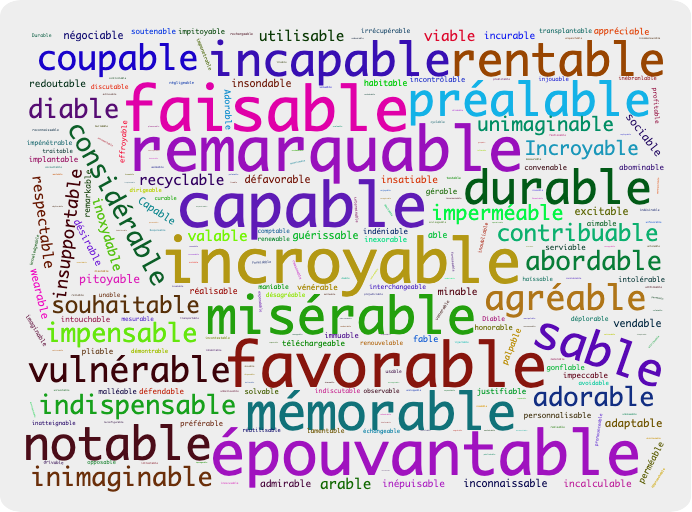}
         \caption{Words ending with ``able''}
         \label{fig:mtm-fr-en-0-encoder-c24}
     \end{subfigure}
     \hfill
     \begin{subfigure}{0.3\textwidth}
         \centering
         \includegraphics[width=\textwidth]{figures/many-to-many-sample-clusters/fr-en/mtm-fr-en-0-encoder-c172.png}
         \caption{Words ending with ``tive''}
         \label{fig:mtm-fr-en-0-encoder-c172}
     \end{subfigure}
     \hfill
     \begin{subfigure}{0.3\textwidth}
         \centering
         \includegraphics[width=\textwidth]{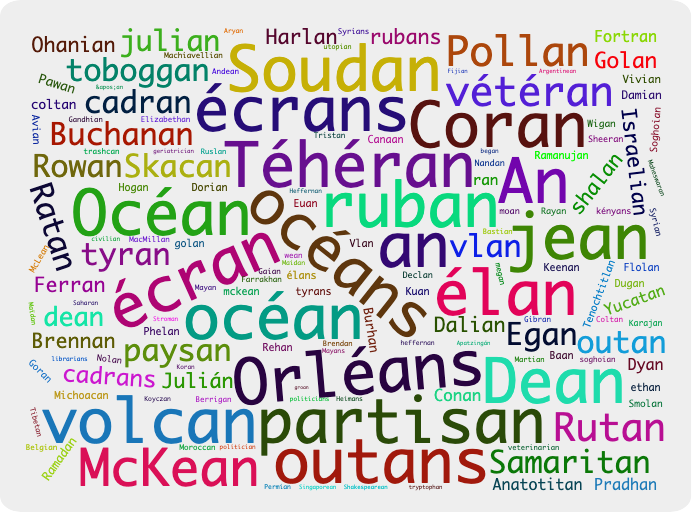}
         \caption{words ending with ``an''}
         \label{fig:mtm-fr-en-0-encoder-c262}
     \end{subfigure}

      \begin{subfigure}{0.3\textwidth}
         \centering
         \includegraphics[width=\textwidth]{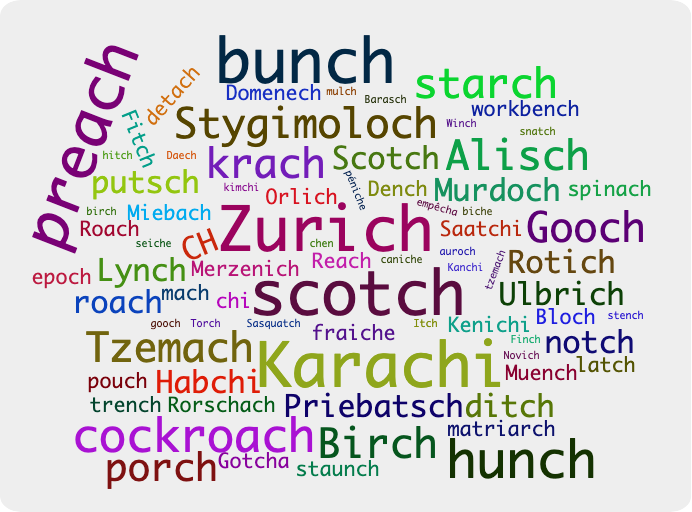}
         \caption{words ending with ``ch''}
         \label{fig:mtm-fr-en-0-encoder-c448}
     \end{subfigure}
        \hfill
      \begin{subfigure}{0.3\textwidth}
         \centering
         \includegraphics[width=\textwidth]{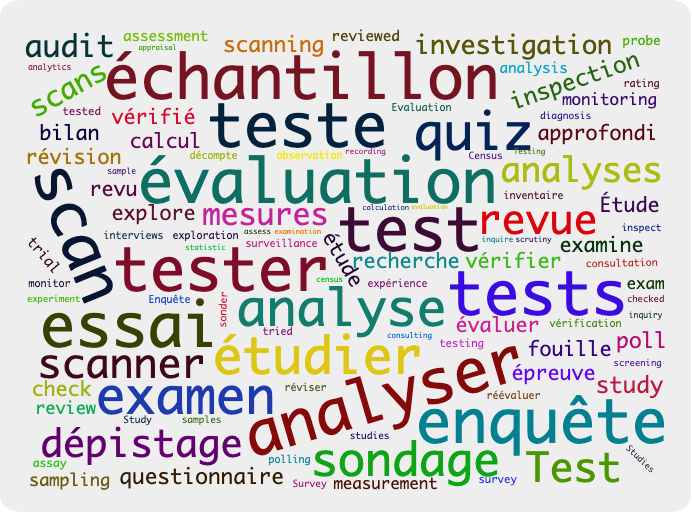}
         \caption{Research Terminology}
         \label{fig:mtm-fr-en-12-encoder-c52}
     \end{subfigure}
     \hfill
      \begin{subfigure}{0.3\textwidth}
         \centering
         \includegraphics[width=\textwidth]{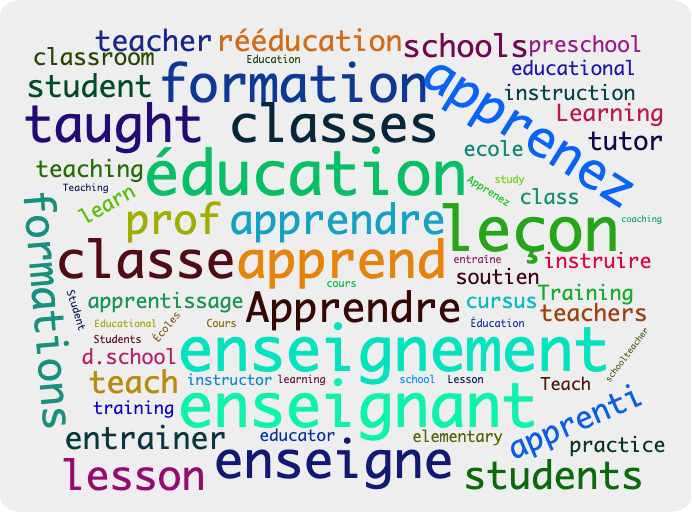}
         \caption{Educational terms}
         \label{fig:mtm-fr-en-12-encoder-c68}
     \end{subfigure}

      \begin{subfigure}{0.3\textwidth}
         \centering
         \includegraphics[width=\textwidth]{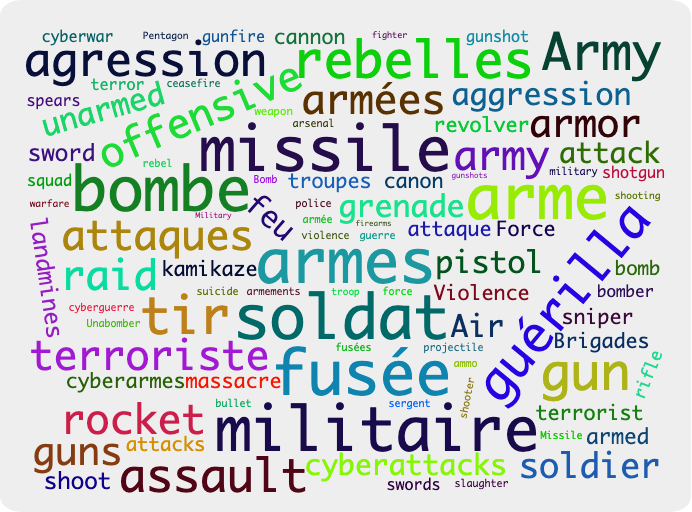}
         \caption{Military and Violence}
         \label{fig:mtm-fr-en-12-encoder-320}
     \end{subfigure}
     \hfill
     \begin{subfigure}{0.3\textwidth}
         \centering
         \includegraphics[width=\textwidth]{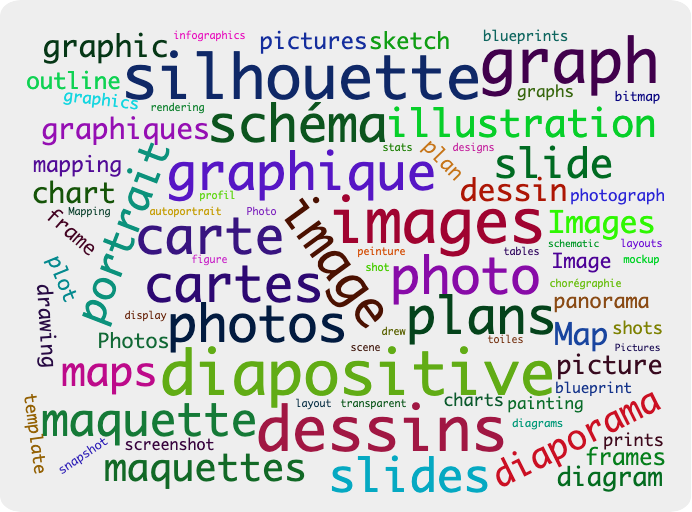}
         \caption{Visual representation vocabulary}
         \label{fig:mtm-fr-en-12-encoder-c325}
     \end{subfigure}
     \hfill 
    \begin{subfigure}{0.3\textwidth}
         \centering
         \includegraphics[width=\textwidth]{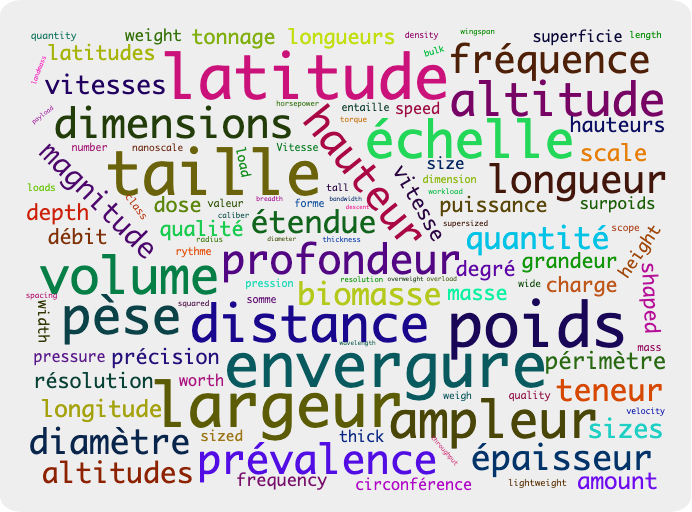}
         \caption{Measurements Vocabulary}
         \label{fig:mtm-fr-en-12-encoder-c394}
     \end{subfigure}
     
    \begin{subfigure}{0.3\textwidth}
         \centering
         \includegraphics[width=\textwidth]{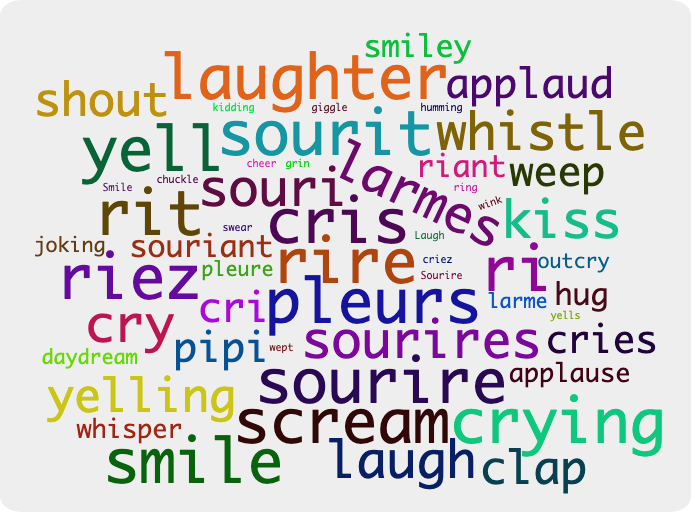}
         \caption{Emotional Expression}
         \label{fig:mtm-fr-en-12-encoder-c413}
     \end{subfigure}
    \hfill
     \begin{subfigure}{0.3\textwidth}
         \centering
         \includegraphics[width=\textwidth]{figures/many-to-many-sample-clusters/fr-en/mtm-fr-en-12-encoder-c420.png}
         \caption{Chemical Elements}
         \label{fig:mtm-fr-en-12-encoder-c420}
     \end{subfigure}
     \hfill
     \begin{subfigure}{0.3\textwidth}
         \centering
         \includegraphics[width=\textwidth]{figures/many-to-many-sample-clusters/fr-en/mtm-fr-en-12-encoder-c451.png}
         \caption{Modes of Transportation}
         \label{fig:mtm-fr-en-12-encoder-c451}
     \end{subfigure}
 
    \caption{Overlapping French-English Concepts in the MT-tuned mT5 model}
    \label{fig:fr_en_mixed_clusters}
\end{figure*}

\begin{figure*}
     \centering
     \begin{subfigure}{0.3\textwidth}
         \centering
         \includegraphics[width=\textwidth]{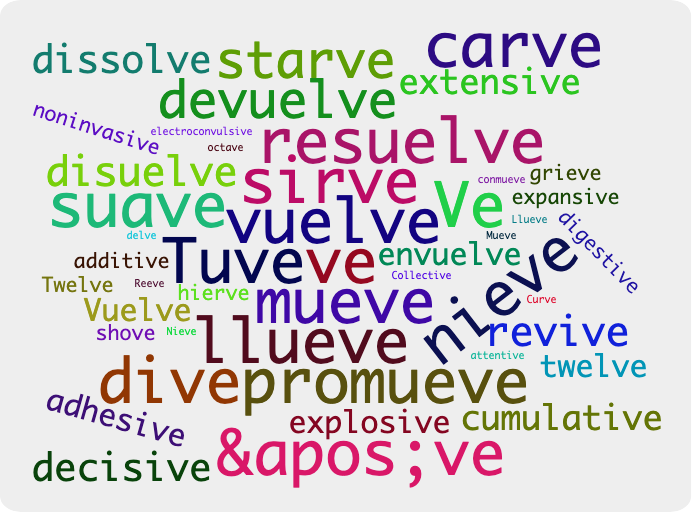}
         \caption{Words containing ``ve''}
         \label{fig:mtm-es-en-0-encoder-c64}
     \end{subfigure}
     \hfill
     \begin{subfigure}{0.3\textwidth}
         \centering
         \includegraphics[width=\textwidth]{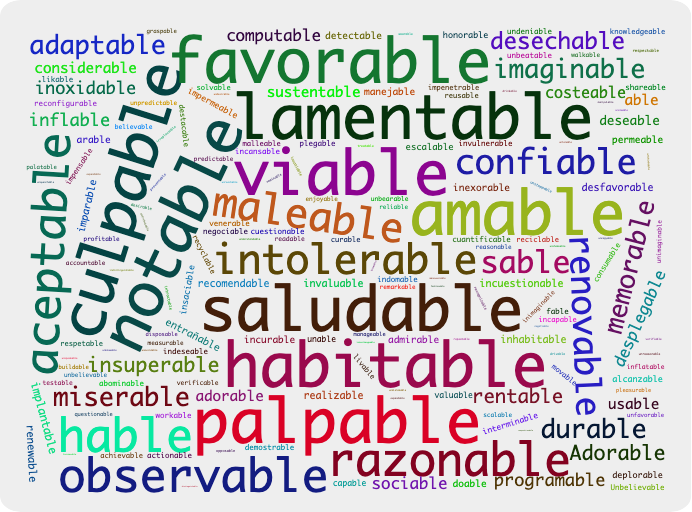}
         \caption{words containing "able"}
         \label{fig:mtm-es-en-0-encoder-c81}
     \end{subfigure}
     \hfill
     \begin{subfigure}{0.3\textwidth}
         \centering
         \includegraphics[width=\textwidth]{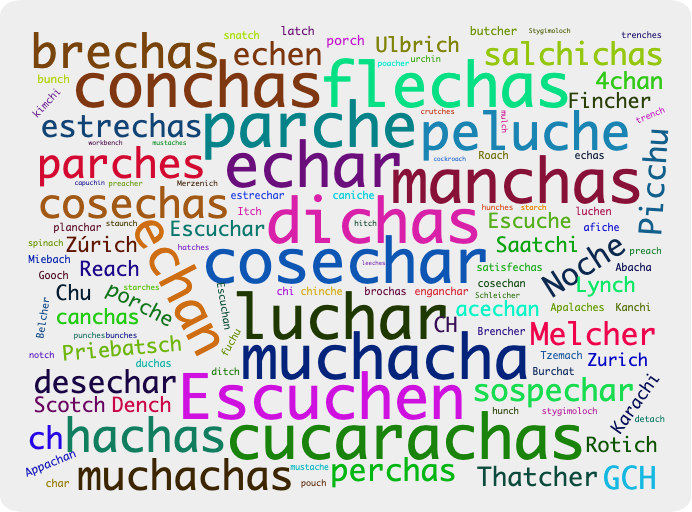}
         \caption{"ch" infix}
         \label{fig:mtm-es-en-0-encoder-425}
     \end{subfigure}

      \begin{subfigure}{0.3\textwidth}
         \centering
         \includegraphics[width=\textwidth]{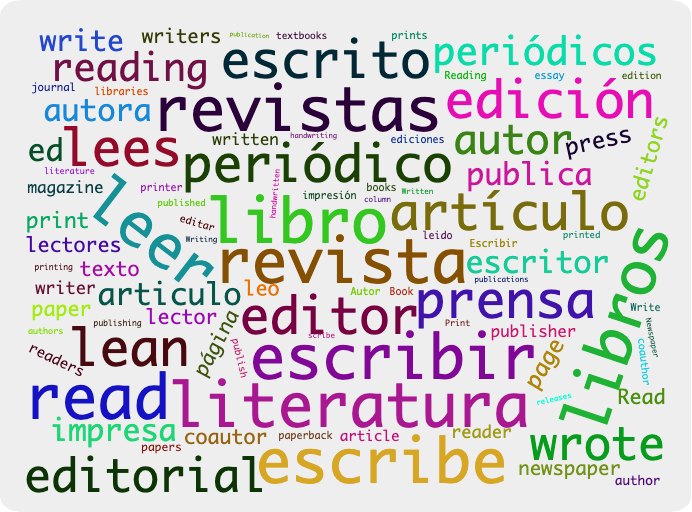}
         \caption{Literature Writing and Vocabulary}
         \label{fig:mtm-es-en-12-encoder-c16}
     \end{subfigure}
        \hfill
      \begin{subfigure}{0.3\textwidth}
         \centering
         \includegraphics[width=\textwidth]{figures/many-to-many-sample-clusters/es-en/mtm-es-en-12-encoder-c28.png}
         \caption{Family and Relationships}
         \label{fig:mtm-es-en-12-encoder-c28}
     \end{subfigure}
     \hfill
      \begin{subfigure}{0.3\textwidth}
         \centering
         \includegraphics[width=\textwidth]{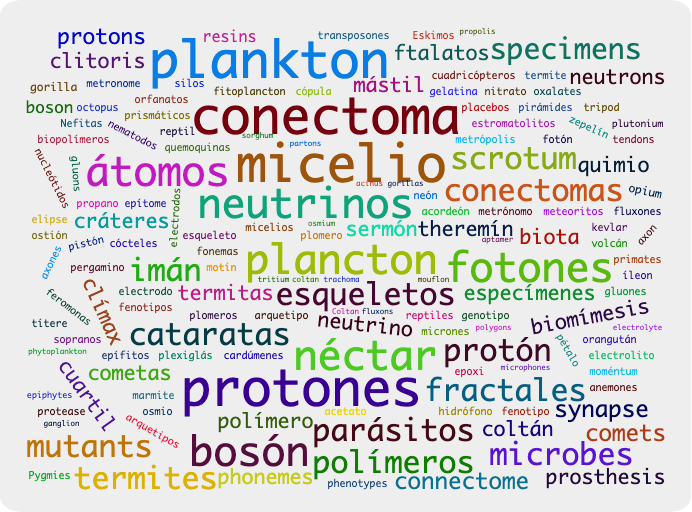}
         \caption{Scientific Terms}
         \label{fig:mtm-es-en-12-encoder-c66}
     \end{subfigure}

      \begin{subfigure}{0.3\textwidth}
         \centering
         \includegraphics[width=\textwidth]{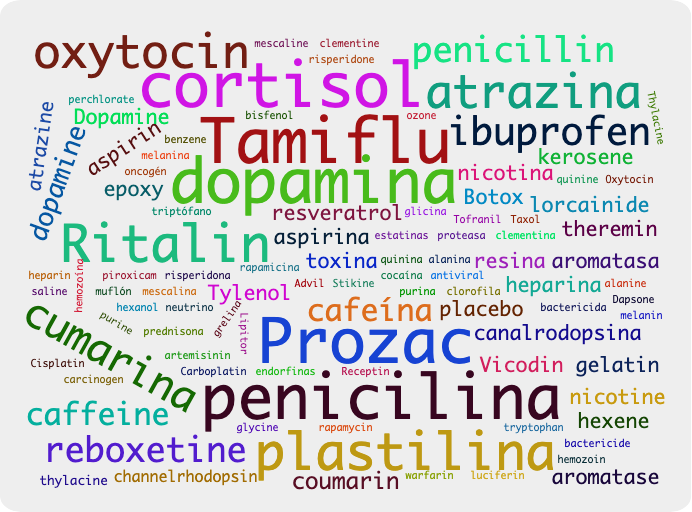}
         \caption{Chemical compounds}
         \label{fig:mtm-es-en-12-encoder-c73}
     \end{subfigure}
     \hfill
     \begin{subfigure}{0.3\textwidth}
         \centering
         \includegraphics[width=\textwidth]{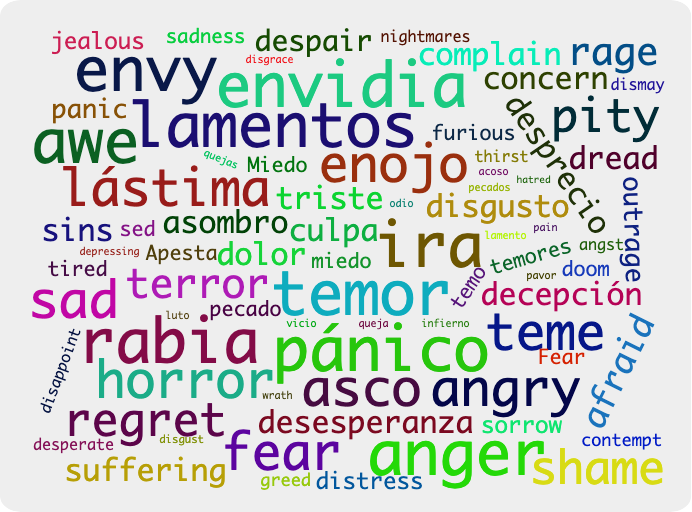}
         \caption{Emotions and States of mind}
         \label{fig:mtm-es-en-12-encoder-c120}
     \end{subfigure}
     \hfill 
    \begin{subfigure}{0.3\textwidth}
         \centering
         \includegraphics[width=\textwidth]{figures/many-to-many-sample-clusters/es-en/mtm-es-en-12-encoder-c211.png}
         \caption{Technological devices and tools}
         \label{fig:mtm-es-en-12-encoder-c211}
     \end{subfigure}
    \caption{Overlapping Spanish-English Concepts in the MT-tuned mT5 model}
    \label{fig:es_en_mixed_clusters_mtm}
\end{figure*}

\begin{figure*}
     \centering
     \begin{subfigure}{0.3\textwidth}
         \centering
         \includegraphics[width=\textwidth]{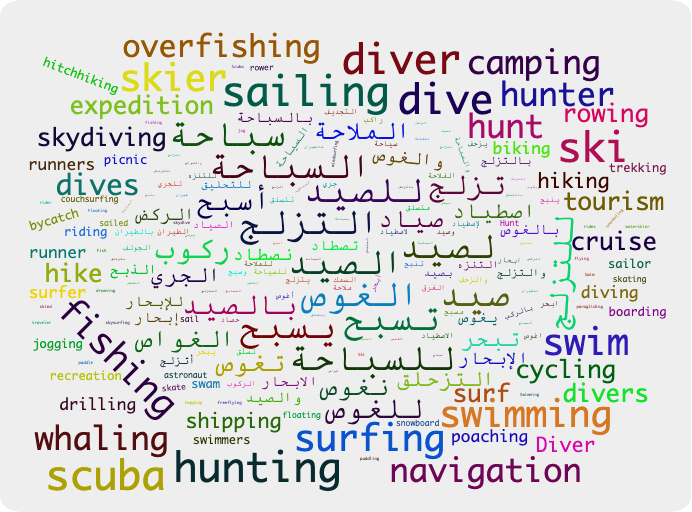}
         \caption{Recreational Sports and Activities}
         \label{fig:mtm-ar-en-12-encoder-c119}
     \end{subfigure}
     \hfill
     \begin{subfigure}{0.3\textwidth}
         \centering
         \includegraphics[width=\textwidth]{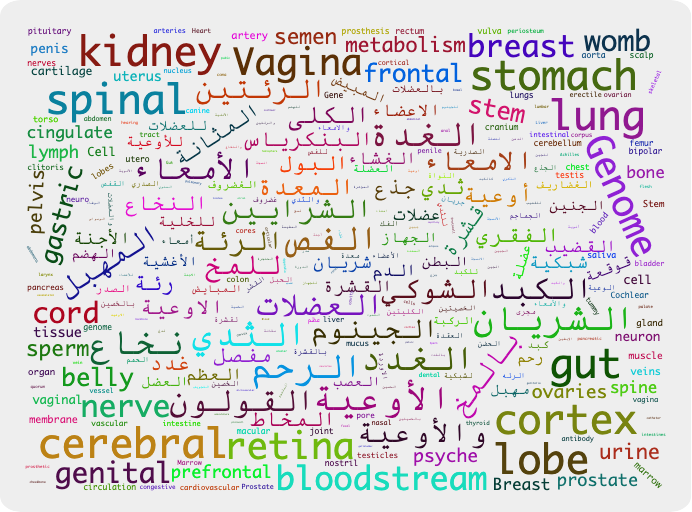}
         \caption{Anatomical Terminology}
         \label{fig:mtm-ar-en-12-encoder-c23}
     \end{subfigure}
     \hfill
     \begin{subfigure}{0.3\textwidth}
         \centering
         \includegraphics[width=\textwidth]{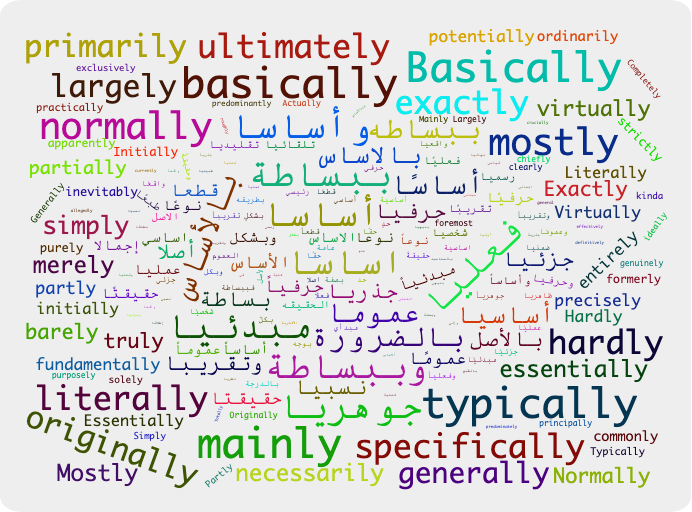}
         \caption{Adverbs of Emphasis and Certainty}
         \label{fig:mtm-ar-en-12-encoder-c27}
     \end{subfigure}

      \begin{subfigure}{0.3\textwidth}
         \centering
         \includegraphics[width=\textwidth]{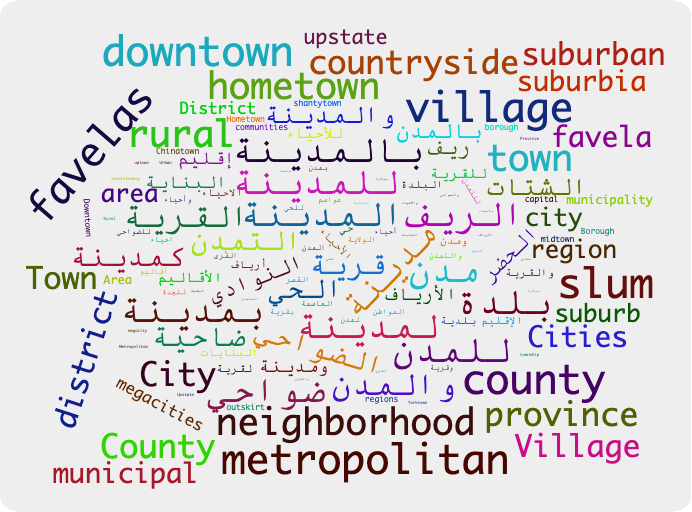}
         \caption{Geographical and Urban Terms}
         \label{fig:mtm-ar-en-12-encoder-c65}
     \end{subfigure}
        \hfill
      \begin{subfigure}{0.3\textwidth}
         \centering
         \includegraphics[width=\textwidth]{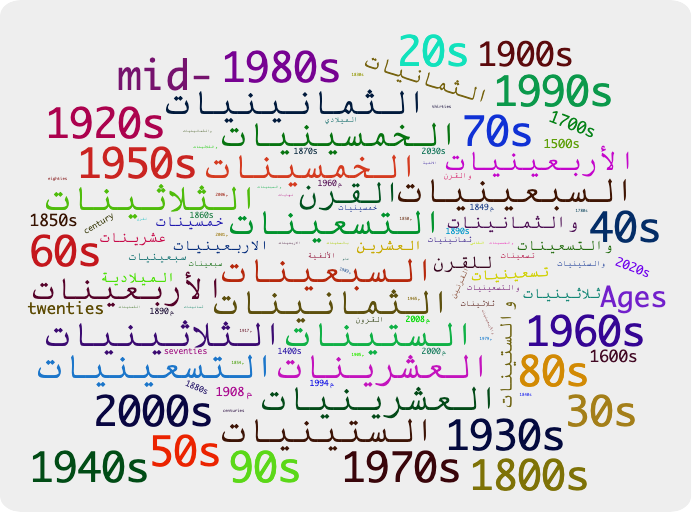}
         \caption{Time periods and Decades}
         \label{fig:mtm-ar-en-12-encoder-c114}
     \end{subfigure}
     \hfill
      \begin{subfigure}{0.3\textwidth}
         \centering
         \includegraphics[width=\textwidth]{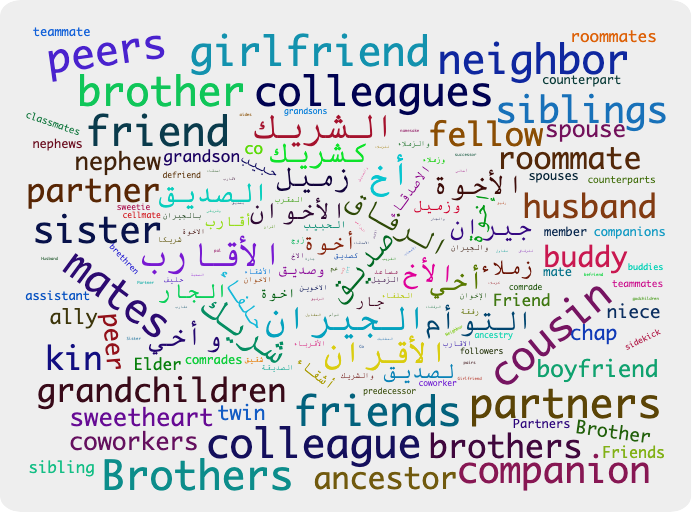}
         \caption{Relationships and Connections}
         \label{fig:mtm-ar-en-12-encoder-c277}
     \end{subfigure}

      \begin{subfigure}{0.3\textwidth}
         \centering
         \includegraphics[width=\textwidth]{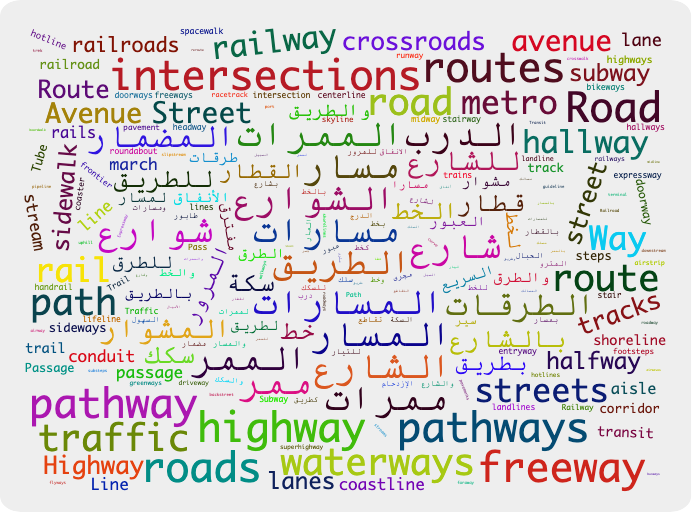}
         \caption{Paths and Transportation}
         \label{fig:mtm-ar-en-12-encoder-c349}
     \end{subfigure}
     \hfill
     \begin{subfigure}{0.3\textwidth}
         \centering
         \includegraphics[width=\textwidth]{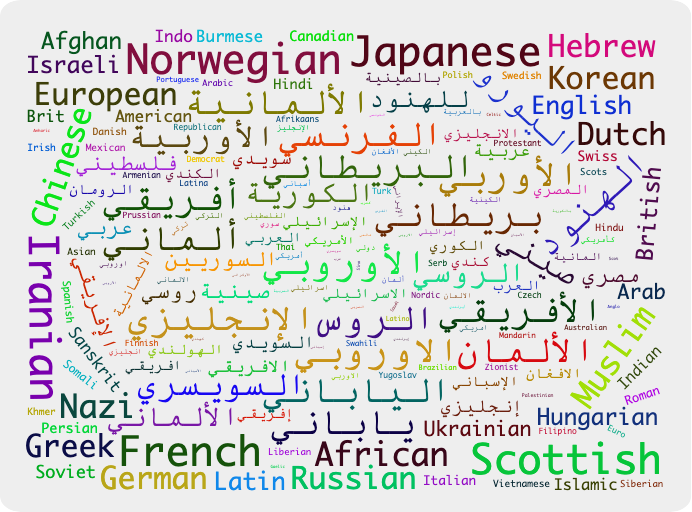}
         \caption{Nationalities and Ethnicities}
         \label{fig:mtm-ar-en-12-encoder-c582}
     \end{subfigure}
     \hfill 
    \begin{subfigure}{0.3\textwidth}
         \centering
         \includegraphics[width=\textwidth]{figures/many-to-many-sample-clusters/ar-en/mtm-ar-en-12-encoder-c409.png}
         \caption{Units of Measurment}
         \label{fig:mtm-ar-en-12-encoder-c409}
     \end{subfigure}
    \caption{Overlapping Arabic-English Concepts in the MT-tuned mT5 model}
    \label{fig:ar_en_mixed_clusters_mtm}
\end{figure*}

\section{Thresholds}
\label{sec:appendix:thresholds}

\begin{figure*}
    \centering

      \begin{subfigure}[b]{0.24\textwidth}
         \centering
         \includegraphics[width=\textwidth]{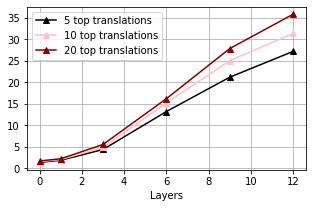}
         \caption{N-best translations}
         \label{fig:de-en-encoder-varying-top-n}
     \end{subfigure}
     \begin{subfigure}[b]{0.24\textwidth}
         \centering
         \includegraphics[width=\textwidth]{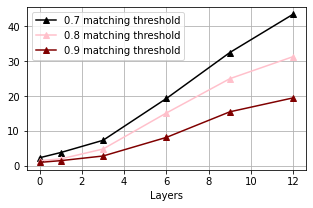}
         \caption{Matching threshold}
         \label{fig:de-en-encoder-vary-matching-threshold}
     \end{subfigure}
    \begin{subfigure}[b]{0.24\textwidth}
         \centering
         \includegraphics[width=\textwidth]{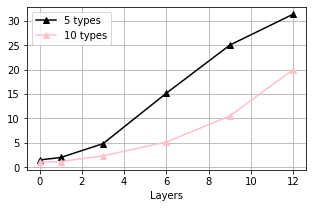}
         \caption{Minimum types per concept}
         \label{fig:de-en-encoder-vary-types}
     \end{subfigure}
      \begin{subfigure}[b]{0.24\textwidth}
         \centering
         \includegraphics[width=\textwidth]{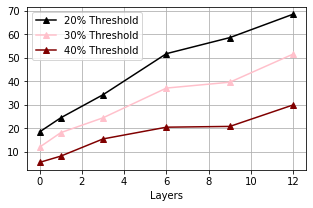}
         \caption{Overlapping threshold}
         \label{fig:multilinguality-thresholds}
     \end{subfigure}
     
    \caption{Varying different threshold parameters in \CA{} and \CO{}}
    \label{fig:parameter-variations}
\end{figure*}

In Section \ref{subsec:threshold} we mentioned the threshold we used for our experiments including the matching threshold, n-best translations to estimate $\mathcal{T}(w_s,w_t)$ and minimum number of types per concept.  The choice of these parameters is arbitrary. We experimented with various configurations, such as using a 70--90\% matching types, using 5--20 best translations. The overall patterns of the results remained consistent across different configurations (please refer to Figure \ref{fig:parameter-variations}). The selected thresholds were chosen based on a qualitative examination of the concepts, allowing for some noise in the concept representations.

\section{Data Statistics}
\label{sec:sppendix:data-stats}

In this section, we report the data statistics that we used for the experiment. Table \ref{table:ted-data-stats} shows the number of sentences for the TED data \cite{birch-etal-2014-edinburgh} used for the machine translation experiments, Table \ref{table:xtreme-data-statistics} shows the statistics for the NER data used, and Table \ref{table:sst2-data-stats} shows the statistics for the sentiment analysis data used.

\section{Computing Budget}
\label{sec:computing-budget}

The extraction of the representations from a multilingual model requires 500GB of RAM memory. The clustering experiments for the extracted representations require 30GB of RAM memory each. 

% \usepackage{booktabs}
\begin{table*}[!ht]
\centering
\begin{tabular}{@{}lrrrr@{}}
\toprule
                                                                     
                      & \multicolumn{1}{l}{de-en} & \multicolumn{1}{l}{es-en} & \multicolumn{1}{l}{fr-en} & \multicolumn{1}{l}{ar-en} \\ \midrule
train     & 209330                    & 184724                    & 234033                    & 229194                    \\ \midrule
test11 & 1433                      & 1435                      & 818                       & 1199                      \\
test12 & 1700                      & 1701                      & 1124                      & 1702                      \\
test13 & 992                       & 1197                      & 1026                      & 1167                      \\
test14 & 1305                      & 1305                      & 1305                      & 1107    \\
\bottomrule
\end{tabular}
\caption{TED data statistics (number of sentences).}
\label{table:ted-data-stats}
\end{table*}
\begin{table*}[]
\centering
\begin{tabular}{@{}lrrrrr@{}}
                                      \\ \toprule
                       & \multicolumn{1}{l}{de} & \multicolumn{1}{l}{es} & \multicolumn{1}{l}{fr} & \multicolumn{1}{l}{ar} & \multicolumn{1}{l}{en} \\ \midrule
train - sentences      & 20000                  & 20000                  & 20000                  & 20000                  & 20000                  \\
train - tokens         & 195387                 & 129283                 & 136788                 & 129184                 & 160394                 \\
validation - sentences & 10000                  & 10000                  & 10000                  & 10000                  & 10000                  \\
validation - tokens    & 97805                  & 64329                  & 68220                  & 64291                  & 80536                  \\
test - sentences       & 10000                  & 10000                  & 10000                  & 10000                  & 10000                  \\
test - tokens          & 97646                  & 64727                  & 68754                  & 64347                  & 80326  \\
\bottomrule
\end{tabular}
\caption{Xtreme NER data statisics}
\label{table:xtreme-data-statistics}
\end{table*}

\begin{table*}[]
\centering
\begin{tabular}{@{}lrr@{}}
\toprule
           & \multicolumn{1}{l}{en}  &   \multicolumn{1}{l}{de} \\  \midrule
train      & 67437      &    67437         \\
validation & 872      &        872         \\
test       & 1821     &     1821        \\ \bottomrule   
\end{tabular}
\caption{SST2 data statistics}
\label{table:sst2-data-stats}

\end{table*}

\end{document}